\documentclass[preprint]{elsarticle}
\usepackage[utf8]{inputenc}
\usepackage{amsfonts}
\usepackage{mathtools}
\usepackage{hhline}
\usepackage{longtable,booktabs}
\usepackage{threeparttablex, caption}
\usepackage{hyperref}
\usepackage{graphicx}
\usepackage{amsmath}
\usepackage[export]{adjustbox}
\graphicspath{{Images/}}
\usepackage{subcaption}
\usepackage{array}
\usepackage{xcolor}

\journal{Information Fusion}

\usepackage{ragged2e}
\newcolumntype{P}[1]{>{\RaggedRight\hspace{0pt}}p{#1}}

\begin{document}

\begin{frontmatter}
\title{Exploration in Deep Reinforcement Learning: A Survey\tnoteref{t1}}

\author[1]{Pawel Ladosz}
\ead{pladosz@unist.ac.kr}
\author[2]{Lilian Weng}
\ead{lilian@openai.com}
\author[1]{Minwoo Kim}
\ead{red9395@unist.ac.kr}
\author[1]{Hyondong Oh\corref{cor1}}
\ead{h.oh@unist.ac.kr}

\address[1]{Department of Mechanical Engineering, Ulsan National Institute
of Science and Technology (UNIST), 50 UNIST-gil, Ulsan, Republic of Korea}
\address[2]{OpenAI LP, 3180 18th St, San Francisco, CA 94110}

\tnotetext[t1]{This research was supported by Basic Science Research Program through the National Research Foundation of Korea(NRF) funded by the Ministry of Education (2020R1A6A1A03040570), National Research Foundation of Korea(NRF) grant funded by the Korea government(MSIT) (2020R1F1A1049066), and Unmanned Vehicles Core Technology Research and Development Program through the National Research  Foundation of Korea(NRF), Unmanned Vehicle Advanced Research Center(UVARC) funded by the Ministry of Science and ICT, the Republic of Korea (2020M3C1C1A01082375)}
\cortext[cor1]{Corresponding author}

\begin{abstract}
This paper reviews exploration techniques in deep reinforcement learning. Exploration techniques are of primary importance when solving sparse reward problems. In sparse reward problems, the reward is rare, which means that the agent will not find the reward often by acting randomly. In such a scenario, it is challenging for reinforcement learning to learn rewards and actions association. Thus more sophisticated exploration methods need to be devised. This review provides a comprehensive overview of existing exploration approaches, which are categorized based on the key contributions as follows reward novel states, reward diverse behaviours, goal-based methods, probabilistic methods, imitation-based methods, safe exploration and random-based methods. Then, the unsolved challenges are discussed to provide valuable future research directions. Finally, the approaches of different categories are compared in terms of complexity, computational effort and overall performance.
\end{abstract}

\begin{keyword}
Deep reinforcement learning \sep Exploration \sep Intrinsic motivation \sep Sparse reward problems
\end{keyword}

\end{frontmatter}

\section{Introduction}
In numerous real-world problems, the outcomes of a certain event are only visible after a significant number of other events have occurred. These types of problems are called sparse reward problems since the reward is rare and without a clear link to previous actions. We note that sparse reward problems are common in a real world. For example, during search and rescue missions, the reward is only given when an object is found, or during delivery, the reward is only given when an object is delivered. In sparse reward problems, thousands of decisions might need to be made before the outcomes are visible. Here, we present a review on a group of techniques that can solve this issue, namely exploration in reinforcement learning.
 
In reinforcement learning, an agent is given a state  and a reward from the environment. The task of the agent is to determine an appropriate action. In reinforcement learning, the appropriate action is such that it maximises the reward, or it could be said that the action is exploitative. However, solving problems with just exploitation may not be feasible owing to reward sparseness. With reward sparseness, the agent is unlikely to find a reward quickly, and thus, it has nothing to exploit. Thus, an exploration algorithm is required to solve sparse reward problems. 

The most common technique for exploration in reinforcement learning is random exploration \cite{Sutton2020}. In this type of approach, the agent decides what to do randomly regardless of its progress. The most commonly-used technique of this type, called $\epsilon$-greedy, uses the time decaying parameter $\epsilon$ to reduce exploration over time. This can theoretically solve the sparse reward problem given a sufficient amount of time. However, this is often impractical in real-world applications because learning times can be very large. However, we note that even just with random exploration, deep reinforcement learning has shown some impressive performance in Atari games~\cite{Mnih2015b}, Mujoco simulator~\cite{Lillicrap2016}, controllers tuning~\cite{Lee2020}, autonomous landing~\cite{Polvara2017}, self-driving cars~\cite{Kiran2021} and healthcare~\cite{Yu2019}.

Another solution for exploration could be reward shaping. In reward shaping, the designer 'artificially' imposes a reward more often. For example, for search and rescue missions, agents can be given a negative reward every time they do not find the victim. However, reward shaping is a challenging problem that is heavily dependent on the experience of the designer. Punishing the agent too much could lead to the agent not moving at all \cite{Irpan2018}, while rewarding it too much may cause the agent to repeat certain actions infinitely \cite{Clark2016}. Thus, with the issues of random exploration and reward shaping, there is a need for more sophisticated exploration algorithms.

While exploration in reinforcement learning was considered as early as 1991 \cite{Schmidhuber1991, Schmidhuber1991a}, it is still under development. Recently, exploration has shown a significant gain in performance compared to non-exploratory algorithms: Diversity is all you need (DIYAN)~\cite{Eysenbach2019a} improved on MuJoCo benchmarks; random network distillation (RND)~\cite{Burda2018b} and pseudocounts~\cite{Bellemare2016} were the first to score on difficult Montezuma's Revenge problem; and Agent57~\cite{Badia2020b} is the first to beat humans in all 57 Atari games.

This review focuses on exploratory approaches which fulfil at least one of the following criteria: (i) determines the exploration degree based on the agent's learning, (ii) actively decides to take certain actions in hopes of finding new outcomes, and (iii) motivates itself to continue exploring despite a lack of environmental rewards. In addition, this review focuses on approaches that have been applied to deep reinforcement learning. Note that this review is intended for beginners in exploration for deep reinforcement learning; thus, the focus is on the breadth of approaches and their relatively simplified description. Note also that, throughout the paper, we will use 'reinforcement learning' as it is a more general term rather than 'deep reinforcement learning'. 

Several review articles exist in the field of reinforcement learning. Aubert et al.~\cite{Aubret2019} presented an overview of intrinsic motivation in reinforcement learning, Li~\cite{Li2018c} presented a comprehensive overview of techniques and applications, \citet{Nguyen2018} considered an application to multi-agent problems, \citet{Levine2018a} provided a tutorial and extensive comparison with probabilistic inference methods and \cite{Lazaridis2019} provided an extensive description of the key breakthrough methods in reinforcement learning, including ones in exploration. However, none of the aforementioned reviews focused on exploration or considered it in great detail. The only other review focused on exploration is from 1999 and is now outdated and inaccurate \cite{Mcfarlane1999}. 

The contributions of this study are as follows.  First, the systematic overview of exploration in deep reinforcement learning is presented. As mentioned above, no other modern review exists with this focus. Second, a categorization of exploration in reinforcement learning is provided. The categorization is devised to provide a good way of comparing different approaches. Finally, future challenges are identified and discussed.

\section{Preliminaries}
\subsection{Introduction to Reinforcement Learning}
\subsubsection{Markov Decision Process}
We consider a standard reinforcement setting in which an agent interacts with a stochastic and fully observable environment by sequentially choosing actions in a discrete time step to maximise cumulative rewards. This series of processes is called \textit{Markov decision process } (MDP). An MDP has a tuple of $(S, A, P, R, \gamma)$, where $S$ is a set of states, $A$ is a set of actions the agent can select, $P$ is a transition probability that satisfies the Markov property given as:
\begin{equation}
    p(s_{t+1}|s_1, a_1, s_2, a_2, \cdots, s_t, a_t) = p(s_{t+1}|s_t, a_t),
\end{equation}
$R$ is a set of rewards, and $\gamma \in (0,1]$ is a discount factor. At each time step $t$, an agent receives states $s_t \in S$ from the environment and selects the best possible actions $a_t \in A$ according to policy $\pi(a_t|s_t)$, which maps from states $s_t$ to actions $a_t$. The agent receives a reward $r_t \in R$ from the environment to take an action $a_t$. The goal of the agent is to maximise the discounted expected reward $G_t = \sum_{k=0}^\infty \gamma^k r_{k+t}$ from each state $s_t$.

\subsubsection{Value-Based Methods}

Given that the agent follows policy $\pi$, a state-value function is defined as $V^\pi(s_t) = \mathbb{E}_{\pi}[G_t|S_t = s]$. Similarly, the action-value function, $Q^\pi(s_t,a_t) = \mathbb{E}_{\pi}[G_t|S_t = s, A_t = a]$, is an expected estimate value for a given state $s_t$ for taking an action $a_t$. Q-learning is a typical type of \textit{off-policy learning} that updates a target policy $\pi$ using samples generated by any stochastic behaviour policy in an environment. Following the \textit{Bellman equation} and \textit{temporal difference} (TD) for the action-value function, the Q-learning algorithm is recursively updated using the following equation:
\begin{equation}
    Q(s_t,a_t) = Q(s_t, a_t) + \alpha[r_t + \gamma \max_{a' \in A} Q(s_{t+1}, a') - Q(s_t, a_t)],
\end{equation}
where $a'$ follows the target policy $a' \sim \pi(\cdot|s_t)$ and $\alpha$ is the learning rate. While updating Q-learning, the next actions $a_{t+1}$ are sampled from the behaviour policy which follows an $\epsilon$-greedy exploration strategy, and among them, the action that makes the largest Q-value, $a'$, is selected.

\subsubsection{Policy-Based Methods}

In contrast to value-based methods, policy-based methods directly update the policy parameterized by $\theta$. In reinforcement learning, because the goal is to maximise the expected return throughout states, the objective function for the policy is defined as $J(\theta) = \mathbb{E}_{\pi_\theta}[G_t]$. Williams et al. \cite{Williams1992} suggested the REINFORCE algorithm which updates the policy network by taking a gradient ascent in the direction of $\nabla_{\theta} J(\theta)$. The gradient of the objective function is expressed as:
\begin{equation}
    \nabla_{\theta} J(\theta) = \mathbb{E}_{s \sim p^{\pi}, a \sim \pi_{\theta}}[ \nabla_{\theta} \log \pi_{\theta}(a|s) G_t],
\end{equation}
where $p^{\pi}$ denotes the state distribution. A general overview of reinforcement learning can be found in \cite{Arulkumaran2017}.

\subsection{Exploration}
Exploration can be defined as the activity of searching and finding out about something \cite{Exploration}. In the context of reinforcement learning, "something" is a reward function and the "searching and finding" is an agent's attempt to try to maximise the reward function. Exploration in reinforcement learning is of particular importance because a reward function is often complex and agents are expected to improve over their lifetime. Exploration can take various forms such as randomly taking certain actions and seeing the output, following the best known solution, or actively considering moves that are good for novel discoveries.

Problems that can be solved by exploration are common in nature. Exploration is the act of searching for a solution to a problem. We note that exploration is the most useful in problems in which a route to the actual solution (i.e. reward) is obstructed by the local minima (maxima) or areas of flat rewards. These conditions mean that discovering the true nature of rewards is challenging. The following examples are intuitive illustrations of those problems: (i) search and rescue–the agent needs to explore to find a target (victim); the agent is only rewarded when it finds the victim; otherwise, the reward is 0; and (ii) delivery–trying to deliver an object in the unknown areas; the agent is only rewarded when the appropriate drop-off point has been found; otherwise, the reward is 0. Exploration could be considered as a ubiquitous problem that is highly relevant to many domains with ongoing research.

\subsection{Challenging Problems}
In this section, some of the challenging problems for exploration in reinforcement learning are described, namely noisy-TV and sparse reward problems.

\subsubsection{Noisy-TV}
In a noisy-TV~\cite{Burda2018b} problem, the agent is stuck in exploring an infinite number of states which lead to no reward. This phenomenon can be easily explained with an example. Imagine a state consisting of a virtual TV where the agent can operate the remote, but operating the remote controller leads to no reward. A new random image is generated on the TV every time a remote is operated. Thus, the agent will experience novelty all the time. This keeps the agent's attention high infinitely but clearly leads to no meaningful progress. This kind of behaviour can also be described as a couch potato problem.

\subsubsection{Sparse Reward Problems}

Sparse rewards are a classical problem in exploration. In the sparse reward problem, the reward is relatively rare. In other words, there is a long gap between an action and a reward. This is problematic for reinforcement learning because for a long time (or at all times) it has no reward to learn from. The agent cannot learn any useful behaviours and eventually converges to a trivial solution. As an example, consider a maze where the agent has to complete numerous steps before reaching the end and being rewarded. The larger the maze is, the less likely it is for the agent to see the reward. Eventually, the maze will be so large that the agent will never see the reward; thus, it will have no opportunity to learn.

\subsection{Benchmarks}
In this section, the most commonly used benchmarks for reinforcement learning are briefly introduced and described. We highlight four benchmarks: Atari Games, VizDoom, Minecraft, and Mujoco.

\subsubsection{Atari Games}
The Atari games benchmark are a set of 57 Atari games combined under the Atari Learning Environment (ALE)~\cite{Bellemare2013}. In Atari games, the state space is normally either images or random-access memory (RAM) snapshots. The action space consists of five joystick actions (up, down, left, right, and action button). Atari games can be largely split into two groups: easy (54 games) and difficult exploration (3 games)~\cite{Aytar2018a}. In the easy exploration problem, the reward is relatively easy to find. In hard exploration problems, the reward is not often given, and the association between states and rewards is complex.

\subsubsection{VizDoom}

VizDoom~\cite{Kempka2016} is a benchmark based on the Doom game. The game has a first-person perspective (i.e., view from characters’ eyes), and the image seen by the character is normally used as a state space. The action space is normally eight directional control and two action buttons (picking up key cards and opening doors). Note that more actions can be added, if needed. One of the key advantages of VizDoom is the availability of easy-to-use tools for editing scenarios and low computational burden.

\subsubsection{Malmo}

Malmo~\cite{Johnson2016a} is a benchmark based on the game Minecraft. In Minecraft, environments are built using same-shaped blocks, similar  to how Lego bricks are used for building. Similar to VizDoom, it is also from the first-person perspective, and the image is the state space. The key advantage of Malmo is its flexibility in terms of the environment structure, domain size, custom scripts, and reward functions.

\subsubsection{Mujoco}

MuJoCo~\cite{Todorov2012} represents multi-joint dynamics with contact. Mujoco is a popular benchmark used for physics-based simulations. In reinforcement learning, Mujoco is typically used to simulate walking robots. These are typically cheetah, ant, humanoids, and their derivatives. The task of reinforcement learning is to control various joint angles and forces to develop walking behaviour. Normally, the task is to walk as far as possible or to reach a specific goal.

\section{Exploration in Reinforcement Learning}
\begin{figure*}[htbp]
   \centering
    \includegraphics[width=0.85\textwidth]{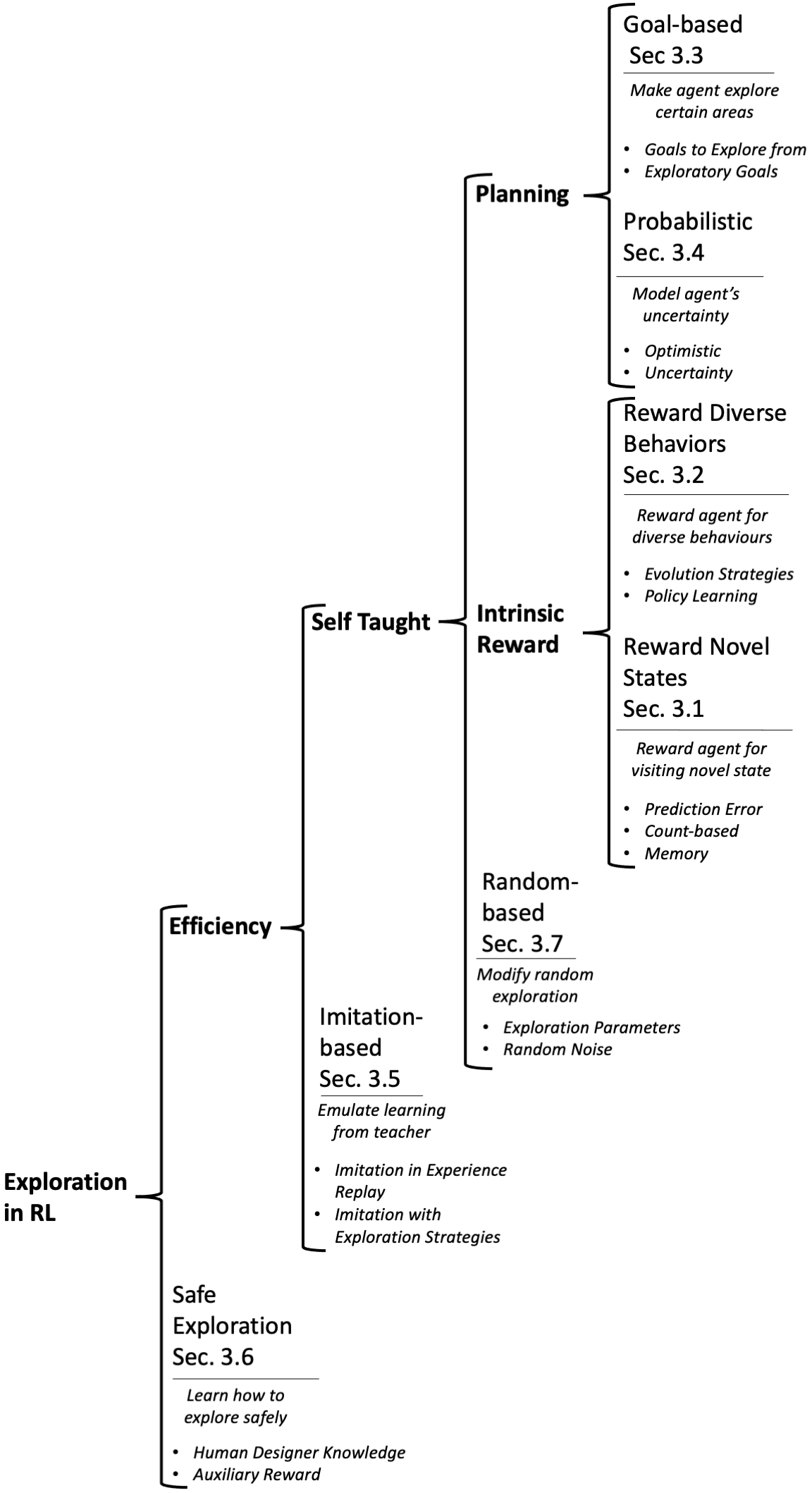}
   \caption{Overview of exploration in reinforcement learning.}
\label{fig:exploration_review_overview}
\end{figure*}

Exploration in reinforcement learning can be split into two main streams: efficiency and safe exploration. In efficiency, the idea is to make exploration more sample efficient so that the agent can explore in as few steps as possible. In safe exploration, the focus is on ensuring safety during exploration. We suggest splitting efficiency-based methods further into imitation-based and self-taught methods. In imitation-based learning, the agent learns how to utilise a policy from an expert to improve exploration. In self-taught methods, learning is performed from scratch. Self-taught methods can be further divided into planning, intrinsic rewards, and random methods. In planning methods, the agent plans its next action to gain a better understanding of the environment. In random methods, the agent does not make conscious plans; rather, it explores and then sees a consequence of this exploration. We distinguish intrinsic reward methods into two categories: (i) reward novel states–reward agents for visiting novel states; and (ii) reward diverse behaviours-reward agents for discovering novel behaviours. Note that intrinsic rewards are a part of a larger notion of intrinsic motivation. For an extensive review  of intrinsic motivation, see \cite{Aubret2019} and \cite{Schmidhuber2010b}. In planning methods, two distinguished categories are considered: (i) goal-based: an agent is given an exploratory goal to reach; and (ii) probability- probabilistic models are used for an environment. Review of the entire categorizations is represented in Fig.~\ref{fig:exploration_review_overview}. From the following, each category is described in detail. The main objective of the categorization is to highlight the key contribution of each approach. Note that a certain approach could be a combination of various techniques. For example, Go-explore \cite{Ecoffet2019} utilizes reward novel states methods, but the main contribution is best described by goal-based methods.

\subsection{Reward Novel States} \label{sec:Intrinsic_Reward}
\begin{figure*}[htbp]
   \centering
    \includegraphics[width=0.85\textwidth]{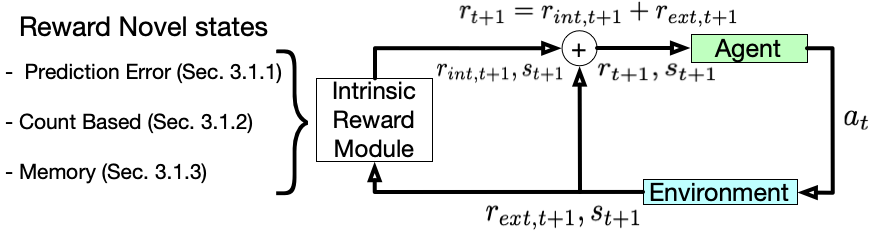}
   \caption{Overview of the reward novel state methods. In general, in reward novel states, the agent is given additional reward $r_{int}$ for discovering novelty. This additional reward is generated from intrinsic reward module $r_{ext}$.}
\label{fig:intrinsic_motivation_overview}
\end{figure*}

In this section, approaches on reward novel state are discussed and compared. Reward novel state approaches give agents a reward for discovering new states. This reward is called an intrinsic reward. As can be observed in Fig.~\ref{fig:intrinsic_motivation_overview}, the intrinsic reward ($r_{int}$) supplements rewards given by the environment ($r_{ext}$ called an extrinsic reward). By rewarding novel states, agents will incorporate exploration into their behaviours \cite{Schmidhuber2010b}.

These approaches were generalised in \cite{Schmidhuber2010b}. In general, there are two necessary components: "an adaptive predictor or compressor or model of the growing data history as the agent is interacting with its environment to provide an intrinsic reward, and a general reinforcement learner to learn behaviours"~\cite{Schmidhuber2010b}. In this division, the reinforcement learner is asked to invent things which predictor does not know yet. In our review, the former is simply referred to as an intrinsic reward module, and the latter is referred to as an agent.

There are different ways of classifying intrinsic rewards \cite{Oudeyer2007b,Aubret2019}. Here, we largely follow the classification of \cite{Aubret2019} with the following categories: (i) prediction error methods, (ii) count-based methods and (iii) memory methods.

\subsubsection{Prediction Error Methods}
In prediction error methods, the error of a prediction model when predicting a previously visited state is used to compute the intrinsic reward. For a certain state, if a model's prediction is inaccurate, it means that a given state has not been seen often and the intrinsic reward is high. One of the key questions that needs to be addressed is how to use the model's error to compute the intrinsic reward. To this end, Achiam et al.~\cite{Achiam2017} compared two intrinsic reward functions: (i) how big the error is in a prediction model and (ii) the learning progress. The first method has shown better performance and is therefore recommended, which can be formalised as: 
\begin{equation}
r_{int}=f(z(s_{t+1})-M(z(s_{t},a_t)))
\end{equation}
where $s$ represents a state, $M$ is an environmental model, $t$ and $t+1$ are two consecutive time steps, $z$ is an optional model for state representation, and $f$ is an optional reward scaling function.

The simplest method of this type was described in \cite{Schmidhuber1991, Schmidhuber1991a}. The intrinsic reward is measured as the Euclidean distance between the prediction of a state from a model and that state. This simple idea was revisited in \cite{Li2019a}. Generative adversarial networks (GAN~\cite{NIPS2014_5ca3e9b1}), distinguishing real from fake states as a prediction error method, were proposed in \cite{Hong2019}. Since then many other approaches were devised which can be further divided into (i) state representation prediction, (ii) a priori knowledge and (iii) uncertainty about the environment.

\paragraph{State representation prediction methods}
In state representation prediction methods, the state is represented in a higher-dimensional space. Then, a model is tasked with predicting the next state representation given the previous state representation. The larger the error is in the prediction, the larger the intrinsic reward is. One way of providing state representation is using an autoencoder \cite{Stadie2015}. Both pre-trained and online trained autoencoders were considered and showed similar performance. Improvements to autoencoder-based approaches were proposed in \cite{Bougie2020c, Bougie2020}, where a slow-trained autoencoder was added. Thus, the intrinsic reward decays slower and the agent explores for longer while increasing the chance of finding the optimal reward.

\begin{figure*}[htbp]
   \centering
    \includegraphics[width=0.8\textwidth]{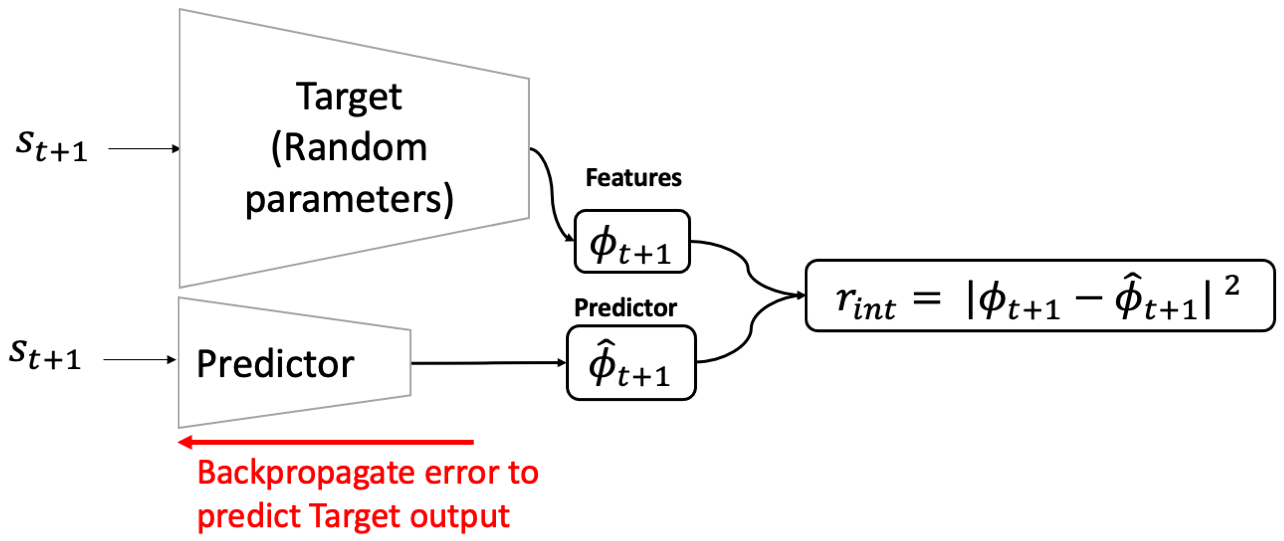}
   \caption{RND overview. The predictor is trying to predict output of a randomly parameterized target.}
\label{fig:RND_overview}
\end{figure*}

Another method of providing state representation involves utilising fixed networks with random weights. Then, another network is used to predict the outputs of randomly initialised networks as shown in Fig. ~\ref{fig:RND_overview}. The most popular approach of this type is called random network distillation (RND)~\cite{Burda2018b}. A similar approach was considered in \cite{Osband2018}.

\begin{figure*}[htbp]
   \centering
    \includegraphics[width=0.8\textwidth]{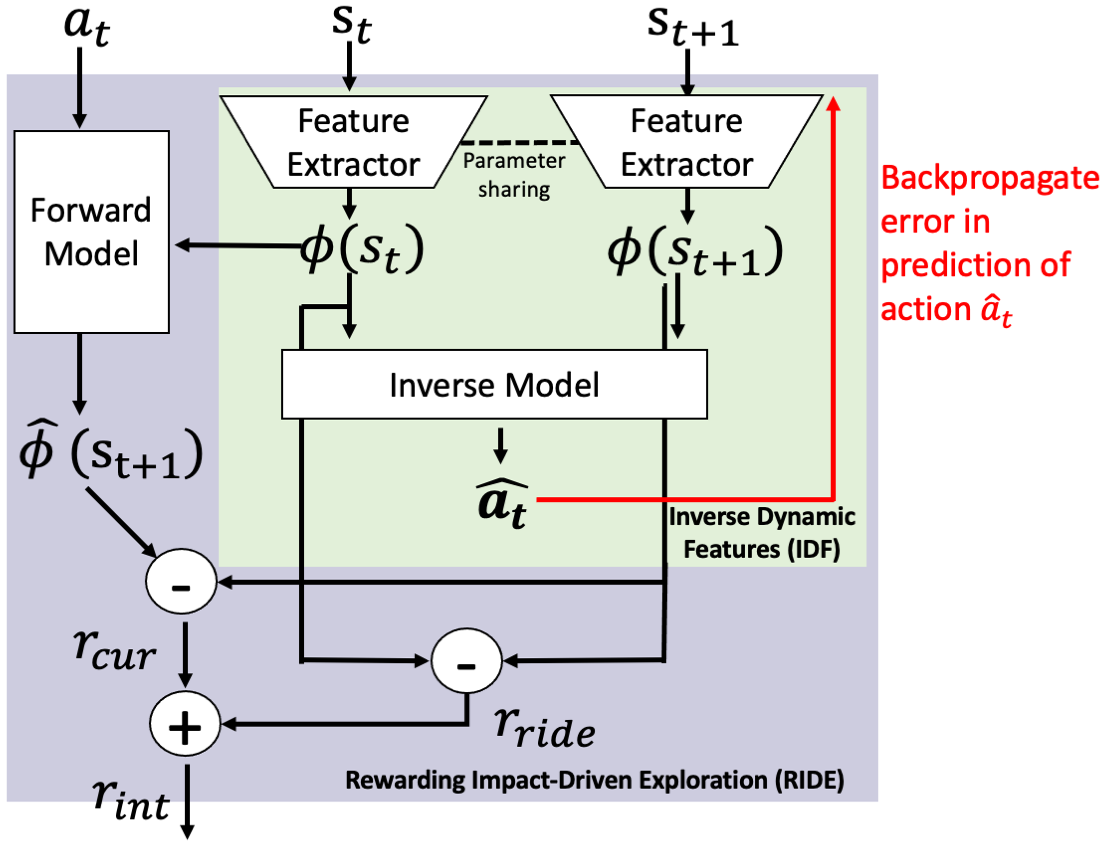}
   \caption{IDF and rewarding impact driven exploration (RIDE) overview. In IDF, the features are extracted based on the network predicting the next action. In RIDE, the intrinsic reward is based on the difference in state representation. (adapted from \cite{Pathak2019})}
\label{fig:RIDE_overview}
\end{figure*}

A state representation method derived from inverse dynamic features (IDF) was used in \cite{Pathak2017b}. In IDF, the representation comes from forcing an agent to predict the action as illustrated in Fig.~\ref{fig:RIDE_overview}. IDF was compared against the state prediction method and random representation in \cite{Burda2018c} with the following conclusions: IDF had the best performance and it scales the best to the unseen environments. IDF was utilised  in \cite{Raileanu2020}, where the Euclidean distance between two consecutive state representations was used as an intrinsic reward, as shown in Fig.~\ref{fig:RIDE_overview}. Intuitively, the more significant the transition is, the larger the change is in IDF's state representation. In another study, RND and IDF were combined into a single intrinsic reward \cite{Li2020a}.

A compact representation using information theory was proposed in \cite{Kim2019c}. Information theory is used to represent states that are close to the representation space in the environment space.
Information theory can also be used to create a bottleneck latent representation \cite{Kim2019b}. Bottleneck latent representation occurs when mutual information between the input to the network and latent representation is minimised.

\paragraph{A priori knowledge methods}
In some types of problems, it makes sense to use certain parts of the state space as an error and use it for computing the intrinsic reward. Those parts could be depth point cloud, position, and sound, and they rely on a priori knowledge from the designer.

Depth point cloud prediction error was used in \cite{Mirowski2017}. The scalability of this approach was analysed by \cite{Dhiman2018a}. It was found that the performance was good in the same environment with different starting positions, but it did not scale to a new scenario. Positions in a 3D space can also be used \cite{Li2020}. An approach using the position was proposed in \cite{Stanton2018}. The environment is split into the x-y grid where each node’s intrinsic reward is placed. When the episode terminates, the rewards are restored to a default value.

Sound as a source of intrinsic reward was used in \cite{Dean2020}. To model sounds, the model is trained to recognise when the sound and the corresponding frame match. If the  model indicates misalignment between frames and sounds, it means that the state is novel.

\paragraph{Uncertainty about the environment methods}
In these methods, the intrinsic reward is based on the uncertainty the agent has. If the agent is exploring highly uncertain areas of the environment, the reward is high. Uncertainty can be utilized using the following techniques Bayesian, ensembles of models and information-theoretic.

\begin{figure*}[htbp]
   \centering
    \includegraphics[width=0.8\textwidth]{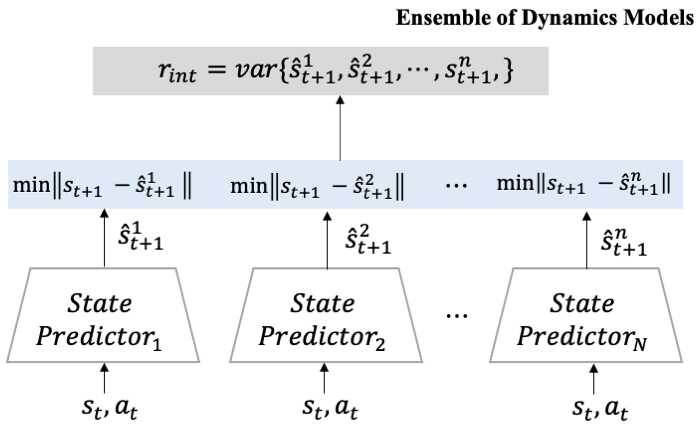}
   \caption{Overview of self-supervised exploration via the disagreement method. The intrinsic reward is based on disagreement between models. (adapted from \cite{Pathak2019})}
\label{fig:Ensambles_overview}
\end{figure*}

Bayesian approaches are generally intractable for large problem spaces; thus, approximations are used. Kotler et al. \cite{Kolter2009a} presented a close to optimal approximation method using the Dirichlet probability distribution over state, action, and next state triplet. Another approximation could be to use ensembles of models as proposed in \cite{Pathak2019}. The intrinsic reward is given based on model disagreement as shown in Fig.~\ref{fig:Ensambles_overview}. The models were initialised with different random weights and were trained on different mini-batches to maintain diversity.

In information-theoretic approaches, the intrinsic reward is computed using the information gained from agent actions. The higher the gain is, the more the agent learns, and the higher the intrinsic reward is. The general framework for these types of approaches was presented in \cite{Still2012a, Still}. One of the most popular information-theoretic approach is called variational information maximization exploration (VIME) \cite{Houthooft2016b}. In this approach, the information gain is approximated as a  Kullback–Leibler (KL) divergence between the weight distribution of the Bayesian neural network, before and after seeing new observations. In \cite{Mohamed2015a}, maximising mutual information between a sequence of actions leads to a state that is rewarded. Rewarding this mutual information gain means maximising the information contained in the action sequence about a state. Mutual information gain was combined with the state prediction error into a single intrinsic reward in \cite{DeAbril2018a, Chien2020}.

\paragraph{Discussion}
The key advantages of prediction error methods are that they rely only on a model of the environment. Thus, there is no need for buffers or complex approximation methods. Each of the four different categories of methods has unique advantages and challenges.

While predicting the state directly requires little to no a priori knowledge, the model needs to learn how to recognise different states. Additionally, they struggle when many states are present in the environment. State representation methods can cope with large state spaces at the cost of increased designer burden and reduced accuracy. Moreover, in a state representation method, the agent cannot affect the state representation, which can often lead to different states being represented similarly. Utilising a priori knowledge relies on defining a special element of the state space as a source of an error for computing the intrinsic reward. These methods do not suffer from problems with the speed of prediction and state recognition. However, they rely on the designer experience to define parts of the state space appropriately. Finally, in uncertainty about the environmental approaches, the agent's uncertainty is used to generate the intrinsic reward. The key advantage of this approach is its high scalability and automatic transition between exploration and exploitation. Prediction error methods have also shown the ability to solve the couch-potato (noisy-TV) problem by storing observations in a memory buffer \cite{Savinov2019}. An intrinsic reward is given only when observation is sufficiently far away (in terms of time steps) from the observations stored in the buffer. This mitigates the couch potato problem since repeatedly visiting states close to each other is not rewarded.

\subsubsection{Count-Based Methods}
In count-based methods, each state is associated with the visitation count number $N(s)$. If the state has a low count, the agent will be given a high intrinsic reward to encourage revisiting. The method of computing the reward based on the count was discussed in \cite{Menard2020a}. It has been shown that 1/$N(s)$ guarantees a faster convergence rate than the commonly used 1/$\sqrt{N(s)}$. 

In problems with large number of states, counting visits to states is difficult because it requires saving the count for each state. To solve this problem, count is normally done on a reduced-size state representation.


\paragraph{Count on state representation methods}
In count on state representation methods, the states are represented in a reduced form to alleviate memory requirements. This allows storing the count and a state with minimal memory in a table, even in the case of a large state space. 

One of the popular methods of this type was proposed in \cite{Tang2017a}, where static hashing was used. Here, a technique called SimHash \cite{Charikar2002} was used, which represents images as a set of numbers called a hash. To generate an even more compact representation, in \cite{Choi2019a}, the state was represented as the learned x-y position of an agent. This was achieved using an attentive dynamic model (ADM). Successor state representation (SSR) \cite{Machado2020a} is a method which combines count and representation. The SSR is based on the count and order between the states. Intuitively, the SSR can be used as a count replacement.

It is also possible to approximate count on state representation by using a function. For example, Bellemare et al. \cite{Bellemare2016} proposed an approximation based on a density model. The density models include context tree switching (CTS) \cite{Bellemare2016}, Gaussian mixture models~\cite{Zhao2019} or PixelCNN~\cite{Ostrovski2017a}. Martin et al. \cite{Martin2017a} proposed an improvement in the approximate count by making counts on the feature space rather than raw inputs.

\paragraph{Discussion}
Count-based methods approximate the intrinsic reward by counting the number of times a given state has been visited. To reduce computational efforts of count-based methods, usually counts are associated with state representations rather than states. This, however, relies on being able to efficiently represent states. State representations can still require a lot of memory and careful design.

\subsubsection{Memory Methods}
In these methods, an intrinsic reward is given on how easy it is to distinguish a state from all others. The easier it is to distinguish from the others, the more novel the given state is. As comparing states directly is computationally expensive, several approximation methods have been devised. Here, we categorize them into comparison models and experience replay.

Models can be trained for comparing state-to-state to reduce the computational load. One example method is to use exemplar model~\cite{Malisiewicz2011} developed in~\cite{Fu2017}. Exemplar models are a set of $n$ classifiers, each of which is trained to distinguish a specific state from the others. Training multiple classifiers is generally computationally expensive. To further reduce the computational cost, the following two strategies are proposed: updating a single exemplar with a each new data point and sampling $k$ exemplars from a buffer. 

Instead of developing models for comparison, a limited size of experience replay was combined with prediction error methods in \cite{Badia2020c}. To devise intrinsic rewards, two rewards are combined: (i) intrinsic episodic experience replay is used to store states and compare them to others; and (ii) intrinsic motivation RND \cite{Burda2018b} is used to determine the state's long-term novelty. Additionally, multiple policies are trained, each with a different ratio between the extrinsic and intrinsic reward. A meta learner to automatically choose different ratios of extrinsic and intrinsic rewards at each step was proposed in \cite{Badia2020b}. 

\paragraph{Discussion}
In memory-based approaches, the agent derives an intrinsic reward by comparing its current state with states stored in the memory. The comparison model method has the advantage of small memory requirements, but requires careful model parameter tuning. On the other hand, using a limited experience buffer does not suffer from model inaccuracies and has shown a great performance in difficult exploratory Atari games. 

\subsubsection{Summary}
The reward novel state-based approaches are summarised in Table ~\ref{tab:novel_state_based}. The table utilizes the following legend:   Legend: A - action space, Ac - action, R - reference, MB - model based, MF - model free, D - discrete, C - continuous, Q - Q values, V - values, P - policy, O - output, S - state space, U - underlying algorithm and Top score on a key benchmark explanation - [benchmark]:[scenario] [score] ([baseline approach] [score]). Prediction error methods are the most commonly-used methods. In general, they have shown very good performance (for example, RND \cite{Burda2018b} with 6,500 in Montezuma's Revenge). However, they normally require a hand-designed state representation method for computational efficiency. This requires problem-specific adaptations, thus reducing the applicability of those approaches. Count-based methods are computationally efficient but they can either require memory to store counts or complex models \cite{Bellemare2016}. Also, counting states in continuous-state domains is challenging and requires combining continuous states into discontinuous chunks. Recently, memory methods have shown good performance in games such as Montezuma Revenge, scoring as much as 11,000 \cite{Badia2020c}. Memory methods require a careful balance of how much data to remember for comparison. Otherwise computing the comparison can take a long time.

\begin{scriptsize}
\begin{ThreePartTable}
\begin{TableNotes}[para, flushleft]
  Legend: A - action space, Ac - action, R - reference, MB - model based, MF - model free, D - discrete, C - continuous, Q - Q values, V - values, P - policy, O - output, S - state space, U - underlying algorithm and Top score on a key benchmark explanation - [benchmark]:[scenario] [score] ([baseline approach] [score]).
\end{TableNotes}
\begin{longtable}   {|P{0.105\linewidth} | P{0.105\linewidth} | P{0.075\linewidth} | P{0.09\linewidth} | P{0.17\linewidth} | P{0.08\linewidth} | P{0.02\linewidth} | P{0.032\linewidth} | P{0.02\linewidth} | }
\caption{Comparison of reward novel state approaches} \label{tab:novel_state_based}\\
    \hline
    R & Prior Knowledge & U & Method  & Top score on a key benchmark & Input Types & O & MB/ MF & A/ S \\ \hhline{|=|=|=|=|=|=|=|=|=|}
\hline
\citet{Pathak2017b} &       & A3C   & Prediction error  & Vizdoom: very sparse 0.6 (A3C 0) & Vizdoom image & P     & MB    & D/ D \\ \hline
\citet{Stadie2015} & autoencoder & DQN   & Prediction error  & Atari: Alien 1436 (DQN 300) & Atari images & Q     & MB    & D/ D \\ \hline
\citet{Savinov2019} & pretrained discrimnator (non-online) & PPO   & Prediction error  & Vizdoom: very sparse  1 (PPO 0); Dmlab: very sparse 30 (PPO+ICM 0) & Vizdoom images/ Mujoco joints angles & Ac    & MB    & C/ C \\ \hline
\citet{Burda2018b} &       & PPO   & Prediction error  & Atari: Montezuma Revenge 7500 (PPO 2500) & Atari images & P     & MB    & D/ D \\ \hline
\citet{Bougie2020c} &       & PPO   & Prediction error  & Atari: Montezuma Revenge 20 rooms found (RND 14) & images & Ac    & MB    & D/ D \\ \hline
\citet{Hong2019} &       & DQN   & Prediction error  & Atari: Montezuma Revenge 200 (DQN 0) & Enumarated state id/ Atari Image & Q     & MB    & D/ D \\ \hline
\citet{Kim2019c} &       & TRPO  & Prediction error  & Atari: Frostbite 6000 (ICM 3000) & Atari Images & Ac    & MB    & D/ D \\ \hline
\citet{Stanton2018} & agent position, reward grid & A2C   & Prediction error  & Atari: Montezuma Revenge 3200 (A2C 0) & Atari images & Ac    & MB    & D/ D \\ \hline
\citet{Achiam2017} &       & TRPO  & Prediction error  & Mujoco: halfcheetah 80 (VIME 40); Atari: Venture 400 (VIME 0) & Atari RAM states/ Mujoco joints angles & Ac    & MB    & C/ C \\ \hline
\citet{Li2020a} &       & A2C   & Prediction error  & Atari: Asterix  500000 (RND 10000) & Atari images & Ac    & MB    & D/ D \\ \hline
\citet{Kim2019b} &       & PPO   & Prediction error  & Atari: Montezuma Revenge with distraction 1500 (RND 0) & Atari images & Ac    & MB    & D/ D \\ \hline
\citet{Chien2020} &       & DQN   & Prediction error  & PyDial: 85 (CME 80); OpenAI: Mario 0.8 (CME 0.8) & Images & Q     & MB    & D/ D \\ \hline
\citet{Li2019a} &       & DDPG  & Prediction error  & Robot: FetchPush 1 (DDPG 0) & Robot joints angles & Ac    & MB    & C/ C \\ \hline
\citet{Raileanu2020} &       & IMPALA & Prediction error  & Vizdoom: 0.95 (ICM 0.95) & Vizdoom Images & Ac    & MB    & D/ D \\ \hline
\citet{Mirowski2017} &       & A3C   & Prediction error  & DM Lab: Random Goal  96 (LSTM-A3C 65) & DM Lab images & Ac    & MB    & C/ C \\ \hline
\citet{Tang2017a} &       & TRPO  & Count-based  & Atari: Montezuma Revenge 238 (TRPO 0); Mujoco: swimmergather 0.3 (VIME 0.15) & Atari images/ Mujoco joints angles & P     & MF    & C/ C \\ \hline
\citet{Martin2017a} & Blob-PROST features & SARSA-e & Count-based  & Atari: Montezuma Revenge 2000 (SARSA 200) & Blob-PROST features & Q     & MB    & D/ D \\ \hline
\citet{Machado2020a} &       & DQN   & Count-based  & Atari: Montezuma Revenge 1396 (Psuedo counts 1671) & Atari images & Q     & MF    & D/ D \\ \hline
\citet{Ostrovski2017a} &       & DQN and Reactor & Count-based  & Atari: Gravitar 1500 (Reactor 1000) & Atari images & Ac    & MB    & D/ D \\ \hline
\citet{Badia2020c} &       & R2D2  & Memory  & Atari: Pitfal 15000 (R2D2 -0.5) & Atari images & P     & MB    & D/ D \\ \hline
\citet{Badia2020b} &       & R2D2  & Memory  & Beat humans in all 57 atari games & Atari images & P     & MB    & D/ D \\ \hline
\citet{Fu2017} & state encoder & TRPO  & Memory  & Mujoco: SparseHalfCheetah 173.2 (VIME 98); Atari: Frostbite 4901 (TRPO 2869);  Doom: MyWayHome 0.788 (VIME 0.443) & Atari images/ Mujoco joints angles & Ac    & MB    & C/ C \\ \hline

\insertTableNotes \\
\end{longtable}%
\end{ThreePartTable}
\end{scriptsize}

\subsection{Reward Diverse Behaviours} \label{sec:Diversity}

\begin{figure*}[htbp]
   \centering
    \includegraphics[width=0.9\textwidth]{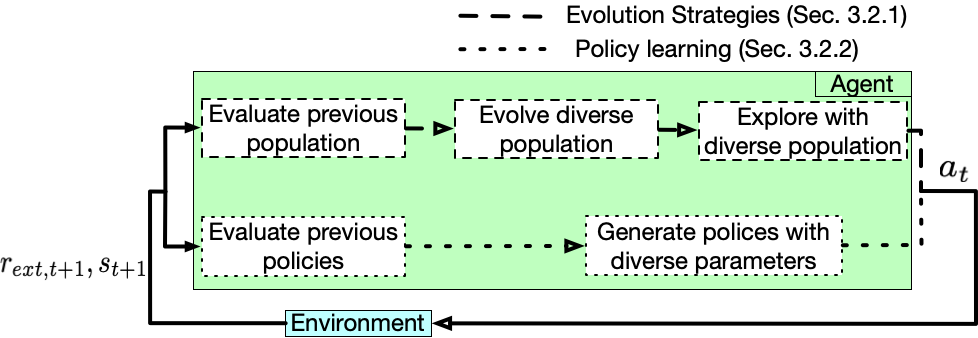}
   \caption{Overview of reward diverse behaviour-based methods. The key idea is for the agent to experience as many things as possible, in which either evolution or policy learning can be used to generate a set of diverse experiences}
\label{fig:diversity_overview}
\end{figure*}

In reward diverse behaviours, the agent collects as many different experiences as possible, as shown in Fig.~\ref{fig:diversity_overview}. This makes exploration an objective rather than a reward finding. These types of approaches can also be called diversity and can be split into evolution strategies and policy learning. 

\subsubsection{Evolution Strategies}
Reward diverse behaviours were initially used with an evolutionary-based approach. In evolutionary approaches, a group of sample solutions (population) is tested and evolves over time to get closer to the optimal solution. Note that evolutionary approaches are generally not considered as the part of reinforcement learning but can be used to solve the same type of problems \cite{Such2017,Salimans2017a}.

One of the earliest methods called novelty search was devised in \cite{Lehman2011} and \cite{Risi2009}. In novelty search, the agent is encouraged to generate numerous different behaviours using a metric called diversity measure. The diversity measure must be hand-designed for each environment, limiting transferability between different domains. Recently, novelty search has been combined with other approaches, such as reward maximization~\cite{Conti2018a} and reward novel state method~\cite{Gravina2018}. In Conti et al. \cite{Conti2018a}, the novelty-search policy is combined with a reward maximisation policy to encourage diverse behaviours and search for the reward. Gravina et al. \cite{Gravina2018} compared three ways of combining novelty search and reward novel state: (i) novelty search, (ii) sum of reward novel state and novelty search, and (iii) sequential optimisation where the second one performed the best in a simulated robot environment. More detailed reviews of exploration in evolution strategies can be found in \cite{Mouret2012} and \cite{Pugh2016}.

\paragraph{Discussion}
Initially, novelty search was used as a stand-alone technique; however, recently, combining it with other techniques~\cite{Gravina2018,Conti2018a} has shown more promise. Such a combination is more beneficial (in terms of reward) as diverse behaviours are more directed toward highly scoring ones.

\subsubsection{Policy Learning}
\begin{figure*}[htbp]
   \centering
    \includegraphics[width=0.9\textwidth]{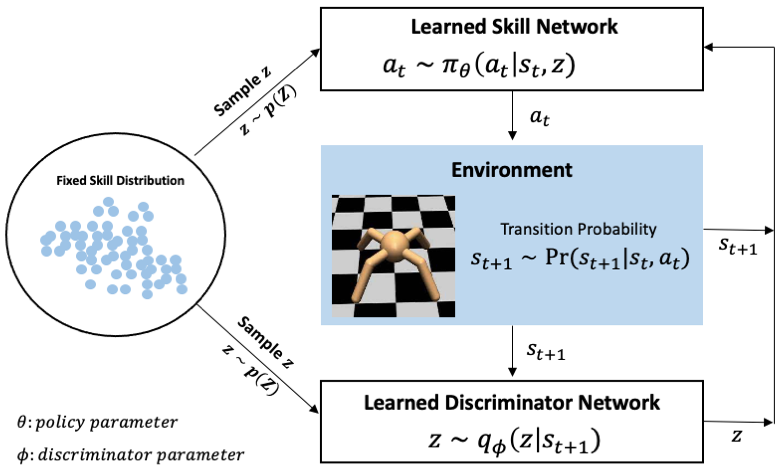}
   \caption{ An overview of  Diversity is all you need (DIAYN), where the agent is encouraged to have as many diverse policies as possible. (adapted from~\cite{Pathak2019})}
\label{fig:DIYAN_overview}
\end{figure*}

Recently, diversity measures have been applied in policy learning approaches. The diversity among policies was measured in \cite{Hong2018}. Diversity is computed by measuring the distance between policies (either KL divergence or simple mean squared error). Very promising results for diversity are presented in \cite{Eysenbach2019a}, as shown in Fig.~\ref{fig:DIYAN_overview}. To generate diverse policies, the objective function consists of (i) maximising the entropy of skills, (ii) inferring behaviours from the current state, and (iii) maximising randomness within a skill. A similar approach was proposed in \cite{Cohen2019} with a new entropy-based objective function. A combination of diversity with a goal-based approach was proposed in \cite{Pong2019}. In this study, the agent learns diverse goals and goals useful for rewards using the skew-fit algorithm. In the skew-fit algorithm, the agent skews the empirical goal distribution so that rarely visited states can be visited more frequently. The algorithm was tested using both simulations and real robots.

In \cite{Hess2019}, the agent stores a set of successful policies in an experience replay and then minimises the difference between the current policy and the best policies from storage. To allow exploration at the same time, the entropy of parameters between policies is maximised. The results show an advantage over evolution strategies and PPO in sparse reward Mujoco problems. 

\paragraph{Discussion}
Diversity in policy-based approaches is a relatively new concept that is still being developed. Careful design of a diversity criterion shows very promising performance, beating standard reinforcement learning with significant margins~\cite{Eysenbach2019a}.

\subsubsection{Summary}
Reward diverse behaviour methods are summarised in Table ~\ref{tab:diversity_based}. In evolution strategies approaches, a diverse population is used, whereas in policy learning, a diverse policy is found. Evolution strategies have the potential to find solutions that are not envisioned by designers as they search for the neural network structure as well as diversity. Evolution strategies suffer from the low sample efficiency, making the training either computationally expensive or slow. Policy learning is not able to go beyond pre-specified structures but can also show some remarkable results \cite{Eysenbach2019a}. Another advantage of the policy learning method is suitability to both continuous and discrete state-space problems.

\begin{scriptsize}
\begin{ThreePartTable}
\begin{TableNotes}[para, flushleft]
  Legend: A - action space, Ac - action, R - reference, MB - model based, MF - model free, D - discrete, C - continuous, Q - Q values, V - values, P - policy, O - output, S - state space, U - underlying algorithm and Top score on a key benchmark explanation - [benchmark]:[scenario] [score] ([baseline approach] [score]).
\end{TableNotes}
\begin{longtable}   {|P{0.085\linewidth} | P{0.105\linewidth} | P{0.075\linewidth} | P{0.09\linewidth} | P{0.17\linewidth} | P{0.1\linewidth} | P{0.02\linewidth} | P{0.032\linewidth} | P{0.02\linewidth} | }
\caption{Comparison of reward diverse behaviour-based approaches} \label{tab:diversity_based}\\ 
    \hline
    R & Prior Knowledge & U & Method  & Top score on a key benchmark & Input Types & O & MB/ MF & A/ S \\ \hhline{|=|=|=|=|=|=|=|=|=|}
\hline
\citet{Conti2018a} & domain specific behaviours & Reinforce & Evolution strategies & Atari: Frostbite 3785 (DQN 1000) & Atari RAM state/ Mujoco joints angles & Ac    & MF    & C/ C \\ \hline
\citet{Gravina2018} &       & NS population based & Evolution strategies & Robotic navigation: 400 successes & six range finders, pie-slice goal-direction sensor & Ac    & MB    & C/ C \\ \hline
\citet{Lehman2011} & measure of policies distance & NEAT  & Evolution strategies & maze: 295 (maximum achievable) & six range finders, pie-slice goal-direction sensor & Ac    & MF    & D/ D \\ \hline
\citet{Risi2009} & measure of policies distance & NEAT  & Evolution strategies & T-maze: solved after 50,000 evaluations & enumarated state id & Ac    & MF    & D/ D \\ \hline
\citet{Cohen2019} &       & SAC   & Policy learning & Mujoco: Hopper 3155 (DIAYN 3120) & Mujoco joint angles & Ac    & MB    & C/ C \\ \hline
\citet{Pong2019} &       & RIG   & Policy learning & Door Opening (distance to the objective): 0.02 (RIG + DISCERN-g 0.04) & Robots joint angles & Ac    & MB    & C/ C \\ \hline
\citet{Eysenbach2019a} &       & SAC   & Policy learning & Mujoco: half cheetah 4.5 (TRPO 2) &  Mujoco joints angles & Ac    & MF    & C/ C \\ \hline
\citet{Hong2018} &       & DQN, DDPG, A2C & Policy learning & Atari: Venture 900 (others 0); Mujoco: SparseHalfCheetah  80 (Noisy-DDPG 5) & Atari images/ Mujoco joints angles & Ac/ Q & MB    & C/ C \\ \hline
\citet{Hess2019} &       & Itself & Policy learning & Mujoco: SparseHalfCheetah 1000 (PPO 0) & Robot joints angles & Ac    & MF    & C/ C \\ \hline

\insertTableNotes \\
\end{longtable}%
\end{ThreePartTable}
\end{scriptsize}

\subsection{Goal-Based Methods} \label{sec:Goal_Based}
\begin{figure*}[htbp]
   \centering
    \includegraphics[width=0.9\textwidth]{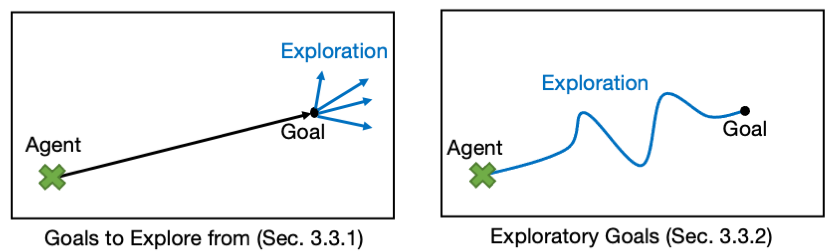}
   \caption{Illustration of goal-based methods. In goal-based methods, the agent's task is to reach a specific goal (or state). Then, this goal is explored using another exploration method (left) or to generate an exploratory target goal (right). The key concept is to guide agents directly to unexplored areas.}
\label{fig:goal_based_overview}
\end{figure*}

In goal-based methods, the states of interest for exploration are used to guide the agent's exploration. In this way, the exploration can immediately focus on largely unknown areas. In those types of methods, the agent requires a goal generator, a policy to find a goal, and an exploration strategy (see Fig.~\ref{fig:goal_based_overview}). The goal generator is responsible for creating goals for the agent. The policy is used to achieve the desired goals. An exploration strategy is used to explore once a goal has been achieved or while trying to achieve goals.

Here, we split goal-based methods into two categories: goals to explore from and exploratory goal methods.

\subsubsection{Goals to Explore from Methods} \label{sec:Generate_Goals_to_Explore_from}
The main technique used for these methods are (i) memorize visited states and trajectories - storing the past states in a buffer and choosing an exploratory goal from the buffer; and (ii) learn from the goal - assuming the goal state is known but a path to it is unknown.

One of the most famous approach when goal is chosen from a buffer of this type is called the go-explore \cite{Ecoffet2019}. The states and trajectories are saved in a buffer and are selected probabilistically. Once the state to explore from has been found, the agent is teleported there and explores it randomly. In \cite{Ecoffet2020}, teleportation was replaced with policy learning. Go-exploration was extended to continuous domains in \cite{Matheron2020a}. Concurrently, similar concepts were developed  in \cite{Guo2019, Guo2019a, Oh2018, Guo2020}. In these approaches, a trajectory from the past is selected as an agent to exploit or explore randomly. If exploration is selected, a sample state from the trajectory is selected as a goal to explore based on the visitation count.

Another goal method was proposed in \cite{Liu2020b}, where the least visited state was selected as a goal from the x-y grid on an Atari game. This reduces the computational effort of remembering where the agent has been significantly.

Learn from goal methods assume that the agent knows how the state with maximum reward looks like, but does not know how to get there. In this case, it is plausible to utilise this knowledge, as described in \cite{Edwards2018, Florensa2017a}. In \cite{Edwards2018}, the model was trained to predict the backward steps in reinforcement learning. With such a model, the agent can 'imagine' states before the goal and thus can explore from the goal state. Similarly, another scenario, in which the agent can start at the reward position, can be conceived; then, it can also explore the starting position from the goal \cite{Florensa2017a}.

\paragraph{Discussion}
Memorise visited states and trajectories methods have shown some remarkable results in hard exploration benchmarks such as Montezuma's revenge and pitfall. By utilising a reward state as a goal, as outlined in learn from the goal methods, the exploration problem can be mitigated, as the agent knows where to look for the reward.

\subsubsection{Exploratory Goal Methods} \label{sec: Generate_Goals_to_Explore_to}
In this subsection, an exploratory goal is given to the agent to try to reach. Exploration occurs when the agent attempts to reach the goal. The following techniques are considered: (i) meta-controllers, (ii) goals in the region of the highest uncertainty, and (iii) sub-goal methods.

\paragraph{Meta-controllers}
In meta-controllers, the algorithm consists of two parts: a controller and a worker. The controller has a high-level overview and provides goals that the worker is trying to find.

One of the simple approaches is to generate and sample goals randomly \cite{Forestier2017}. The random goal selection mechanism was refined in \cite{Colas2019} with goal selection based on the learning progress. A similar approach in two phases was proposed by Pere et al.~\cite{Pere2018}. First, the agent explores randomly to learn the environment representation. Second, the goals are randomly sampled from the learned representation. An approach in which both goal creation and selection mechanisms are devised by a meta-controller was proposed in \cite{Vezhnevets2017a}. In this work, a meta-controller proposes goals within a certain horizon for a worker to find.

In \cite{Hester2013a}, a multi-arm bandit-based method to choose one strategy from a group of hand-designed strategies was proposed. At each episode, every ten steps, the agent chooses a strategy based on its performance. The goal selection mechanism from a group of hand-designed goals is also discussed in \citet{Kulkarni2016a}. The low-level controller is trained on a state-action space, and the meta-controller is trained on a goal-state space. An approach in which each subtask is learned by one learner was proposed in \cite{Riedmiller2018a}. To allow any sub-task learner to perform its task from all states, the starting points for learning are shared between sub-task learners.

\paragraph{Sub-goals}
In sub-goal methods, the algorithms find the sub-goals for the agent to reach. In general, sub-goal methods can be split into: (i) bottleneck states which lead to many others as exploratory goals, (ii) progress towards the main goal which is likely to lead to the reward and (iii) uncertainty based sub-goals.

One of the early methods of discovering bottleneck states was described in \cite{Ghafoorian2013a} using an ant colony optimisation method. Bottleneck states are said to be the states often visited by ants when exploring (by measuring pheromone levels). To discover bottleneck states, \cite{Machado2017} proposed the use of proto-value functions based on the eigenvalue of representations. This allows the computation of eigenvector centrality~\cite{zaki_meira}, which has a high value if the node has many connections. This was later improved in \cite{Machado2018} by replacing the handcrafted adjacency matrix with successor representations.

To design sub-goals which lead to a reward, Fang et al.~\cite{Fang2020} proposed progressively generating sub-tasks that are closer to the main task. To this end, two components are used: the learning progress estimator and task generator. The learning progress estimator determines the learning progress on the main task. The task generator then uses the learning progress to generate sub-tasks closer to the main tasks. 

In uncertainty based methods, sub-goals the goals for the agent are positioned at the most uncertain states. One of the earliest attempts of this type was proposed by \citet{Guestrin2002a}. Here, the upper and lower bounds of the reward are estimated. Then, states with high uncertainty regarding the reward are used as exploratory goals. Clustering states using k-means and visiting least-visited clusters were proposed in \cite{Abel2016a}. Clustering can also help to solve the couch potato problem, as described in \cite{Kovac2020}. In this approach, the states are clustered using Gaussian mixture models. The agent avoids the couch potato problem by clustering all states from a TV into a single avoidable cluster.

\paragraph{Discussion}
There are two main categories of exploratory goal methods: meta-controllers, and sub-goals. The key advantage of meta-controllers is that they allow the agent to set its own goals without excessively rewarding itself. However, training the controller is a challenge, which was not fully solved yet. In sub-goals methods, what constitutes a goal is defined by human designers. This puts a significant burden on the designer to provide suitable and meaningful goals.

\subsubsection{Summary}
The goal-based methods are summarised in Table~\ref{tab:goal_based}. Goals to explore from methods have shown very good performance recently~\cite{Ecoffet2020, Guo2020} in difficult exploratory games such as Montezuma's Revenge. The key challenges of these methods are the need to store states and trajectories as well as how to navigate to the goal. This issue is partially mitigated in \cite{Guo2020} by using the agent's position as the state representation, however, this is highly problem-specific. Exploratory goal methods are limited as devising an exploratory goal becomes more challenging with increasing sparsity of the reward. This is somewhat mitigated in \cite{Colas2019} or \cite{Fang2020}, but these approaches rely on the ability to parametrize the task.

\begin{scriptsize}
\begin{ThreePartTable}
\begin{TableNotes}[para, flushleft]
  Legend: A - action space, Ac - action, R - reference, MB - model based, MF - model free, D - discrete, C - continuous, Q - Q values, V - values, P - policy, O - output, S - state space, U - underlying algorithm and Top score on a key benchmark explanation - [benchmark]:[scenario] [score] ([baseline approach] [score]).
\end{TableNotes}
\begin{longtable}   {|P{0.085\linewidth} | P{0.105\linewidth} | P{0.075\linewidth} | P{0.09\linewidth} | P{0.17\linewidth} | P{0.1\linewidth} | P{0.02\linewidth} | P{0.032\linewidth} | P{0.02\linewidth} | }
\caption{Comparison of Goal-based approaches} \label{tab:goal_based}\\    \hline
    R & Prior Knowledge & U & Method  & Top score on a key benchmark & Input Types & O & MB/ MF & A/ S \\ \hhline{|=|=|=|=|=|=|=|=|=|}
\hline
\citet{Guo2019a} &       & A2C and PPO & Goals to Explore from  & Atari: Montezume Revenage 20158 (A2C+CoEX 6600) & Atari images & P     & MF    & C/ C \\ \hline
\citet{Guo2019} &       & DQN, DDPG & Goals to Explore from  & Mujoco: Fetch Push 0.9 after 400 epoch (DDPG 0.5) &  Mujoco joints angles & Q     & MB    & C/ C \\ \hline
\citet{Florensa2017a} & goal position & TRPO  & Goals to Explore from  & Mujoco: Key Insertion  0.55 (TRPO 0.01) &  Mujoco joints angles & Ac    & MF    & C/ C \\ \hline
\citet{Edwards2018} & goal state information & DDQN  & Goals to Explore from  & Gridworld 0 (DDQN -1) & Enumarated state id & Q     & MF    & D/ D \\ \hline
\citet{Matheron2020a} & state storage method & DDPG  & Goals to Explore from  & Maze: reach reward after 146k (TD3 never) & x-y position & Ac    & MB    & C/ C \\ \hline
\citet{Oh2018} &       & A2C   & Goals to Explore from  & Atari: Montezuma Revenge 2500 (A2C 0) & Atari images & Ac    & MF    & D/ D \\ \hline
\citet{Guo2020} & access to agent position & Itself & Goals to Explore from  & Atari: Pitfall 11,000 (PPO 0); Robot manipulation task: 40 (PPO 0) & Atari images/ agent positions/ robotics joint angles & Ac    & MF    & C/ C \\ \hline
\citet{Ecoffet2019} & teleportation ability & itself & Goals to Explore from  & Atari: Montezuma Revenge 46000 (RND 11000) & Atari images & Ac    & MB    & D/ D \\ \hline
\citet{Ecoffet2020} & access to agent position & itself & Goals to Explore from  & Atari: Montezuma Revenge 46000 (RND 11000) & Atari images & Ac    & MB    & D/ D \\ \hline
\citet{Hester2013a} & Strategies set & texpl- ore-vanir & Exploratory Goal  &  Sensor Goal: -53 (greedy -54) & Enumarated state id & Ac    & MB    & D/ D \\ \hline
\citet{Machado2017} & hadcrafted features & itself & Exploratory Goal  & 4-room domain: 1 & Enumarated state id & Ac    & MB    & D/ D \\ \hline
\citet{Machado2018} &       & itself & Exploratory Goal  & 4-room domain: 1 & Enumarated state id & Ac    & MB    & D/ D \\ \hline
\citet{Abel2016a} &       & DQN   & Exploratory Goal  & Malmo: Visual Hill Climbing 170 (DQN+boosted 60) & Image/ Vehicle positions & Q     & MB    & C/ C \\ \hline
\citet{Forestier2017} & randomly generated goals & Itself & Exploratory Goal  & Minecraft: mountain car 84\% explored ($\epsilon$-greedy 3\%) & State Id & Ac    & MB    & C/ C \\ \hline
\citet{Colas2019} &       & M-UVFA & Exploratory Goal  & OpenAI: Goal Fetch Arm 0.8 (M-UVFA 0.78) & Robot joints angles & Ac    & MB    & C/ C \\ \hline
\citet{Pere2018} &       & IMGEP & Exploratory Goal  & Mujoco (KLC): ArmArrow 7.4 (IMGEP with handcrafted features 7.7) &  Mujoco joints angles & Ac    & MF    & C/ C \\ \hline
\citet{Ghafoorian2013a} &       & Q-learning & Exploratory Goal  & Taxi Driver: Found goal after 50 episodes (Q-learning after 200) & State Id & Q     & MF    & D/ D \\ \hline
\citet{Riedmiller2018a} & rewards for axuillary tasks & Itself & Exploratory Goal  & Block stacking: 140 (DDPG 0) & Robot joints angles & Ac    & MB    & C/ C \\ \hline
\citet{Fang2020} & tasks parameterization & itself & Exploratory Goal  & GirdWorld: 1 (GoalGAN 0.6) & Robot joints angles, images & Ac    & MB    & C/ C \\ \hline

\insertTableNotes \\
\end{longtable}%
\end{ThreePartTable}
\end{scriptsize}

\subsection{Probabilistic Methods} \label{sec:Uncertainty}

\begin{figure*}[htbp]
   \centering
    \includegraphics[width=0.99\textwidth]{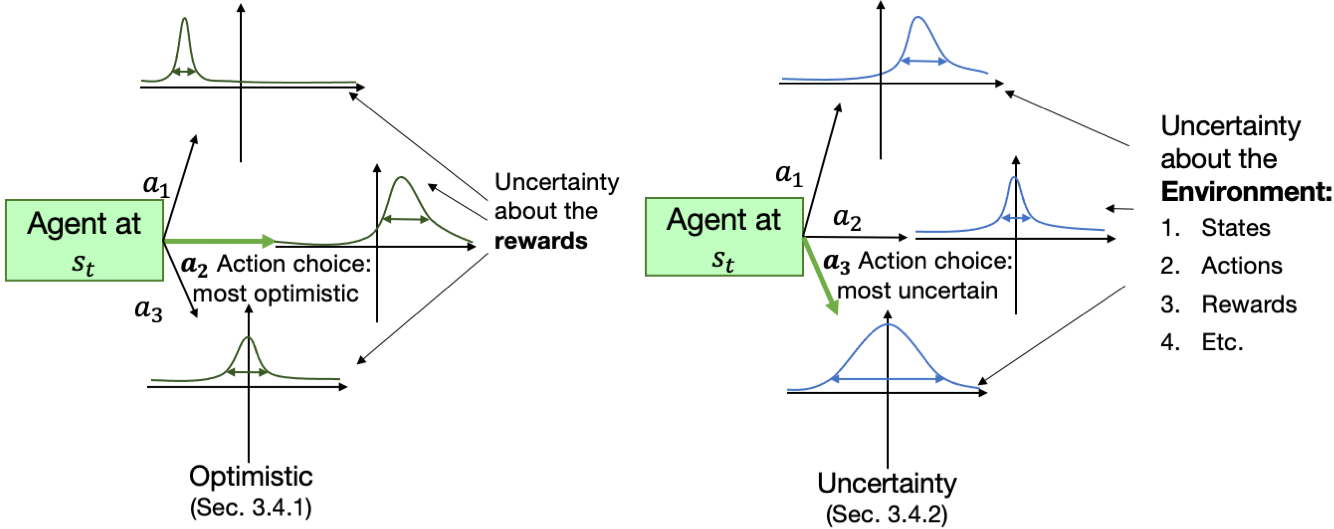}
   \caption{Overview of probabilistic methods. The agent uses uncertainty over the environment model to either behave optimistically (left) or follow the most uncertain solution (right). Both should lead to a reduction in the uncertainty of the agent.}
\label{fig:uncertainty_overview}
\end{figure*}

In probabilistic approaches, the agent holds a probability over states, actions, values, rewards or their combination and chooses the next action based on that probability. Probabilistic methods can be split into optimistic and uncertain methods \cite{Osband2017a}. The main difference between them is how they model a probability and how the agent utilises the probability, as shown in Fig. \ref{fig:uncertainty_overview}. In optimistic methods, the estimation needs to depend on a reward, either implicitly or explicitly. Then, the upper bound of the estimate is used to make the action. In uncertainty-based methods, the estimate is the uncertainty about the environment, such as the value function and state prediction. In the uncertainty-based method, the agent takes actions that minimise environmental uncertainty. Note that uncertainty methods can use estimations from optimistic methods but they utilise them differently.

\subsubsection{Optimistic Methods}
In optimistic approaches, the agent follows \emph{optimism under the uncertainty} principle. In other words, the agent follows the upper confidence bound of the reward estimate. The use of Gaussian process (GP) as a reward model was presented in \cite{Jung2010}. The GP readily provides uncertainty, which can be used for reward estimation. The linear Gaussian algorithm can also be used as a model of the reward \cite{Xie2016}. Bootstrapped deep-Q networks (DQN) and Thomson sampling  were utilised in \cite{DEramo2019}. Bootstrapped DQNs naturally provide a distribution over rewards and values so that optimistic decisions can be taken.

It is also possible to hold a set of value functions and samples during exploration \cite{Osband2016a, Osband2019b}. The most optimistic value function is used by the agent for an episode. At the end of the episode, the distribution of the value functions was updated.

\paragraph{Discussion}
In optimistic approaches, the agent attempts to utilise \emph{optimism under the uncertainty} principle. To utilize this principle the agent needs to be able to model the reward. It is possible to do this modeling by either modelling reward directly or by approximating value functions. Value function approximation can be advantageous as reward sparsity increases. With increased reward sparsity, the agent can utilize the partial reward from value functions for learning.

\subsubsection{Uncertainty Methods}
In uncertainty-based methods, the agent holds a probability distribution over actions and/or states which represent the uncertainty of the environment. Then, it chooses an action that minimises the uncertainty. Here, five subcategories are distinguished: parameter uncertainty, value uncertainty, network ensemble, and information-theoretic.

\paragraph{Parameter uncertainty}
In parameter uncertainty, the agent holds uncertainty over the parameters defining a policy. Then, the agent samples from those and follows this policy for a certain time and  updates the parameters based on the performance. One of the simplest approaches is to hold a distribution over the parameters of the network~\cite{Tang2018}. Here, the network parameters were sampled from the weight distribution. Colas et al. \cite{Colas2018} split the exploration into two phases: (i) explore randomly and (ii) compare experiences to an expert-created imitation to determine the good behavior.

In \cite{Janz2019}, the successor state representation was utilised as a model of the environment. The exploration was performed by sampling parameters from the Bayesian linear regression model which predicts successor representation. 

\paragraph{Policy and Q-value uncertainty}
In policy and Q-value uncertainty, the agent holds uncertainty over Q-values/actions and samples the appropriate action. Some of the simplest approaches rely on optimisation to determine the distribution parameters. For example, in \cite{Stulp2012}, the cross-entropy method (CEM) was used to control the variance of a Gaussian distribution from which actions were drawn. Alternatively, policies can be sampled~\cite{Akiyama2010}. In this study, a set of sampling policies sampled from a base policy were used. At the end of the episode, the best policy was chosen as an update to the base policy.

The most prevalent approach of this type is to use the Bayesian framework. In \cite{Strens2000a}, the hypothesis is generated once and then followed for a certain number of steps, which saves computational time. This idea was further developed in \cite{Guez2012}, where Bayesian sampling was combined with a tree-based state representation for further efficiency gains. To enable Bayesian uncertainty approaches to deep learning, \citet{ODonoghue2018a} derived Bayesian uncertainty such that it can be computed using the Bellman principle and the output of the neural network. 

To minimize the uncertainty about policy and/or Q-values, information-theoretic approaches can be used. Agents choose actions that will result in maximal information gain, thus reducing uncertainty about the environment. An example of this approach, called information-directed sampling (IDS), is discussed in \cite{Nikolov2019}. In IDS, the information gain function is expressed as a ratio between regret and how informative the action is.

\paragraph{Network ensembles}
In the network ensemble method, the agent uses several models (initialised with different parameters) to approximate the distribution. Sampling one model from the ensemble to follow was discussed in \cite{Osband2016}. In this study, a DQN with multiple heads, each estimating Q-value, was proposed. At each episode, one head was chosen randomly for use.

It is difficult to determine the model convergence by sampling one model at a time. Therefore, multiple models to approximate the distribution over states were devised in \cite{Pearce2018}. In this approach, Q-values estimated by different models were computed and fitted into a Gaussian distribution. A similar approach was developed in \cite{Shyam2019a}, using the information gain among the environmental models to decide where to go. Another ensemble model was presented in \cite{Henaff2019a}. Exploration is achieved by finding a policy which results in the highest disagreement among the environmental models.

\paragraph{Discussion}
In parameter sampling, the policy is parameterized (i.e. represented by the neural network), and the probability over parameters is devised. The agent samples the parameters and continues the update-exploitation cycle. In contrast, in policy and Q-value sampling methods, the probability distribution is not based on policy parameters but on actions and Q-values. The advantage of doing this over parameter sampling is faster updates because the policy can be adjusted dynamically. The disadvantage is that estimating the exact probability is intractable, and thus, simplifications need to be made. Another method is to use network ensembles to approximate the distribution over the action/states. This agent can either sample from the distribution or choose one model to follow. While more computationally intensive, this approach can also be updated instantaneously.

\subsubsection{Summary}
Tabular summary of optimistic and uncertainty approaches is shown in Table~\ref{tab:uncertainty_based} and have been extensively compared in \cite{Osband2017a}. The article concludes that the biggest issue for optimistic exploration is that the confidence sets are built independent of each other. Thus, an agent can have multiple states with high confidence. This results in unnecessary exploration as the agent visits states which do not lead to the reward. Remedying this issue would be computationally intractable. In uncertainty methods, the confidence bounds are built depending on each other; thus, it does not have this problem.

\begin{scriptsize}
\begin{ThreePartTable}
\begin{TableNotes}[para, flushleft]
  Legend: A - action space, Ac - action, R - reference, MB - model based, MF - model free, D - discrete, C - continuous, Q - Q values, V - values, P - policy, O - output, S - state space, U - underlying algorithm and Top score on a key benchmark explanation - [benchmark]:[scenario] [score] ([baseline approach] [score]).
\end{TableNotes}
\begin{longtable}   {|P{0.11\linewidth} | P{0.105\linewidth} | P{0.075\linewidth} | P{0.09\linewidth} | P{0.17\linewidth} | P{0.08\linewidth} | P{0.02\linewidth} | P{0.032\linewidth} | P{0.02\linewidth} | }
\caption{Comparison of probabilistic approaches} \label{tab:uncertainty_based}\\  \hline
    R & Prior Knowledge & U & Method  & Top score on a key benchmark & Input Types & O & MB/ MF & A/ S \\ \hhline{|=|=|=|=|=|=|=|=|=|}
\hline
\citet{DEramo2019} &       & bDQN, SARSA & Optimistic  & Mujoco: acrobot -100 (Thomson -120) &  Mujoco joints angles & Q     & MF    & C/ C \\ \hline
\citet{Osband2016a} &       & LSVI  & Optimistic  & Tetris: 5000 (LSVI 4000) & Hand tuned 22 features & Ac    & MF    & D/ D \\ \hline
\citet{Jung2010} &       &       & Optimistic  & Mujoco:  Inverted Pendulum 0 (SARSA -10) & State Id & Ac    & MB    & D/ C \\ \hline
\citet{Xie2016} &       & MPC   & Optimistic  & Robotics hand simulation: complete each of 10 poses & joints angles & Ac    & MB    & C/ C \\ \hline
\citet{Osband2019b} &       & LSVI  & Optimistic  & Cartpole Swing up: 600 (DQN 0) & State Id & Ac    & MF    & D/ D \\ \hline
\citet{Nikolov2019} &       & bDQN and C51 & Uncertainty  & 55 atari games: 1058\% of reference human performance & Atari images & Q     & MB    & D/ D \\ \hline
\citet{Colas2018} & a set of goal policies $O$ & DDPG  & Uncertainty  & Mujoco: Half Cheetah 6000 (DDP 5445) & Mujoco joints angles & P     & MF    & C/ C \\ \hline
\citet{Osband2016} &       & DQN   & Uncertainty  & Atari: James Bond 1000 (DQN 600) & Atari imgaes & Q     & MB    & D/ D \\ \hline
\citet{Tang2018} &       & DDPG  & Uncertainty  & Mujoco: sparse mountaincar 0.2 (NoisyNet 0) &  Mujoco joints angles & Ac    & MF    & D/ C \\ \hline
\citet{Strens2000a} &       & Dynamic Programming & Uncertainty  & Maze: 1864 (QL SEMI-UNIFORM 1147) & Enumarated state id & Ac    & MB    & D/ D \\ \hline
\citet{Akiyama2010} & initial policy guess & LSPI  & Uncertainty  & Ball bating 2-DoF simulation: 67 (Passive learning:61) & Robot angles & Ac    & MB    & D/ C \\ \hline
\citet{Henaff2019a} &       & DQN   & Uncertainty  & Maze: -4 (UE2 -14) & Enumarated state id & Q     & MB    & D/ D \\ \hline
\citet{Guez2012} & guess of a prior & Policy learning & Uncertainty  & Dearden Maze: 965.2 (SBOSS 671.3) & Enumarated state id & Ac    & MB    & D/ D \\ \hline
\citet{Pearce2018} & guess of a prior & DQN   & Uncertainty  & Cart pole: 200 & Enumarated state id & Q     & MB    & C/ C \\ \hline
\citet{ODonoghue2018a} & prior distribution & DQN   & Uncertainty  & Atari: Montezuma Revenge 3000 (DQN 0) & Atari images & Q     & MB    & D/ D \\ \hline
\citet{Shyam2019a} &       & SAC   & Uncertainty  & Chain: 100\% explored (bootstrapped-DQN  30\%) & Enumarated state id/ Mujoco joints angles & Ac    & MB    & C/ C \\ \hline
\citet{Stulp2012} &       & $PI^2$ & Uncertainty  & Ball batting: learned after 20 steps & Robot joints angles & Ac    & MB    & C/ C \\ \hline
\citet{Janz2019} &       & DQN   & Uncertainty  & 49 Atari games: 77.55\% superhuman (Bootstrapped DQN  67.35\%) & Atari images & Q     & MB    & D/ D \\ \hline

\insertTableNotes \\
\end{longtable}%
\end{ThreePartTable}
\end{scriptsize}

\subsection{Imitation-Based Methods} \label{sec:Imitation_based}

\begin{figure*}[htbp]
   \centering
    \includegraphics[width=0.9\textwidth]{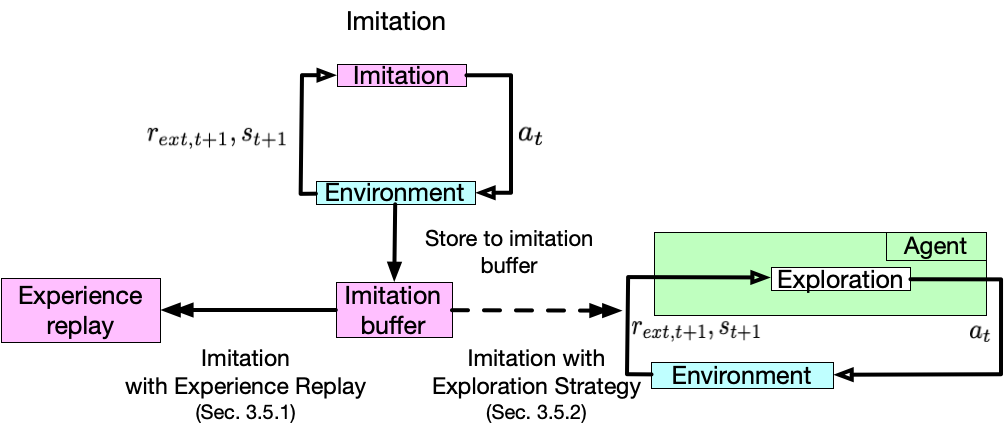}
   \caption{Overview of imitation-based methods. In imitation-based methods, the agent receives demonstrations from expert on how to behave. These are then used in two ways: (i) directly learning on demonstrations or (ii) using demonstrations as a start for other exploration techniques.}
\label{fig:imitation_learning_overview}
\end{figure*}

In imitation learning, the exploration is 'kick-started' with demonstrations from different sources (usually humans). This is similar to how humans learn because we are initially guided in what to do by society and teachers. Thus, it is plausible to see imitation learning as a supplement to standard reinforcement learning. Note that demonstrations do not have to be perfect; rather, they just need to be a good starting point. Imitation learning can be categorized to imitation in experience replay and imitation with exploration strategy as illustrated in  Fig.~\ref{fig:imitation_learning_overview}.

\subsubsection{Imitations in Experience Replay Methods}
One of the most common techniques is combining samples from demonstrations with samples collected by an agent in a single experience replay. This guarantees that imitations can be used throughout the learning process while using new experiences.

In \cite{Vecerik2017}, the demonstrations were stored in a prioritised experience replay alongside the agent's experience. The transitions from demonstrations have a higher probability of being selected. Deep Q learning from demonstration (DQfD)~\cite{Hester2018a} differs in two aspects from \cite{Vecerik2017}. First, the agent was pre-trained on demonstrations only. Second, the ratio between the samples from the agent's run and demonstrations was controlled by a parameter. A similar work with R2D2 was reported in~\cite{Gulcehr2020}. Storing states in two different replays was presented in~\cite{Nair2018c}. Every time the agent samples for learning, it samples a certain amount from each buffer.  

\paragraph{Discussion}
Using one or two experience replays seems to have negligible impact on performance. However, storing in one experience replay is conceptually and implementation-wise easier. Moreover, it allows agents to stop using imitation experiences when they are not needed anymore.

\subsubsection{Imitation with Exploration Strategy Methods}
Instead of using experience replays, imitations and exploration strategies can be combined directly. In such an approach, imitations are used as a 'kick-start' for exploration.

A single demonstration was used as a starting point for exploration in \cite{Salimans2018a}. The agent randomly explores from a state alongside a single demonstration run. The agent trained from a mediocre demonstration can score highly in Montezuma's Revenge. The auxiliary reward approach was proposed in \citet{Aytar2018a}. The architecture can combine several YouTube videos into a single embedding space for training. The auxiliary reward is added to every $N$ frame from the demonstration video. The agent that can ask for help from the demonstrator was proposed in~\cite{Subramanian2016a}. If the agent detects an unknown environment, the human demonstrator is asked to show the agent how to navigate.

\paragraph{Discussion}
Using imitations as a starting point for exploration has shown impressive performance in difficult exploratory games. In particular, \cite{Aytar2018a} and \cite{Salimans2018a} scored highly in Montezuma's Revenge. This is the effect of overcoming the initial burden of exploration through demonstrations. Approach from \cite{Aytar2018a} can score highly in Montezuma's revenge with just a single demonstration, making it very sample efficient. Meanwhile, the approach from \cite{Aytar2018a} can combine data from multiple sources, making it more suitable for problems with many demonstrations.

\subsubsection{Summary}
A comparison of the imitation methods is presented in Table~\ref{tab:imitation_based}. Imitations in experience replay allow the agent to seamlessly and continuously learn from demonstration experiences. However, imitations with exploration strategies have the potential to find good novel strategies around existing ones. Imitations with exploration strategies have shown a great capability to overcome initial exploration difficulty. Imitations with exploration strategies achieve better performance than using imitations in experience replay only.

\begin{scriptsize}
\begin{ThreePartTable}
\begin{TableNotes}[para, flushleft]
  Legend: A - action space, Ac - action, R - reference, MB - model based, MF - model free, D - discrete, C - continuous, Q - Q values, V - values, P - policy, O - output, S - state space, U - underlying algorithm and Top score on a key benchmark explanation - [benchmark]:[scenario] [score] ([baseline approach] [score]).
\end{TableNotes}
\begin{longtable}   {|P{0.06\linewidth} | P{0.105\linewidth} | P{0.075\linewidth} | P{0.11\linewidth} | P{0.17\linewidth} | P{0.1\linewidth} | P{0.02\linewidth} | P{0.032\linewidth} | P{0.02\linewidth} | }
\caption{Comparison of imitation-based approaches} \label{tab:imitation_based}\\  \hline
    R & Prior Knowledge & U & Method  & Top score on a key benchmark & Input Types & O & MB/ MF & A/ S \\ \hhline{|=|=|=|=|=|=|=|=|=|}
\citet{Hester2018a} & imitation trained policy & DQN   & Imitations in Experience Replay  & Atari: Pitfall 50.8 (Baseline 0) & Atari Images & Q     & MF    & D/ D \\ \hline
\citet{Vecerik2017} & demonstrations & DDPG  & Imitations in Experience Replay  & Peg insertion: 5 ( DDPG -15) & Robot joints angles & Ac    & MF    & C/ C \\ \hline
\citet{Nair2018c} & demonstrations & DDPG  & Imitations in Experience Replay  & Brick stacking: Pick and Place 0.9 (Behaviour cloning 0.8) & Robot joints angles & Ac    & MF    & C/ C \\ \hline
\citet{Gulcehr2020} & demonstrations & R2D2  & Imitations in Experience Replay  & Hard-eight: Drawbridge 12.5 (R2D2:0) & Vizdoom Images & Ac    & MF    & D/ D \\ \hline
\citet{Aytar2018a} & youtube embbeding  & IMPALA & Imitation with Exploration Strategy  & Atari: Montezuma's Revenge 80k (DqfD 4k) & Atari Images & Ac    & MF    & D/ D \\ \hline
\citet{Salimans2018a} & single demonstration & PPO   & Imitation with Exploration Strategy  & Atari: Montezuma Revenge with distraction 74500 (Playing by youtube 41098) & Atari images & Ac    & MF    & D/ D \\ \hline

\insertTableNotes \\
\end{longtable}%
\end{ThreePartTable}
\end{scriptsize}

\subsection{Safe Exploration} \label{sec:Safe_exploration}
\begin{figure*}[htbp]
   \centering
    \includegraphics[width=0.71\textwidth]{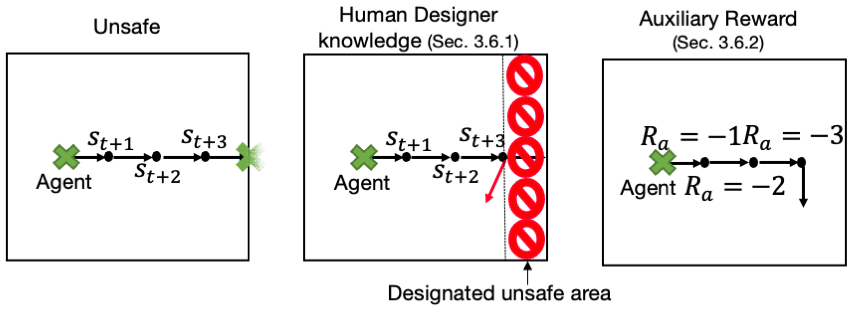}
   \caption{Illustration of safe exploration methods. In safe exploration methods, attempts are made to prevent unsafe behaviours during exploration. Here, three techniques are highlighted: (i) human designer knowledge–the agent's behaviours are limited by human-designed boundaries; (ii) prediction models–the agent learns unsafe behaviours and how to avoid them; and (iii) auxiliary rewards–agents are punished in dangerous states.}
\label{fig:safe_exploration}
\end{figure*}

In safe exploration, the problem of preventing agents from unsafe behaviours is considered. This is an important aspect of exploration research, as the agent's safety needs to be ensured. Safe exploration can be split into three categories: (i) human designer knowledge, (ii) prediction model, and (iii) auxiliary reward as illustrated in Fig.~\ref{fig:safe_exploration}. For more details about safe exploration in reinforcement learning, the reader is invited to read \cite{Garcia2015}.

\subsubsection{Human Designer Knowledge Methods}
Human-designated safety boundaries are used in human designer knowledge methods. Knowledge from the human designer can be split into baseline behaviours, direct human intervention and prediction models.

\begin{figure*}[htbp]
   \centering
    \includegraphics[width=0.8\textwidth]{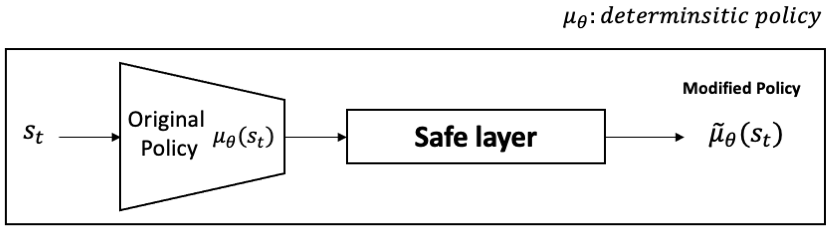}
   \caption{Overview of safe exploration in continuous action spaces~\cite{Dalal2018a}. The additional model is modifying the actions of the original policy.}
\label{fig:Safety_layer}
\end{figure*}

Baseline behaviours impose an impassable safety baseline. Garcia et al. \cite{Garcia2012} proposed the addition of a risk function (which determines unsafe states) and baseline behaviour (which decides what to do in unsafe states). In \cite{Dalal2018a}, the agent was constrained by an additional pre-trained module to prevent unsafe actions as shown in Fig.~\ref{fig:Safety_layer}, while in \cite{Garcelon2020a}, agents are expected to perform no worse than the a priori known baseline. Classifying which object is dangerous and how to avoid them before the training of an agent was proposed in \cite{Hunt2020}. The agent learns how to avoid certain objects rather than states; thus, this approach can be generalised to new scenarios.

The human intervention approach was discussed in \cite{Saunders2018}. During the initial phases of exploration, humans in the loop stop disasters. Then, a supervised trained network of data collected from humans is used as a replacement for humans.

In the prediction model, the human designed safety model determines if the agent's next action leads to an unsafe position and avoids it. In \cite{Turchetta2016a}, a rover traversing a terrain of different heights was considered. The Gaussian process model provides estimates of the height at a given location. If the height is lower than the safe behaviour limit, the robot can explore safely. A heuristic safety model using a priori knowledge was proposed in \cite{Gao2019a}. To this end, they proposed an algorithm called action pruning, which uses the heuristics to prevent agent from committing to unsafe actions.

\paragraph{Discussion}
In human designer knowledge methods, the barriers to unsafe behaviours are placed by a human designer. Baseline behaviours and human intervention methods guarantee certain performance in certain situations but they will only work in pre-defined situations. Prediction model methods require a model of the environment. This can be either in the form of a mathematical model~\cite{Turchetta2016a} or heuristic rules~\cite{Gao2019a}. Prediction models have a higher chance of working on previously unseen environments and have a higher chance of adaptability than baseline behaviours and human intervention methods.


\subsubsection{Auxiliary Reward Methods}
In auxiliary rewards, the agent is punished for putting itself into a dangerous situation. This approach requires the least human intervention, but it generates the weakest safety behaviours.

One of the methods is to find states in which an episode terminates and discourages an agent from approaching using an intrinsic fear \cite{Lipton2016}. The approach counts back a certain number of states from death and applies the distance-to-death penalty. Additionally, they made a simple environment in which a highest positive reward was next to the negative reward. The DQN eventually jumps to the negative rewards. The authors state "We might critically ask, in what real-world scenario, we could depend upon a system [DQN] that cannot solve [these kinds of problems]". A similar approach, but with more stochasticity, was later proposed in \cite{Fatemi2019}.

Allowing the agent to learn undesirable states from previous experiences autonomously was discussed in \cite{Karimpanal2020a}. The states and their advantage values were stored in a common buffer. Then, frequently visited states with the lowest advantage have additional negative rewards associated with them.

\paragraph{Discussion}
Auxiliary rewards can be an effective method of discouraging agents from unsafe behaviours. For example, in~\cite{Lipton2016}, half of the agent's death was prevented. Moreover, some approaches, such as ~\citet{Karimpanal2020a}, have shown the ability to fully automatically determine undesirable states and avoid them.~This, however, assumes that when the agent perishes, it has a low score; this may not always be the case.

\subsubsection{Summary}
An overview of the safety approaches is shown in Table ~\ref{tab:safety_based}. Safety is a vital aspect of reinforcement learning for practical applications in many domains. There are three general approaches: human designer knowledge, prediction models, and auxiliary rewards. Human designer knowledge guarantees safe behaviour in certain states. However, the agent struggles to learn new safe behaviours. Auxiliary reward approaches can adjust to new scenarios, but they require time to train and design of the negative reward.

\begin{scriptsize}
\begin{ThreePartTable}
\begin{TableNotes}[para, flushleft]
  Legend: A - action space, Ac - action, R - reference, MB - model based, MF - model free, D - discrete, C - continuous, Q - Q values, V - values, P - policy, O - output, S - state space, U - underlying algorithm and Top score on a key benchmark explanation - [benchmark]:[scenario] [score] ([baseline approach] [score]).
\end{TableNotes}
\begin{longtable}   {|P{0.095\linewidth} | P{0.105\linewidth} | P{0.075\linewidth} | P{0.09\linewidth} | P{0.15\linewidth} | P{0.1\linewidth} | P{0.02\linewidth} | P{0.032\linewidth} | P{0.02\linewidth} | }
\caption{Comparison of Safe approaches} \label{tab:safety_based}\\  \hline
    R & Prior Knowledge & U & Method  & Top score on a key benchmark & Input Types & O & MB/ MF & A/ S \\ \hhline{|=|=|=|=|=|=|=|=|=|}
\hline
\citet{Garcelon2020a} & Baseline safe policy  & Policy-based UCRL2 & Human Designer Knowledge  & stochastic inventory control: never breaching the safety baseline & amount of products in inventory & P     & MB    & D/ D \\ \hline
\citet{Garcia2012} & baseline behaviour &       & Human Designer Knowledge  & car parking problem 6.5 & angles and positions of respective controllable vehicles & Ac    & MB    & D/ D \\ \hline
\citet{Hunt2020} & pretrained safety network & PPO   & Human Designer Knowledge  & Point mass environment:  0 unsafe actions (PPO 3000) & bird's eye view of the problem & Ac    & MB    & D/ D \\ \hline
\citet{Saunders2018} & human intervention data & DQN   & Human Designer Knowledge  & Atari: Space Invaders 0 catastrophes (DQN 800000) & Atari images & Q     & MB    & D/ D \\ \hline
\citet{Dalal2018a} & pretrained safety model & DDPG  & Human Designer Knowledge  & spaceship: Arena 1000  (DDPG 300) & x-y position & Ac    & MB    & C/ C \\ \hline
\citet{Gao2019a} & environmental knowledge & PPO   & Human Designer Knowledge  & Pommerman: 0.8 (Baseline 0) & Agent, enemy agents and bombs positions & Ac    & MB    & D/ D \\ \hline
\citet{Turchetta2016a} &       & Bayesian optimisation & Human Designer Knowledge  & Simulated rover: 80\% exploration (Random 0.98\%) & x-y position & Ac    & MB    & C/ C \\ \hline
\citet{Fatemi2019} &       & DQN   & Auxiliary Reward  & Bridge: optimal after 14k episodes (ten times faster then competitor) & card types/ atari images & Q     & MB    & D/ D \\ \hline
\citet{Lipton2016} &       & DQN   & Auxiliary Reward  & Atari: Asteroids  total death 40,000 (DQN 80,000) & Atari images & Ac    & MB    & C/ C \\ \hline
\citet{Karimpanal2020a} &       & Q-learning and DDPG & Auxiliary Reward  & Navigation environment: -3 (PQRL -3.5) & enumarated state id & Q     & MF    & C/ C \\ \hline

\insertTableNotes \\
\end{longtable}%
\end{ThreePartTable}
\end{scriptsize}

\subsection{Random-Based Methods} \label{sec:Random}

\begin{figure*}[htbp]
   \centering
    \includegraphics[width=0.69\textwidth]{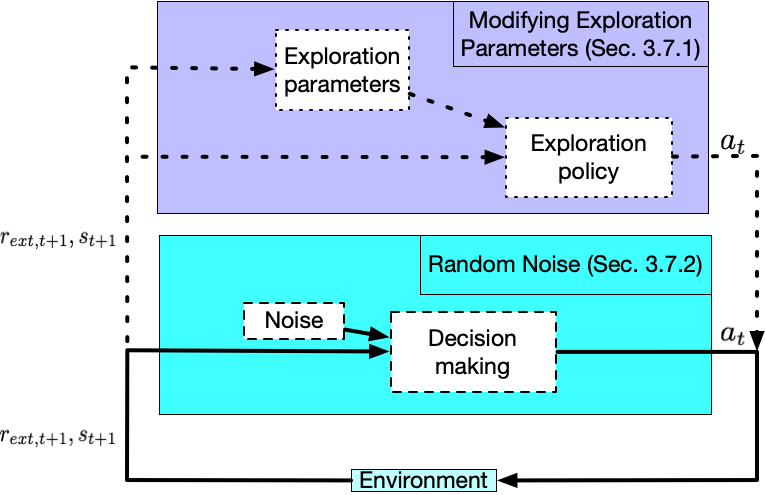}
   \caption{Overview of random based methods. In random-based methods, simple random exploration is modified for improved efficiency. In modifying the states for exploration, the number of actions to be taken randomly is reduced. In modifying the exploration parameters, the exploration is automatically decided. In the network parameter noise, the noise is imposed on the policy parameters.}
\label{fig:random_overview}
\end{figure*}

In random-based approaches, improvements to simple random exploration are discussed. Random exploration tends to be inefficient as it often revisits the same states. To solve this problem, the following approaches are considered: (i) reduced states/actions for exploration methods, (ii) exploration parameters methods, and (iii) network parameter noise methods, as illustrated in Fig. \ref{fig:random_overview}.

\subsubsection{Exploration Parameters Methods}
In this section, exploration is parameterized (for example, $\epsilon$ in $\epsilon$-greedy). Then, the parameters are modified according to the agent's learning progress. 

One technique to adapt the exploration rate is by simply considering a reward and adjusting the random exploration parameter accordingly, as described in \cite{Patrascu1999}. Using a pure reward can lead to problems with sparse rewards. To solve this problem, in \cite{Tokic2010}, $\epsilon$ was made to depend on the error of the value-function estimates instead of the reward. It is also possible to determine the amount of random exploration using the environmental model entropy, as discussed in \cite{Usama2019a}. The learning rate \cite{Shani2019a} can also depend on exploration in which a parameter $\alpha$ that is functionally equivalent to the learning rate is introduced. If the agent is exploring a lot, the value of $\alpha$ slows down the learning to account for uncertainty. Khamassi et al. \cite{Khamassi2017} used long-term and short-term reward averages to control exploration and exploitation. When the short-term average is below the long-term average, exploration should be increased.

Chang et al. \cite{Chang2004a} used multiple agents (ants) to adjust the exploration parameters. At each step, the ants chose their actions randomly, but were skewed by pheromone values left by other ants.

Another approach of this type could be reducing states for exploration based on some predefined metric. An approach using the adaptive resonance theorem (ART)~\cite{Grossberg1987} was presented in \cite{Teng2012a} and was later extended in \cite{Wang2018a}. In ART, knowledge about actions can be split into: (i) positive chunk which leads to positive rewards, (ii) negative chunk which leads to negative results, and (iii) empty chunk which is not yet taken. In this approach, the action is randomly chosen from positive and no chunks; thus, the agent is exploring either new things or ones with the positive reward. Wang et al. \cite{Wang2018a} extended this to include the probability of selecting the remaining actions based on how well they are known.

\paragraph{Discussion}
Different parameters can be changed based on learning progress. Initially, approaches used learning progress, reward, or value of states  to determine the rate of exploration. The challenge with these approaches is determining the parameters controlling the exploration. However, it is also possible to adjust the learning rate based on exploration \cite{Shani2019a}. The advantage is that the agent avoids learning uncertain information, but it slows down the training. Finally, reducing states for exploration can make exploration more sample efficient, but it struggles to account for unseen states that occurs after the eliminated states.

\subsubsection{Random Noise}
In random noise approaches, random noise is used for exploration. The random noise can be either imposed on networks parameters or be produced based on states met during exploration.

The easiest method of including the noise is to include a fixed amount of noise \cite{Ruckstiess2010}. This paper reviews the usage of small perturbations in the parameter space. In \cite{Shibata2015}, chaotic networks were used to induce the noise in the network. It is also possible to adjust the noise strength using backpropagation, as described in \cite{Fortunato2018} where the noise is created by a constant noise source multiplied by a gradient-adaptable parameter. Another way of the using the noise is by comparing the decision made by the noisy and noiseless policy \cite{Plappert2018a}. Exploration is imposed, if decisions are sufficiently different. 

In \cite{Ruckstiess2008}, the problem of assigning rewards when the same state is present multiple times is discussed. In such a problem, the agent will be likely to take different actions for the same state, making credit assignment difficult. To solve this problem, a random action generation function dependent on the input state was developed; if the state is the same, the random action is the same.

\paragraph{Discussion}
Network parameter noise was first developed for evolutionary approaches, such as \cite{Ruckstiess2010}. Recently, the noise of parameters has been used in policy-based methods. In particular, good performance was achieved in \cite{Fortunato2018} which was able to achieve 50\% improvement averaged over 52 Atari games. 

\subsubsection{Summary}
A comparison of the random-based approach is presented in Table ~\ref{tab:random_based}. The key advantage of reduced states for exploration methods is that the exploration can be very effective, but it needs to hold the memory of where it has been. Exploration parameter methods solves a trade-off between exploration and exploitation well; however, the agent can still get stuck in exploring unnecessary states. The random noise approaches are very simple to implement and show promising results, but they rely on careful tuning of parameters by designers.

\begin{scriptsize}
\begin{ThreePartTable}
\begin{TableNotes}[para, flushleft]
  Legend: A - action space, Ac - action, R - reference, MB - model based, MF - model free, D - discrete, C - continuous, Q - Q values, V - values, P - policy, O - output, S - state space, U - underlying algorithm and Top score on a key benchmark explanation - [benchmark]:[scenario] [score] ([baseline approach] [score]).
\end{TableNotes}
\begin{longtable}   {|P{0.105\linewidth} | P{0.105\linewidth} | P{0.075\linewidth} | P{0.09\linewidth} | P{0.15\linewidth} | P{0.1\linewidth} | P{0.02\linewidth} | P{0.032\linewidth} | P{0.02\linewidth} | }
\caption{Comparison of Random based approaches} \label{tab:random_based}\\  \hline
    R & Prior Knowledge & U & Method  & Top score on a key benchmark & Input Types & O & MB/ MF & A/ S \\ \hhline{|=|=|=|=|=|=|=|=|=|}
\citet{Wang2018a} &       & ART & Exploration parameters & minefield navigation (successful rate): 91\% (Baseline 91\%) & Vehicles positions & Q     & MB    & C/ C \\ \hline
\citet{Shani2019a} &       & DDQN and DDPG & Exploration parameters  & Atari: Frostbite 2686 (DDPG 1720); Mujoco:  HalfCheetah 4579 (DDPG 2255) & Atari images, Mujoco joints angles & Ac/ Q & MF    & C/ C \\ \hline
\citet{Patrascu1999} &       & Fuzzy ART MAP architecture & Exploration parameters  & Changing world environment (grid with two alternating paths to reward) 0.9 & Enumerated state id & Ac    & MB    & D/ D \\ \hline
\citet{Usama2019a} &       & DQN & Exploration parameters  & VizDoom: Defend the centre 12.2 ($\epsilon$-greedy 11.8)  & Images & Q     & MB    & C/ C \\ \hline
\citet{Tokic2010} &       & V-learning & Exploration parameters  & Multi-arm bandit: 1.42 (Softmax 1.38) & Enumarated state id & V     & MF    & D/ D \\ \hline
\citet{Khamassi2017} &       & Q-learning & Exploration parameters  & Nao simulator: Engagement 10 (Kalman-QL 5) & Robot joints angles & Q     & MF    & D/ D \\ \hline
\citet{Shibata2015} &       & Actor-critic & Random noise & area with randomly positioned obstacle: 0.6 out of 1 & Enumarated state id & Ac    & MF    & D/ D \\ \hline
\citet{Plappert2018a} & measure of policies distance & DQN, DDPG and TRPO & Random noise & Atari: BeamRdier 9000 ($\epsilon$-greedy 5000); Mujoco: Half cheetah 5000 ($\epsilon$-greedy 1500) & Atari images/Mujoco joints angles & Ac/ Q & MF    & C/ C \\ \hline
\citet{Shibata2015} &       &       & Random noise & Multi Arm Bandit Problem: 1 (Optimal) & Stateless & Ac    & MB    & C/ C \\ \hline
\citet{Fortunato2018} &       &       & Random noise & Atari: 57 games 633 points (Dueling DQN 524) & Atari images & Ac    & MF    & C/ C \\ \hline

\insertTableNotes \\
\end{longtable}%
\end{ThreePartTable}
\end{scriptsize}

\section{Future Challenges} \label{sec:future_challenges}
In this section, we discuss the following future challenges on exploration in reinforcement learning: evaluation, scalability, exploration-exploitation dilemma, intrinsic reward, noisy TV problems, safety, and transferability.

\paragraph{Evaluation}
Currently, evaluating and comparing different exploration algorithms is challenging. This issue arises from three reasons: lack of a common benchmark, lack of a common evaluation strategy, and lack of good metrics to measure exploration. 

Currently, four major benchmarks used by the community are VizDoom \cite{Kempka2016}, Minecraft \cite{Johnson2016a}, Atari Games \cite{Badia2020b} and Mujoco \cite{Todorov2012}. Each benchmark is characterised by different complexities in terms of state space, reward sparseness, and action space. Moreover, each benchmark offers several scenarios with various degrees of complexity. Such a wealth of benchmarks is desirable for exposure of agents to various complexities; however, the difference in complexity between different benchmarks is well-understood. This leads to difficulty in comparing algorithms using different benchmarks. There have been attempts to solve the evaluation issues using a common benchmark, for example, in \cite{Osband2020}. However, this study is not commonly adopted yet.

Regarding the evaluation strategy, most algorithms use a reward after a certain number of steps. Note that in the context of this paragraph, steps could also mean episodes, iterations and epochs. This makes the reporting of results inconsistent in two aspects: (i) the number of steps in which the algorithm was tested and (ii) how the reward is reported. The first makes comparisons between algorithms difficult because performance can vary widely depending on when the comparison is made. The second concern is how rewards are reported. Most authors choose to report the average reward the agent has scored; however, sometimes comparison with the average human performance is used (without clear indication of what average human performance means exactly). Moreover, sometimes the distinction between the average reward or maximum reward is not clearly made. 

Finally, it is arguable if a reward is an appropriate measure for evaluation \cite{Stadie2015}. One of the key issues is that it fails to account for the speed of learning, which should be higher if exploration is more efficient~\cite{Stadie2015}. Attempts have been made to address this issue in \cite{Stadie2015}, but as of the time of writing this review paper, this new metric is not widely adopted. Another issue with rewards is that it does not provide any information regarding the goodness of exploratory behaviour. This is even more difficult in continuous action space problems where computing novelty is considerably more challenging. 

\paragraph{Scalability}
Exploration in reinforcement learning does not scale well to real-world problems. This is caused by two limitations: training time and inefficient state representation. Currently, even the fastest training requires millions of samples in complex environments. Note that even the most complex environments currently used in reinforcement learning are still relatively simple compared to the real world. In the real world, collecting millions of samples for training is unrealistic owing to wear and tear of physical devices. To cope with the real world, either a sim-to-real gap needs to be reduced or exploration needs to become more sample efficient.

Another limitation is efficient state representation so that memorising states and actions is possible in large domains. For example, Go-Explore \cite{Ecoffet2019} does not scale up well if the environment is large. This problem was discussed in \cite{Jaegle2019a} by comparing how the brain stores memories and computes novelty. It states that the human brain is much faster at determining scene novelty and has a much larger capacity. To achieve this, the brain uses an agreement between multiple neurons. The more neurons indicate that the given image is novel, the higher the novelty is. Thus, the brain does not need to remember full states; instead, it trains itself to recognise the novelty. This is currently unmatched in reinforcement learning in terms of the representation efficiency. 

\paragraph{Exploration-exploitation dilemma}
The exploration–exploitation dilemma is an ongoing research topic not only in reinforcement learning but also in a general problem. Most current exploration approaches have a built-in solution to exploration-exploitation, but not all methods do. This is particularly true in goal-based methods that rely on hand-designed solutions. Moreover, even in approaches that solve it automatically, the balance is still mostly decided by the designer-provided threshold. One potential way of solving this problem is to train a set of skills (policies) during exploration and combine skills in greater goal-oriented policies \cite{OpenAI2021}. This is similar to how humans solve problems by learning smaller skills and then using them later to exploit them as a larger policy.

\paragraph{Intrinsic reward}
Reward novel states and diverse behaviour approaches can be improved in two ways: (i) the agent should be more free to reward itself and (ii) better balance between long-term and short-term novelty should be achieved. 

In most intrinsic reward approaches, the exact reward formulation is performed by an expert. Designing a reward that guarantees good exploration is a challenging and time-consuming task. Moreover, there might be ways of rewarding agents which were not conceived by designers. Thus, it could be beneficial if an agent is not only trained in the environment but is also trained on how to reward itself. This would be closer to human behaviour where the self-rewarding mechanism was developed through evolution.

Balancing the long-term novelty and short-term novelty is another challenge. In this problem, the agent tries to balance two factors: revisiting states often to find something new or abandoning states quickly to try to find something new. This is currently a hand-designed parameter, but its tuning is time-consuming. Recently, there has been a fix proposed in \cite{Badia2020b} where meta-learning decides the appropriate balance, but at the cost of computational complexity for training.

\paragraph{Noisy-TV problem}

The noisy-TV (or couch potato problem) remains largely unsolved. While using memory can be used to solve it, they are limited by memory requirements. Thus, it can be envisioned that if the noisy sequence is very long and the state space is complex, memory approaches will also struggle to solve it. One method that has shown some promise is the use of clustering \cite{Kovac2020} to cluster noisy states and avoid that cluster. However, this requires the design of correct clusters.

\paragraph{Optimal exploration}

One area which is rarely considered in the current exploration in reinforcement learning research is how to explore optimally. For optimal exploration, the agent does not revisit states unnecessarily and explores the most promising areas first. This problem and the proposed solution are described in detail in \cite{Zhang2019a}. The solution uses a demand matrix, which is an $m$ by $n$ matrix of $m$ states and $n$ actions, indicating state-action exploration counts. It then defines the exploration cost for exploration policy, which is the number of steps each state-action pair needs to be explored. Note that the demand matrix does not need to be known a priori and can be updated online. This aspect needs further developments.

\paragraph{Safe exploration}

Safe exploration is of paramount importance for real-world applications. However, so far, there have been very few approaches to cope with this issue. Most of them rely heavily on hand-designed rules to prevent catastrophes. Moreover, it has been shown in \cite{Lipton2016} that current reinforcement learning is struggling to prevent catastrophes even with carefully engineered rewards. Thus, there exists a need for the agent to recognise unsafe situations and act accordingly. Moreover, what constitutes an unsafe situation is not well defined beyond hand-designed rules. This leads to problems with regard to the scalability and transferability of safe exploration in reinforcement learning. A more rigorous definition of an unsafe situation would be beneficial to address this problem.

\paragraph{Transferability}
Most exploratory approaches are currently limited to the domain on which they were trained. When faced with new environments (e.g., increased state space and different reward functions), exploration strategies do not seem to perform well~\cite{Dhiman2018a,Raileanu2020}. Coping with this issue would be helpful in two scenarios. First, it would be beneficial to be able to teach the agent behaviours in smaller scenarios and then allow it to perform well on larger scenarios to alleviate computational issues. Second, in some domains, defining state spaces suitable for exploration is challenging and may vary in size significantly between tasks (e.g., search for a victim of an accident).

\section{Conclusions}
This paper presents a review of the exploration in reinforcement learning. The following methods were discussed: reward novel states, reward diverse behaviours, goal-based methods, uncertainty, imitation-based methods, safe exploration, and random methods.

In reward novel state methods, the agent is given a reward for discovering a novel or surprising state. This reward can be computed using prediction error, count, or memory. In prediction error methods, the reward is given based on the accuracy of the agent's internal environmental model. In count-based methods, the reward is given based on how often a given state is visited. In memory-based methods, the reward is computed based on how different a state is compared to other states in a buffer. 

In reward diverse behaviour methods, the agent is rewarded for discovering as many diverse behaviours as possible. Note here that we use word behaviour loosely as a sequence of actions or a policy. Reward diverse behaviour methods can be divided into: evolutionary strategies and policy learning. In evolution strategies, diversity among the population of agents is encouraged. In policy learning, the diversity of policy parameters is encouraged.

In goal-based methods, the agent is given the goal of either exploring from or exploring while trying to reach the goal. In the first method, the agent chooses the goal to get to and then explore from it. This results in a very efficient exploration as the agent visits predominantly unknown areas. In the second method, called the exploratory goal, the agent is exploring while travelling toward a goal. The key idea of this method is to provide goals which are suitable for exploration. 

In probabilistic methods, the agent holds an uncertainty model about the environment and uses it to make its next move. The uncertainty method has two subcategories: optimistic and uncertainty methods. In optimistic methods, the agent follows the \emph{optimism under uncertainty} principle. This means that the agent will sample the most optimistic understanding of the reward. In uncertainty methods, the agent will sample from internal uncertainty to move toward the least known areas.

Imitation-based methods rely on using demonstrations to help exploration. In general, there are two methods: combining demonstrations with experience replay and combining them with an exploration strategy. In the first method, samples from demonstrations and collected by the agent are combined into one buffer for the agent to learn from. In the second method, the demonstrations are used as a starting point for other exploration techniques such as the reward novel state. 

Safe exploration methods were devised to ensure the safe behaviour of the agents during exploration. In safe exploration, the most prevalent method is to use human designer knowledge to develop boundaries for the agent. Furthermore, it is possible to train a model that predicts and stops agents from making a disastrous move. Finally, the agent can be discouraged from visiting dangerous states with a negative reward.

Random exploration methods improve standard random exploration. These improvements include modifying the states for exploration, modifying exploration parameters, and putting the noise on network parameters. In modifying states for exploration, certain states and actions are removed from the random choice if they have been sufficiently explored. In modifying exploration parameter methods, the parameters affecting when to randomly explore are automatically chosen based on the agent's learning progress. Lastly, in the network parameter noise approach, random noise is applied to the parameters to induce exploration before the weight convergence.


Finally, the best approaches in terms of ease of implementation, computational cost and overall performance are highlighted. The easiest methods to implement are reward novel states, reward diverse behaviours and random-based approaches. Basic implementation of those approaches can be used with almost any other existing reinforcement learning algorithms; they might require a few additions and tuning to work. In terms of computational efficiency, random-based, reward novel states and reward divers behaviours generally require the least resources. Particularly, random-based approaches are computationally efficient as the additional components are lightweight. Currently, best-performing methods are goal-based and reward novel states methods where goal-based methods have achieved high scores in difficult exploratory problems such as Montezuma's revenge. However, goal-based methods tend to be the most complex in terms of implementation. Overall, reward novel states methods seem like a good compromise between ease of implementation and performance.

\section*{Declaration of Competing Interest}
The authors declare that they have no known competing financial interests or personal relationships that could have appeared to influence the work reported in this paper.


\bibliographystyle{elsarticle-num-names}
\bibliography{RL_exploration}

\begin{thebibliography}{161}
\expandafter\ifx\csname natexlab\endcsname\relax\def\natexlab#1{#1}\fi
\providecommand{\url}[1]{\texttt{#1}}
\providecommand{\href}[2]{#2}
\providecommand{\path}[1]{#1}
\providecommand{\DOIprefix}{doi:}
\providecommand{\ArXivprefix}{arXiv:}
\providecommand{\URLprefix}{URL: }
\providecommand{\Pubmedprefix}{pmid:}
\providecommand{\doi}[1]{\href{http://dx.doi.org/#1}{\path{#1}}}
\providecommand{\Pubmed}[1]{\href{pmid:#1}{\path{#1}}}
\providecommand{\bibinfo}[2]{#2}
\ifx\xfnm\relax \def\xfnm[#1]{\unskip,\space#1}\fi
\bibitem[{Sutton and Barto(2018)}]{Sutton2020}
\bibinfo{author}{R.~S. Sutton}, \bibinfo{author}{A.~G. Barto},
  \bibinfo{title}{Reinforcement Learning: An Introduction},
  \bibinfo{publisher}{A Bradford Book}, \bibinfo{address}{Cambridge, MA, USA},
  \bibinfo{year}{2018}.
\bibitem[{Mnih et~al.(2015)Mnih, Kavukcuoglu, Silver, Rusu, Veness, Bellemare,
  Graves, Riedmiller, Fidjeland, Ostrovski, Petersen, Beattie, Sadik,
  Antonoglou, King, Kumaran, Wierstra, Legg, and Hassabis}]{Mnih2015b}
\bibinfo{author}{V.~Mnih}, \bibinfo{author}{K.~Kavukcuoglu},
  \bibinfo{author}{D.~Silver}, \bibinfo{author}{A.~A. Rusu},
  \bibinfo{author}{J.~Veness}, \bibinfo{author}{M.~G. Bellemare},
  \bibinfo{author}{A.~Graves}, \bibinfo{author}{M.~Riedmiller},
  \bibinfo{author}{A.~K. Fidjeland}, \bibinfo{author}{G.~Ostrovski},
  \bibinfo{author}{S.~Petersen}, \bibinfo{author}{C.~Beattie},
  \bibinfo{author}{A.~Sadik}, \bibinfo{author}{I.~Antonoglou},
  \bibinfo{author}{H.~King}, \bibinfo{author}{D.~Kumaran},
  \bibinfo{author}{D.~Wierstra}, \bibinfo{author}{S.~Legg},
  \bibinfo{author}{D.~Hassabis},
\newblock \bibinfo{title}{{Human-level Control Through Deep Reinforcement
  Learning}},
\newblock \bibinfo{journal}{Nature} \bibinfo{volume}{518}
  (\bibinfo{year}{2015}) \bibinfo{pages}{529--533}. \URLprefix
  \url{http://dx.doi.org/10.1038/nature14236}.
  \DOIprefix\doi{10.1038/nature14236}.
\bibitem[{Lillicrap et~al.(2016)Lillicrap, Hunt, Pritzel, Heess, Erez, Tassa,
  Silver, and Wierstra}]{Lillicrap2016}
\bibinfo{author}{T.~P. Lillicrap}, \bibinfo{author}{J.~J. Hunt},
  \bibinfo{author}{A.~Pritzel}, \bibinfo{author}{N.~Heess},
  \bibinfo{author}{T.~Erez}, \bibinfo{author}{Y.~Tassa},
  \bibinfo{author}{D.~Silver}, \bibinfo{author}{D.~Wierstra},
\newblock \bibinfo{title}{{Continuous Control with Deep Reinforcement
  Learning}},
\newblock \bibinfo{journal}{4th International Conference on Learning
  Representations, ICLR 2016}  (\bibinfo{year}{2016}).
  \href{http://arxiv.org/abs/1509.02971}{{\tt arXiv:1509.02971}}.
\bibitem[{Lee and Bang(2020)}]{Lee2020}
\bibinfo{author}{S.~Lee}, \bibinfo{author}{H.~Bang},
\newblock \bibinfo{title}{{Automatic Gain Tuning Method of a Quad-Rotor
  Geometric Attitude Controller Using A3C}},
\newblock \bibinfo{journal}{International Journal of Aeronautical and Space
  Sciences} \bibinfo{volume}{21} (\bibinfo{year}{2020})
  \bibinfo{pages}{469--478}. \URLprefix
  \url{https://doi.org/10.1007/s42405-019-00233-x}.
  \DOIprefix\doi{10.1007/s42405-019-00233-x}.
\bibitem[{Polvara et~al.(2017)Polvara, Patacchiola, Sharma, Wan, Manning,
  Sutton, and Cangelosi}]{Polvara2017}
\bibinfo{author}{R.~Polvara}, \bibinfo{author}{M.~Patacchiola},
  \bibinfo{author}{S.~Sharma}, \bibinfo{author}{J.~Wan},
  \bibinfo{author}{A.~Manning}, \bibinfo{author}{R.~Sutton},
  \bibinfo{author}{A.~Cangelosi},
\newblock \bibinfo{title}{{Autonomous Quadrotor Landing using Deep
  Reinforcement Learning}}  (\bibinfo{year}{2017}). \URLprefix
  \url{http://arxiv.org/abs/1709.03339}.
  \href{http://arxiv.org/abs/1709.03339}{{\tt arXiv:1709.03339}}.
\bibitem[{Kiran et~al.(2021)Kiran, Sobh, Talpaert, Mannion, Sallab, Yogamani,
  and Perez}]{Kiran2021}
\bibinfo{author}{B.~R. Kiran}, \bibinfo{author}{I.~Sobh},
  \bibinfo{author}{V.~Talpaert}, \bibinfo{author}{P.~Mannion},
  \bibinfo{author}{A.~A. Sallab}, \bibinfo{author}{S.~Yogamani},
  \bibinfo{author}{P.~Perez},
\newblock \bibinfo{title}{{Deep Reinforcement Learning for Autonomous Driving:
  A Survey}},
\newblock \bibinfo{journal}{IEEE Transactions on Intelligent Transportation
  Systems}  (\bibinfo{year}{2021}) \bibinfo{pages}{1--18}.
  \DOIprefix\doi{10.1109/TITS.2021.3054625}.
  \href{http://arxiv.org/abs/2002.00444}{{\tt arXiv:2002.00444}}.
\bibitem[{Yu et~al.(2019)Yu, Liu, and Nemati}]{Yu2019}
\bibinfo{author}{C.~Yu}, \bibinfo{author}{J.~Liu}, \bibinfo{author}{S.~Nemati},
\newblock \bibinfo{title}{{Reinforcement Learning in Healthcare: A Survey}}
  (\bibinfo{year}{2019}). \URLprefix \url{http://arxiv.org/abs/1908.08796}.
  \href{http://arxiv.org/abs/1908.08796}{{\tt arXiv:1908.08796}}.
\bibitem[{Irpan(2018)}]{Irpan2018}
\bibinfo{author}{A.~Irpan}, \bibinfo{title}{Deep reinforcement learning doesn't
  work yet},
  \bibinfo{howpublished}{\url{https://www.alexirpan.com/2018/02/14/rl-hard.html}},
  \bibinfo{year}{2018}.
\bibitem[{Clark and Amodei(2016)}]{Clark2016}
\bibinfo{author}{J.~Clark}, \bibinfo{author}{D.~Amodei}, \bibinfo{title}{Faulty
  reward functions in the wild},
  \bibinfo{howpublished}{\url{https://openai.com/blog/faulty-reward-functions/}},
  \bibinfo{year}{2016}.
\bibitem[{Schmidhuber(1991{\natexlab{a}})}]{Schmidhuber1991}
\bibinfo{author}{J.~Schmidhuber},
\newblock \bibinfo{title}{Curious model-building control systems},
\newblock \bibinfo{journal}{1991 IEEE International Joint Conference on Neural
  Networks, IJCNN 1991}  (\bibinfo{year}{1991}{\natexlab{a}})
  \bibinfo{pages}{1458--1463}. \DOIprefix\doi{10.1109/ijcnn.1991.170605}.
\bibitem[{Schmidhuber(1991{\natexlab{b}})}]{Schmidhuber1991a}
\bibinfo{author}{J.~Schmidhuber},
\newblock \bibinfo{title}{{A Possibility for Implementing Curiosity And Boredom
  in Model-Building Neural Controllers}},
\newblock \bibinfo{journal}{Proceedings of the First International Conference
  on Simulation of Adaptive Behavior} \bibinfo{volume}{1}
  (\bibinfo{year}{1991}{\natexlab{b}}) \bibinfo{pages}{5--10}. \URLprefix
  \url{ftp://ftp.idsia.ch/pub/juergen/curiositysab.pdf}.
\bibitem[{Eysenbach et~al.(2019)Eysenbach, Gupta, Ibarz, and
  Levine}]{Eysenbach2019a}
\bibinfo{author}{B.~Eysenbach}, \bibinfo{author}{A.~Gupta},
  \bibinfo{author}{J.~Ibarz}, \bibinfo{author}{S.~Levine},
\newblock \bibinfo{title}{{Diversity Is All You Need}},
\newblock \bibinfo{journal}{7th International Conference on Learning
  Representations, ICLR 2019}  (\bibinfo{year}{2019}). \URLprefix
  \url{https://openreview.net/pdf?id=SJx63jRqFm}.
\bibitem[{Burda et~al.(2018)Burda, Edwards, Storkey, and Klimov}]{Burda2018b}
\bibinfo{author}{Y.~Burda}, \bibinfo{author}{H.~Edwards},
  \bibinfo{author}{A.~Storkey}, \bibinfo{author}{O.~Klimov},
\newblock \bibinfo{title}{{Exploration by Random Network Distillation}},
\newblock \bibinfo{journal}{7th International Conference on Learning
  Representations, ICLR 2019}  (\bibinfo{year}{2018}). \URLprefix
  \url{http://arxiv.org/abs/1810.12894}. \DOIprefix\doi{arXiv:1810.12894v1}.
  \href{http://arxiv.org/abs/1810.12894}{{\tt arXiv:1810.12894}}.
\bibitem[{Bellemare et~al.(2016)Bellemare, Srinivasan, Ostrovski, Schaul,
  Saxton, and Munos}]{Bellemare2016}
\bibinfo{author}{M.~G. Bellemare}, \bibinfo{author}{S.~Srinivasan},
  \bibinfo{author}{G.~Ostrovski}, \bibinfo{author}{T.~Schaul},
  \bibinfo{author}{D.~Saxton}, \bibinfo{author}{R.~Munos},
\newblock \bibinfo{title}{{Unifying Count-Based Exploration and Intrinsic
  Motivation}},
\newblock \bibinfo{journal}{Conference on Neural Information Processing
  Systems, NeurIPS 2016}  (\bibinfo{year}{2016}). \URLprefix
  \url{http://arxiv.org/abs/1606.01868}. \DOIprefix\doi{10.1002/pola.10609}.
  \href{http://arxiv.org/abs/1606.01868}{{\tt arXiv:1606.01868}}.
\bibitem[{Badia et~al.(2020)Badia, Piot, Kapturowski, Sprechmann, Vitvitskyi,
  Guo, and Blundell}]{Badia2020b}
\bibinfo{author}{A.~P. Badia}, \bibinfo{author}{B.~Piot},
  \bibinfo{author}{S.~Kapturowski}, \bibinfo{author}{P.~Sprechmann},
  \bibinfo{author}{A.~Vitvitskyi}, \bibinfo{author}{D.~Guo},
  \bibinfo{author}{C.~Blundell},
\newblock \bibinfo{title}{{Agent57: Outperforming the Atari Human Benchmark}}
  (\bibinfo{year}{2020}). \URLprefix \url{http://arxiv.org/abs/2003.13350}.
  \href{http://arxiv.org/abs/2003.13350}{{\tt arXiv:2003.13350}}.
\bibitem[{Aubret et~al.(2019)Aubret, Matignon, and Hassas}]{Aubret2019}
\bibinfo{author}{A.~Aubret}, \bibinfo{author}{L.~Matignon},
  \bibinfo{author}{S.~Hassas},
\newblock \bibinfo{title}{{A survey on Intrinsic Motivation in Reinforcement
  Learning}},
\newblock \bibinfo{journal}{arXiv}  (\bibinfo{year}{2019}).
  \href{http://arxiv.org/abs/1908.06976}{{\tt arXiv:1908.06976}}.
\bibitem[{Li(2018)}]{Li2018c}
\bibinfo{author}{Y.~Li},
\newblock \bibinfo{title}{Deep reinforcement learning}  (\bibinfo{year}{2018}).
  \DOIprefix\doi{10.18653/v1/p18-5007}.
  \href{http://arxiv.org/abs/1911.10107}{{\tt arXiv:1911.10107}}.
\bibitem[{Nguyen et~al.(2018)Nguyen, Nguyen, and Nahavandi}]{Nguyen2018}
\bibinfo{author}{T.~T. Nguyen}, \bibinfo{author}{N.~D. Nguyen},
  \bibinfo{author}{S.~Nahavandi},
\newblock \bibinfo{title}{{Deep Reinforcement Learning For Multi-Agent Systems:
  A Review Of Challenges, Solutions and Applications}},
\newblock \bibinfo{journal}{arXiv} \bibinfo{volume}{50} (\bibinfo{year}{2018})
  \bibinfo{pages}{3826--3839}.
\bibitem[{Levine(2018)}]{Levine2018a}
\bibinfo{author}{S.~Levine},
\newblock \bibinfo{title}{{Reinforcement Learning and Control as Probabilistic
  Inference: Tutorial and Review}},
\newblock \bibinfo{journal}{arXiv}  (\bibinfo{year}{2018}).
  \href{http://arxiv.org/abs/1805.00909}{{\tt arXiv:1805.00909}}.
\bibitem[{Lazaridis et~al.(2020)Lazaridis, Fachantidis, and
  Vlahavas}]{Lazaridis2019}
\bibinfo{author}{A.~Lazaridis}, \bibinfo{author}{A.~Fachantidis},
  \bibinfo{author}{I.~Vlahavas},
\newblock \bibinfo{title}{Deep reinforcement learning: A state-of-the-art
  walkthrough},
\newblock \bibinfo{journal}{Journal of Artificial Intelligence Research}
  \bibinfo{volume}{69} (\bibinfo{year}{2020}).
\bibitem[{Mcfarlane(1999)}]{Mcfarlane1999}
\bibinfo{author}{R.~Mcfarlane},
\newblock \bibinfo{title}{{A Survey of Exploration Strategies in Reinforcement
  Learning}}  (\bibinfo{year}{1999}) \bibinfo{pages}{1--10}. \URLprefix
  \url{https://pdfs.semanticscholar.org/0276/1533d794ed9ed5dfd0295f2577e1e98c4fe2.pdf}.
\bibitem[{Williams(1992)}]{Williams1992}
\bibinfo{author}{R.~J. Williams},
\newblock \bibinfo{title}{{Simple Statistical Gradient-Following Algorithms For
  Connectionist Reinforcement Learning}},
\newblock \bibinfo{journal}{Machine Learning} \bibinfo{volume}{8}
  (\bibinfo{year}{1992}) \bibinfo{pages}{229--256}.
  \DOIprefix\doi{10.1007/bf00992696}.
\bibitem[{Arulkumaran et~al.(2017)Arulkumaran, Deisenroth, Brundage, and
  Bharath}]{Arulkumaran2017}
\bibinfo{author}{K.~Arulkumaran}, \bibinfo{author}{M.~P. Deisenroth},
  \bibinfo{author}{M.~Brundage}, \bibinfo{author}{A.~A. Bharath},
\newblock \bibinfo{title}{{Deep Reinforcement Learning: A Brief Survey}},
\newblock \bibinfo{journal}{IEEE Signal Processing Magazine}
  \bibinfo{volume}{34} (\bibinfo{year}{2017}) \bibinfo{pages}{26--38}.
  \DOIprefix\doi{10.1109/MSP.2017.2743240}.
  \href{http://arxiv.org/abs/arXiv:1708.05866v2}{{\tt
  arXiv:arXiv:1708.05866v2}}.
\bibitem[{Exp(2020)}]{Exploration}
\bibinfo{title}{Exploration},
  \bibinfo{howpublished}{\url{https://dictionary.cambridge.org/dictionary/english/exploration}},
  \bibinfo{year}{2020}. \bibinfo{note}{Accessed: 2020-04-09}.
\bibitem[{Bellemare et~al.(2013)Bellemare, Naddaf, Veness, and
  Bowling}]{Bellemare2013}
\bibinfo{author}{M.~G. Bellemare}, \bibinfo{author}{Y.~Naddaf},
  \bibinfo{author}{J.~Veness}, \bibinfo{author}{M.~Bowling},
\newblock \bibinfo{title}{{The Arcade Learning Environment: An Evaluation
  Platform For General Agents}},
\newblock \bibinfo{journal}{Journal of Artificial Intelligence Research}
  \bibinfo{volume}{47} (\bibinfo{year}{2013}) \bibinfo{pages}{253--279}.
  \DOIprefix\doi{10.1613/jair.3912}. \href{http://arxiv.org/abs/1207.4708}{{\tt
  arXiv:1207.4708}}.
\bibitem[{Aytar et~al.(2018)Aytar, Pfaff, Budden, {Le Paine}, Wang, and {De
  Freitas}}]{Aytar2018a}
\bibinfo{author}{Y.~Aytar}, \bibinfo{author}{T.~Pfaff},
  \bibinfo{author}{D.~Budden}, \bibinfo{author}{T.~{Le Paine}},
  \bibinfo{author}{Z.~Wang}, \bibinfo{author}{N.~{De Freitas}},
\newblock \bibinfo{title}{{Playing Hard Exploration Games by Watching
  Youtube}},
\newblock \bibinfo{journal}{Conference on Neural Information Processing
  Systems, NeurIPS 2018}  (\bibinfo{year}{2018}) \bibinfo{pages}{2930--2941}.
  \href{http://arxiv.org/abs/1805.11592}{{\tt arXiv:1805.11592}}.
\bibitem[{Kempka et~al.(2016)Kempka, Wydmuch, Runc, Toczek, and
  Jaskowski}]{Kempka2016}
\bibinfo{author}{M.~Kempka}, \bibinfo{author}{M.~Wydmuch},
  \bibinfo{author}{G.~Runc}, \bibinfo{author}{J.~Toczek},
  \bibinfo{author}{W.~Jaskowski},
\newblock \bibinfo{title}{{ViZDoom: A Doom-based AI Research Platform for
  Visual Reinforcement Learning}},
\newblock \bibinfo{journal}{IEEE Conference on Computatonal Intelligence and
  Games, CIG} \bibinfo{volume}{0} (\bibinfo{year}{2016}).
  \DOIprefix\doi{10.1109/CIG.2016.7860433}.
  \href{http://arxiv.org/abs/1605.02097}{{\tt arXiv:1605.02097}}.
\bibitem[{Johnson et~al.(2016)Johnson, Hofmann, Hutton, and
  Bignell}]{Johnson2016a}
\bibinfo{author}{M.~Johnson}, \bibinfo{author}{K.~Hofmann},
  \bibinfo{author}{T.~Hutton}, \bibinfo{author}{D.~Bignell},
\newblock \bibinfo{title}{{The Malmo Platform For Artificial Intelligence
  Experimentation}},
\newblock \bibinfo{journal}{IJCAI International Joint Conference on Artificial
  Intelligence} \bibinfo{volume}{2016-Janua} (\bibinfo{year}{2016})
  \bibinfo{pages}{4246--4247}.
\bibitem[{Todorov et~al.(2012)Todorov, Erez, and Tassa}]{Todorov2012}
\bibinfo{author}{E.~Todorov}, \bibinfo{author}{T.~Erez},
  \bibinfo{author}{Y.~Tassa},
\newblock \bibinfo{title}{{MuJoCo: A Physics Engine for Model-based Control}},
\newblock \bibinfo{journal}{IEEE International Conference on Intelligent Robots
  and Systems}  (\bibinfo{year}{2012}) \bibinfo{pages}{5026--5033}.
  \DOIprefix\doi{10.1109/IROS.2012.6386109}.
\bibitem[{Schmidhuber(2010)}]{Schmidhuber2010b}
\bibinfo{author}{J.~Schmidhuber},
\newblock \bibinfo{title}{{Formal Theory Of Creativity, Fun, and Intrinsic
  Motivation (1990-2010)}},
\newblock \bibinfo{journal}{IEEE Transactions on Autonomous Mental Development}
  \bibinfo{volume}{2} (\bibinfo{year}{2010}) \bibinfo{pages}{230--247}.
  \DOIprefix\doi{10.1109/TAMD.2010.2056368}.
  \href{http://arxiv.org/abs/1510.05840}{{\tt arXiv:1510.05840}}.
\bibitem[{Ecoffet et~al.(2019)Ecoffet, Huizinga, Lehman, Stanley, and
  Clune}]{Ecoffet2019}
\bibinfo{author}{A.~Ecoffet}, \bibinfo{author}{J.~Huizinga},
  \bibinfo{author}{J.~Lehman}, \bibinfo{author}{K.~O. Stanley},
  \bibinfo{author}{J.~Clune},
\newblock \bibinfo{title}{{Go-Explore: a New Approach for Hard-Exploration
  Problems}}  (\bibinfo{year}{2019}) \bibinfo{pages}{1--37}. \URLprefix
  \url{http://arxiv.org/abs/1901.10995}.
  \href{http://arxiv.org/abs/1901.10995}{{\tt arXiv:1901.10995}}.
\bibitem[{Oudeyer and Kaplan(2007)}]{Oudeyer2007b}
\bibinfo{author}{P.-Y. Oudeyer}, \bibinfo{author}{F.~Kaplan},
\newblock \bibinfo{title}{{What Is Intrinsic Motivation? A Typology Of
  Computational Approaches}},
\newblock \bibinfo{journal}{Frontiers in Neurorobotics} \bibinfo{volume}{1}
  (\bibinfo{year}{2007}) \bibinfo{pages}{1184--1191}. \URLprefix
  \url{http://journal.frontiersin.org/article/10.3389/neuro.12.006.2007/abstract}.
  \DOIprefix\doi{10.3389/neuro.12.006.2007}.
  \href{http://arxiv.org/abs/arXiv:1410.5401v2}{{\tt arXiv:arXiv:1410.5401v2}}.
\bibitem[{Achiam and Sastry(2017)}]{Achiam2017}
\bibinfo{author}{J.~Achiam}, \bibinfo{author}{S.~Sastry},
\newblock \bibinfo{title}{{Surprise-Based Intrinsic Motivation for Deep
  Reinforcement Learning}}  (\bibinfo{year}{2017}) \bibinfo{pages}{1--13}.
  \URLprefix \url{http://arxiv.org/abs/1703.01732}.
  \href{http://arxiv.org/abs/1703.01732}{{\tt arXiv:1703.01732}}.
\bibitem[{Li et~al.(2019)Li, Lu, Li, Lu, Cai, and Wang}]{Li2019a}
\bibinfo{author}{B.~Li}, \bibinfo{author}{T.~Lu}, \bibinfo{author}{J.~Li},
  \bibinfo{author}{N.~Lu}, \bibinfo{author}{Y.~Cai}, \bibinfo{author}{S.~Wang},
\newblock \bibinfo{title}{{Curiosity-driven Exploration for Off-policy
  Reinforcement Learning Methods}},
\newblock \bibinfo{journal}{IEEE International Conference on Robotics and
  Biomimetics, ROBIO 2019}  (\bibinfo{year}{2019}) \bibinfo{pages}{1109--1114}.
  \DOIprefix\doi{10.1109/ROBIO49542.2019.8961529}.
\bibitem[{Goodfellow et~al.(2014)Goodfellow, Pouget-Abadie, Mirza, Xu,
  Warde-Farley, Ozair, Courville, and Bengio}]{NIPS2014_5ca3e9b1}
\bibinfo{author}{I.~Goodfellow}, \bibinfo{author}{J.~Pouget-Abadie},
  \bibinfo{author}{M.~Mirza}, \bibinfo{author}{B.~Xu},
  \bibinfo{author}{D.~Warde-Farley}, \bibinfo{author}{S.~Ozair},
  \bibinfo{author}{A.~Courville}, \bibinfo{author}{Y.~Bengio},
\newblock \bibinfo{title}{Generative adversarial nets},
\newblock \bibinfo{journal}{Conference on Neural Information Processing
  Systems, NeurIPS 2014} \bibinfo{volume}{27} (\bibinfo{year}{2014}).
  \URLprefix
  \url{https://proceedings.neurips.cc/paper/2014/file/5ca3e9b122f61f8f06494c97b1afccf3-Paper.pdf}.
\bibitem[{Hong et~al.(2019)Hong, Zhu, Liu, Zhang, Zhou, Yu, and Sun}]{Hong2019}
\bibinfo{author}{W.~Hong}, \bibinfo{author}{M.~Zhu}, \bibinfo{author}{M.~Liu},
  \bibinfo{author}{W.~Zhang}, \bibinfo{author}{M.~Zhou},
  \bibinfo{author}{Y.~Yu}, \bibinfo{author}{P.~Sun},
\newblock \bibinfo{title}{{Generative Adversarial Exploration for Reinforcement
  Learning}},
\newblock \bibinfo{journal}{ACM International Conference Proceeding Series}
  (\bibinfo{year}{2019}). \DOIprefix\doi{10.1145/3356464.3357706}.
\bibitem[{Stadie et~al.(2015)Stadie, Levine, and Abbeel}]{Stadie2015}
\bibinfo{author}{B.~C. Stadie}, \bibinfo{author}{S.~Levine},
  \bibinfo{author}{P.~Abbeel},
\newblock \bibinfo{title}{{Incentivizing Exploration In Reinforcement Learning
  With Deep Predictive Models}}  (\bibinfo{year}{2015}) \bibinfo{pages}{1--11}.
  \URLprefix \url{http://arxiv.org/abs/1507.00814}.
  \href{http://arxiv.org/abs/1507.00814}{{\tt arXiv:1507.00814}}.
\bibitem[{Bougie and Ichise(2020{\natexlab{a}})}]{Bougie2020c}
\bibinfo{author}{N.~Bougie}, \bibinfo{author}{R.~Ichise},
\newblock \bibinfo{title}{{Fast And Slow Curiosity for High-Level Exploration
  in Reinforcement Learning}},
\newblock \bibinfo{journal}{Applied Intelligence}
  (\bibinfo{year}{2020}{\natexlab{a}}).
  \DOIprefix\doi{10.1007/s10489-020-01849-3}.
\bibitem[{Bougie and Ichise(2020{\natexlab{b}})}]{Bougie2020}
\bibinfo{author}{N.~Bougie}, \bibinfo{author}{R.~Ichise},
\newblock \bibinfo{title}{{Towards High-Level Intrinsic Exploration in
  Reinforcement Learning}},
\newblock \bibinfo{journal}{International Joint Conference on Artificial
  Intelligence (IJCAI-20)}  (\bibinfo{year}{2020}{\natexlab{b}}).
  \href{http://arxiv.org/abs/1810.12894}{{\tt arXiv:1810.12894}}.
\bibitem[{Osband et~al.(2018)Osband, Aslanides, and Cassirer}]{Osband2018}
\bibinfo{author}{I.~Osband}, \bibinfo{author}{J.~Aslanides},
  \bibinfo{author}{A.~Cassirer},
\newblock \bibinfo{title}{{Randomized Prior Functions for Deep Reinforcement
  Learning}},
\newblock \bibinfo{journal}{Conference on Neural Information Processing Systems
  (NeurIPS 2018)}  (\bibinfo{year}{2018}).
\bibitem[{Pathak et~al.(2017)Pathak, Agrawal, Efros, and Darrell}]{Pathak2017b}
\bibinfo{author}{D.~Pathak}, \bibinfo{author}{P.~Agrawal},
  \bibinfo{author}{A.~A. Efros}, \bibinfo{author}{T.~Darrell},
\newblock \bibinfo{title}{{Curiosity-Driven Exploration by Self-Supervised
  Prediction}},
\newblock \bibinfo{journal}{Proceedings of the 34th International Conference on
  Machine Learning}  (\bibinfo{year}{2017}) \bibinfo{pages}{488--489}.
  \DOIprefix\doi{10.1109/CVPRW.2017.70}.
  \href{http://arxiv.org/abs/1705.05363}{{\tt arXiv:1705.05363}}.
\bibitem[{Burda et~al.(2019)Burda, Edwards, Pathak, Storkey, Darrell, and
  Efros}]{Burda2018c}
\bibinfo{author}{Y.~Burda}, \bibinfo{author}{H.~Edwards},
  \bibinfo{author}{D.~Pathak}, \bibinfo{author}{A.~Storkey},
  \bibinfo{author}{T.~Darrell}, \bibinfo{author}{A.~A. Efros},
\newblock \bibinfo{title}{{Large-Scale Study of Curiosity-Driven Learning}},
\newblock \bibinfo{journal}{7th International Conference on Learning
  Representations, ICLR 2019}  (\bibinfo{year}{2019}). \URLprefix
  \url{http://arxiv.org/abs/1808.04355}.
  \href{http://arxiv.org/abs/1808.04355}{{\tt arXiv:1808.04355}}.
\bibitem[{Raileanu and Rockt{\"{a}}schel(2020)}]{Raileanu2020}
\bibinfo{author}{R.~Raileanu}, \bibinfo{author}{T.~Rockt{\"{a}}schel},
\newblock \bibinfo{title}{{RIDE: Rewarding Impact-Driven Exploration for
  Procedurally-Generated Environments}},
\newblock \bibinfo{journal}{8th International Conference on Learning
  Representations, ICLR 2020}  (\bibinfo{year}{2020}). \URLprefix
  \url{http://arxiv.org/abs/2002.12292}.
  \href{http://arxiv.org/abs/2002.12292}{{\tt arXiv:2002.12292}}.
\bibitem[{Li et~al.(2020)Li, Shi, Li, Zhang, and Wang}]{Li2020a}
\bibinfo{author}{J.~Li}, \bibinfo{author}{X.~Shi}, \bibinfo{author}{J.~Li},
  \bibinfo{author}{X.~Zhang}, \bibinfo{author}{J.~Wang},
\newblock \bibinfo{title}{{Random Curiosity-Driven Exploration in Deep
  Reinforcement Learning}},
\newblock \bibinfo{journal}{Neurocomputing} \bibinfo{volume}{418}
  (\bibinfo{year}{2020}) \bibinfo{pages}{139--147}. \URLprefix
  \url{https://doi.org/10.1016/j.neucom.2020.08.024}.
  \DOIprefix\doi{10.1016/j.neucom.2020.08.024}.
\bibitem[{Pathak et~al.(2019)Pathak, Gandhi, and Gupta}]{Pathak2019}
\bibinfo{author}{D.~Pathak}, \bibinfo{author}{D.~Gandhi},
  \bibinfo{author}{A.~Gupta},
\newblock \bibinfo{title}{{Self-Supervised Exploration via Disagreement}},
\newblock \bibinfo{journal}{Proceedings of the 36th International Conference on
  Machine Learning}  (\bibinfo{year}{2019}).
\bibitem[{Kim et~al.(2019{\natexlab{a}})Kim, Kim, Jeong, Levine, and
  Song}]{Kim2019c}
\bibinfo{author}{H.~Kim}, \bibinfo{author}{J.~Kim}, \bibinfo{author}{Y.~Jeong},
  \bibinfo{author}{S.~Levine}, \bibinfo{author}{H.~O. Song},
\newblock \bibinfo{title}{{EMI: Exploration with Mutual Information}},
\newblock \bibinfo{journal}{36th International Conference on Machine Learning,
  ICML 2019}  (\bibinfo{year}{2019}{\natexlab{a}}) \bibinfo{pages}{5837--5851}.
  \href{http://arxiv.org/abs/1810.01176}{{\tt arXiv:1810.01176}}.
\bibitem[{Kim et~al.(2019{\natexlab{b}})Kim, Nam, Kim, Kim, and Kim}]{Kim2019b}
\bibinfo{author}{Y.~Kim}, \bibinfo{author}{W.~Nam}, \bibinfo{author}{H.~Kim},
  \bibinfo{author}{J.~H. Kim}, \bibinfo{author}{G.~Kim},
\newblock \bibinfo{title}{{Curiosity-Bottleneck: Exploration by Distilling
  Task-Specific Novelty}},
\newblock \bibinfo{journal}{36th International Conference on Machine Learning,
  ICML 2019}  (\bibinfo{year}{2019}{\natexlab{b}}) \bibinfo{pages}{5861--5874}.
\bibitem[{Mirowski et~al.(2017)Mirowski, Pascanu, Viola, Soyer, Ballard,
  Banino, Denil, Goroshin, Sifre, Kavukcuoglu, Kumaran, and
  Hadsell}]{Mirowski2017}
\bibinfo{author}{P.~Mirowski}, \bibinfo{author}{R.~Pascanu},
  \bibinfo{author}{F.~Viola}, \bibinfo{author}{H.~Soyer},
  \bibinfo{author}{A.~J. Ballard}, \bibinfo{author}{A.~Banino},
  \bibinfo{author}{M.~Denil}, \bibinfo{author}{R.~Goroshin},
  \bibinfo{author}{L.~Sifre}, \bibinfo{author}{K.~Kavukcuoglu},
  \bibinfo{author}{D.~Kumaran}, \bibinfo{author}{R.~Hadsell},
\newblock \bibinfo{title}{{Learning to Navigate in Complex Environments}},
\newblock \bibinfo{journal}{5th International Conference on Learning
  Representations, ICLR 2017}  (\bibinfo{year}{2017}).
  \href{http://arxiv.org/abs/1611.03673}{{\tt arXiv:1611.03673}}.
\bibitem[{Dhiman et~al.(2018)Dhiman, Banerjee, Griffin, Siskind, and
  Corso}]{Dhiman2018a}
\bibinfo{author}{V.~Dhiman}, \bibinfo{author}{S.~Banerjee},
  \bibinfo{author}{B.~Griffin}, \bibinfo{author}{J.~M. Siskind},
  \bibinfo{author}{J.~J. Corso},
\newblock \bibinfo{title}{{A Critical Investigation of Deep Reinforcement
  Learning for Navigation}}  (\bibinfo{year}{2018}). \URLprefix
  \url{http://arxiv.org/abs/1802.02274}.
  \href{http://arxiv.org/abs/1802.02274}{{\tt arXiv:1802.02274}}.
\bibitem[{Li et~al.(2020)Li, Lu, Li, Lu, Cai, and Wang}]{Li2020}
\bibinfo{author}{B.~Li}, \bibinfo{author}{T.~Lu}, \bibinfo{author}{J.~Li},
  \bibinfo{author}{N.~Lu}, \bibinfo{author}{Y.~Cai}, \bibinfo{author}{S.~Wang},
\newblock \bibinfo{title}{{ACDER: Augmented Curiosity-Driven Experience
  Replay}},
\newblock \bibinfo{journal}{IEEE International Conference on Robotics and
  Automation, ICRA 2020}  (\bibinfo{year}{2020}) \bibinfo{pages}{4218--4224}.
  \DOIprefix\doi{10.1109/ICRA40945.2020.9197421}.
\bibitem[{Stanton and Clune(2018)}]{Stanton2018}
\bibinfo{author}{C.~Stanton}, \bibinfo{author}{J.~Clune}, \bibinfo{title}{{Deep
  Curiosity Search: Intra-life Exploration Can Improve Performance on
  Challenging Deep Reinforcement Learning Problems}}, \bibinfo{year}{2018}.
  \URLprefix \url{http://arxiv.org/abs/1806.00553}.
  \href{http://arxiv.org/abs/1806.00553}{{\tt arXiv:1806.00553}}.
\bibitem[{Dean et~al.(2020)Dean, Tulsiani, and Gupta}]{Dean2020}
\bibinfo{author}{V.~Dean}, \bibinfo{author}{S.~Tulsiani},
  \bibinfo{author}{A.~Gupta}, \bibinfo{title}{{See, Hear, Explore: Curiosity
  via Audio-Visual Association}}, \bibinfo{year}{2020}. \URLprefix
  \url{http://arxiv.org/abs/2007.03669}.
  \href{http://arxiv.org/abs/2007.03669}{{\tt arXiv:2007.03669}}.
\bibitem[{Kolter and Ng(2009)}]{Kolter2009a}
\bibinfo{author}{J.~Z. Kolter}, \bibinfo{author}{A.~Y. Ng},
\newblock \bibinfo{title}{{Near-Bayesian Exploration in Polynomial Time}}
  (\bibinfo{year}{2009}) \bibinfo{pages}{513--520}.
\bibitem[{Still and Precup(2012)}]{Still2012a}
\bibinfo{author}{S.~Still}, \bibinfo{author}{D.~Precup},
\newblock \bibinfo{title}{{An Information-Theoretic Approach to
  Curiosity-Driven Reinforcement Learning}},
\newblock \bibinfo{journal}{Theory in Biosciences} \bibinfo{volume}{131}
  (\bibinfo{year}{2012}) \bibinfo{pages}{139--148}.
  \DOIprefix\doi{10.1007/s12064-011-0142-z}.
\bibitem[{Still(2009)}]{Still}
\bibinfo{author}{S.~Still},
\newblock \bibinfo{title}{{Information Theoretic Approach to Interactive
  Learning}},
\newblock \bibinfo{journal}{Arxiv}  (\bibinfo{year}{2009})
  \bibinfo{pages}{1--6}.
\bibitem[{Houthooft et~al.(2016)Houthooft, Chen, Duan, Schulman, {De Turck},
  and Abbeel}]{Houthooft2016b}
\bibinfo{author}{R.~Houthooft}, \bibinfo{author}{X.~Chen},
  \bibinfo{author}{Y.~Duan}, \bibinfo{author}{J.~Schulman},
  \bibinfo{author}{F.~{De Turck}}, \bibinfo{author}{P.~Abbeel},
\newblock \bibinfo{title}{{VIME: Variational Information Maximizing
  Exploration}},
\newblock \bibinfo{journal}{Conference on Neural Information Processing
  Systems, NeurIPS 2016} \bibinfo{volume}{0} (\bibinfo{year}{2016})
  \bibinfo{pages}{1117--1125}. \href{http://arxiv.org/abs/1605.09674}{{\tt
  arXiv:1605.09674}}.
\bibitem[{Mohamed and Rezende(2015)}]{Mohamed2015a}
\bibinfo{author}{S.~Mohamed}, \bibinfo{author}{D.~J. Rezende},
\newblock \bibinfo{title}{{Variational Information Maximisation for
  Intrinsically Motivated Reinforcement Learning}},
\newblock \bibinfo{journal}{Conference on Neural Information Processing
  Systems, NeurIPS 2015}  (\bibinfo{year}{2015}) \bibinfo{pages}{2125--2133}.
  \href{http://arxiv.org/abs/1509.08731}{{\tt arXiv:1509.08731}}.
\bibitem[{{De Abril} and Kanai(2018)}]{DeAbril2018a}
\bibinfo{author}{I.~M. {De Abril}}, \bibinfo{author}{R.~Kanai},
\newblock \bibinfo{title}{{Curiosity-Driven Reinforcement Learning with
  Homeostatic Regulation}},
\newblock \bibinfo{journal}{Proceedings of the International Joint Conference
  on Neural Networks} \bibinfo{volume}{2018-July} (\bibinfo{year}{2018}).
  \DOIprefix\doi{10.1109/IJCNN.2018.8489075}.
  \href{http://arxiv.org/abs/1801.07440}{{\tt arXiv:1801.07440}}.
\bibitem[{Chien and Hsu(2020)}]{Chien2020}
\bibinfo{author}{J.-T. Chien}, \bibinfo{author}{P.-C. Hsu},
\newblock \bibinfo{title}{{Stochastic Curiosity Maximizing Exploration}},
\newblock \bibinfo{journal}{2020 International Joint Conference on Neural
  Networks (IJCNN)}  (\bibinfo{year}{2020}).
  \DOIprefix\doi{10.1109/ijcnn48605.2020.9207295}.
\bibitem[{Savinov et~al.(2019)Savinov, Raichuk, Marinier, Vincent, Pollefeys,
  Lillicrap, and Gelly}]{Savinov2019}
\bibinfo{author}{N.~Savinov}, \bibinfo{author}{A.~Raichuk},
  \bibinfo{author}{R.~Marinier}, \bibinfo{author}{D.~Vincent},
  \bibinfo{author}{M.~Pollefeys}, \bibinfo{author}{T.~Lillicrap},
  \bibinfo{author}{S.~Gelly},
\newblock \bibinfo{title}{{Episodic Curiosity Through Reachability}},
\newblock \bibinfo{journal}{7th International Conference on Learning
  Representations, ICLR 2019}  (\bibinfo{year}{2019}) \bibinfo{pages}{1--20}.
  \href{http://arxiv.org/abs/1810.02274}{{\tt arXiv:1810.02274}}.
\bibitem[{M{\'{e}}nard et~al.(2020)M{\'{e}}nard, Domingues, Jonsson, Kaufmann,
  Leurent, and Valko}]{Menard2020a}
\bibinfo{author}{P.~M{\'{e}}nard}, \bibinfo{author}{O.~D. Domingues},
  \bibinfo{author}{A.~Jonsson}, \bibinfo{author}{E.~Kaufmann},
  \bibinfo{author}{E.~Leurent}, \bibinfo{author}{M.~Valko},
\newblock \bibinfo{title}{{Fast Active Learning for Pure Exploration in
  Reinforcement Learning}}  (\bibinfo{year}{2020}) \bibinfo{pages}{1--36}.
  \URLprefix \url{http://arxiv.org/abs/2007.13442}.
  \href{http://arxiv.org/abs/2007.13442}{{\tt arXiv:2007.13442}}.
\bibitem[{Tang et~al.(2017)Tang, Houthooft, Foote, Stooke, Chen, Duan,
  Schulman, {De Turck}, and Abbeel}]{Tang2017a}
\bibinfo{author}{H.~Tang}, \bibinfo{author}{R.~Houthooft},
  \bibinfo{author}{D.~Foote}, \bibinfo{author}{A.~Stooke},
  \bibinfo{author}{X.~Chen}, \bibinfo{author}{Y.~Duan},
  \bibinfo{author}{J.~Schulman}, \bibinfo{author}{F.~{De Turck}},
  \bibinfo{author}{P.~Abbeel},
\newblock \bibinfo{title}{{Exploration: A Study Of Count-Based Exploration For
  Deep Reinforcement Learning}},
\newblock \bibinfo{journal}{Conference on Neural Information Processing
  Systems, NeurIPS 2017}  (\bibinfo{year}{2017}) \bibinfo{pages}{2754--2763}.
  \href{http://arxiv.org/abs/1611.04717}{{\tt arXiv:1611.04717}}.
\bibitem[{Charikar(2002)}]{Charikar2002}
\bibinfo{author}{M.~S. Charikar},
\newblock \bibinfo{title}{{Similarity Estimation Techniques From Rounding
  Algorithms}},
\newblock \bibinfo{journal}{Conference Proceedings of the Annual ACM Symposium
  on Theory of Computing}  (\bibinfo{year}{2002}) \bibinfo{pages}{380--388}.
  \DOIprefix\doi{10.1145/509907.509965}.
\bibitem[{Choi et~al.(2019)Choi, Guo, Moczulski, Oh, Wu, Norouzi, and
  Lee}]{Choi2019a}
\bibinfo{author}{J.~Choi}, \bibinfo{author}{Y.~Guo},
  \bibinfo{author}{M.~Moczulski}, \bibinfo{author}{J.~Oh},
  \bibinfo{author}{N.~Wu}, \bibinfo{author}{M.~Norouzi},
  \bibinfo{author}{H.~Lee},
\newblock \bibinfo{title}{{Contingency-aware Exploration in Reinforcement
  Learning}},
\newblock \bibinfo{journal}{7th International Conference on Learning
  Representations, ICLR 2019}  (\bibinfo{year}{2019}) \bibinfo{pages}{1--19}.
  \href{http://arxiv.org/abs/1811.01483}{{\tt arXiv:1811.01483}}.
\bibitem[{Machado et~al.(2020)Machado, Bellemare, and Bowling}]{Machado2020a}
\bibinfo{author}{M.~C. Machado}, \bibinfo{author}{M.~G. Bellemare},
  \bibinfo{author}{M.~Bowling},
\newblock \bibinfo{title}{{Count-Based Exploration with the Successor
  Representation}},
\newblock \bibinfo{journal}{AAAI Conference on Artificial Intelligence}
  (\bibinfo{year}{2020}). \URLprefix \url{http://arxiv.org/abs/1807.11622}.
  \DOIprefix\doi{10.1609/aaai.v34i04.5955}.
  \href{http://arxiv.org/abs/1807.11622}{{\tt arXiv:1807.11622}}.
\bibitem[{Zhao and Tresp(2019)}]{Zhao2019}
\bibinfo{author}{R.~Zhao}, \bibinfo{author}{V.~Tresp},
\newblock \bibinfo{title}{{Curiosity-Driven Experience Prioritization via
  density estimation}},
\newblock \bibinfo{journal}{arXiv}  (\bibinfo{year}{2019}).
  \href{http://arxiv.org/abs/1902.08039}{{\tt arXiv:1902.08039}}.
\bibitem[{Ostrovski et~al.(2017)Ostrovski, Bellemare, {Van Den Oord}, and
  Munos}]{Ostrovski2017a}
\bibinfo{author}{G.~Ostrovski}, \bibinfo{author}{M.~G. Bellemare},
  \bibinfo{author}{A.~{Van Den Oord}}, \bibinfo{author}{R.~Munos},
\newblock \bibinfo{title}{{Count-based Exploration with Neural Density
  Models}},
\newblock \bibinfo{journal}{34th International Conference on Machine Learning,
  ICML 2017} \bibinfo{volume}{6} (\bibinfo{year}{2017})
  \bibinfo{pages}{4161--4175}. \href{http://arxiv.org/abs/1703.01310}{{\tt
  arXiv:1703.01310}}.
\bibitem[{Martin et~al.(2017)Martin, Narayanan, Everitt, and
  Hutter}]{Martin2017a}
\bibinfo{author}{J.~Martin}, \bibinfo{author}{S.~S. Narayanan},
  \bibinfo{author}{T.~Everitt}, \bibinfo{author}{M.~Hutter},
\newblock \bibinfo{title}{{Count-based exploration in feature space for
  reinforcement learning}},
\newblock \bibinfo{journal}{IJCAI International Joint Conference on Artificial
  Intelligence}  (\bibinfo{year}{2017}) \bibinfo{pages}{2471--2478}.
  \DOIprefix\doi{10.24963/ijcai.2017/344}.
  \href{http://arxiv.org/abs/arXiv:1706.08090v1}{{\tt
  arXiv:arXiv:1706.08090v1}}.
\bibitem[{Malisiewicz et~al.(2011)Malisiewicz, Gupta, and
  Efros}]{Malisiewicz2011}
\bibinfo{author}{T.~Malisiewicz}, \bibinfo{author}{A.~Gupta},
  \bibinfo{author}{A.~A. Efros},
\newblock \bibinfo{title}{{Ensemble of Exemplar-SVMs for Object Detection and
  Beyond}},
\newblock \bibinfo{journal}{Proceedings of the IEEE International Conference on
  Computer Vision}  (\bibinfo{year}{2011}) \bibinfo{pages}{89--96}.
  \DOIprefix\doi{10.1109/ICCV.2011.6126229}.
\bibitem[{Fu et~al.(2017)Fu, Co-Reyes, and Levine}]{Fu2017}
\bibinfo{author}{J.~Fu}, \bibinfo{author}{J.~D. Co-Reyes},
  \bibinfo{author}{S.~Levine},
\newblock \bibinfo{title}{{Ex2: Exploration With Exemplar Models for Deep
  Reinforcement Learning}},
\newblock \bibinfo{journal}{Conference on Neural Information Processing
  Systems, NeurIPS 2017}  (\bibinfo{year}{2017}) \bibinfo{pages}{2578--2588}.
  \href{http://arxiv.org/abs/1703.01260}{{\tt arXiv:1703.01260}}.
\bibitem[{Badia et~al.(2020)Badia, Sprechmann, Vitvitskyi, Guo, Piot,
  Kapturowski, Tieleman, Arjovsky, Pritzel, Bolt, and Blundell}]{Badia2020c}
\bibinfo{author}{A.~P. Badia}, \bibinfo{author}{P.~Sprechmann},
  \bibinfo{author}{A.~Vitvitskyi}, \bibinfo{author}{D.~Guo},
  \bibinfo{author}{B.~Piot}, \bibinfo{author}{S.~Kapturowski},
  \bibinfo{author}{O.~Tieleman}, \bibinfo{author}{M.~Arjovsky},
  \bibinfo{author}{A.~Pritzel}, \bibinfo{author}{A.~Bolt},
  \bibinfo{author}{C.~Blundell},
\newblock \bibinfo{title}{{Never Give Up: Learning Directed Exploration
  Strategies}},
\newblock \bibinfo{journal}{8th International Conference on Learning
  Representations, ICLR 2020}  (\bibinfo{year}{2020}). \URLprefix
  \url{http://arxiv.org/abs/2002.06038}.
  \href{http://arxiv.org/abs/2002.06038}{{\tt arXiv:2002.06038}}.
\bibitem[{Such et~al.(2017)Such, Madhavan, Conti, Lehman, Stanley, and
  Clune}]{Such2017}
\bibinfo{author}{F.~P. Such}, \bibinfo{author}{V.~Madhavan},
  \bibinfo{author}{E.~Conti}, \bibinfo{author}{J.~Lehman},
  \bibinfo{author}{K.~O. Stanley}, \bibinfo{author}{J.~Clune},
\newblock \bibinfo{title}{{Deep Neuroevolution: Genetic Algorithms Are a
  Competitive Alternative for Training Deep Neural Networks for Reinforcement
  Learning}}  (\bibinfo{year}{2017}). \URLprefix
  \url{http://arxiv.org/abs/1712.06567}. \DOIprefix\doi{1712.06567}.
  \href{http://arxiv.org/abs/1712.06567}{{\tt arXiv:1712.06567}}.
\bibitem[{Salimans et~al.(2017)Salimans, Ho, Chen, Sidor, and
  Sutskever}]{Salimans2017a}
\bibinfo{author}{T.~Salimans}, \bibinfo{author}{J.~Ho},
  \bibinfo{author}{X.~Chen}, \bibinfo{author}{S.~Sidor},
  \bibinfo{author}{I.~Sutskever},
\newblock \bibinfo{title}{{Evolution Strategies as a Scalable Alternative to
  Reinforcement Learning}},
\newblock \bibinfo{journal}{arXiv}  (\bibinfo{year}{2017})
  \bibinfo{pages}{476--485}. \DOIprefix\doi{10.1109/ICSTW.2011.58}.
  \href{http://arxiv.org/abs/1703.03864v2}{{\tt arXiv:1703.03864v2}}.
\bibitem[{Lehman and Stanley(2011)}]{Lehman2011}
\bibinfo{author}{J.~Lehman}, \bibinfo{author}{K.~O. Stanley},
\newblock \bibinfo{title}{{Abandoning Objectives: Evolution Through the Search
  for Novelty Alone}},
\newblock \bibinfo{journal}{Evolutionary Computation} \bibinfo{volume}{19}
  (\bibinfo{year}{2011}) \bibinfo{pages}{189--222}.
  \DOIprefix\doi{10.1162/EVCO_a_00025}.
\bibitem[{Risi et~al.(2009)Risi, Vanderbleek, Hughes, and Stanley}]{Risi2009}
\bibinfo{author}{S.~Risi}, \bibinfo{author}{S.~D. Vanderbleek},
  \bibinfo{author}{C.~E. Hughes}, \bibinfo{author}{K.~O. Stanley},
\newblock \bibinfo{title}{{How Novelty Search Escapes the Deceptive Trap of
  Learning to Learn}},
\newblock \bibinfo{journal}{Proceedings of the 11th Annual conference on
  Genetic and evolutionary computation - GECCO '09}  (\bibinfo{year}{2009})
  \bibinfo{pages}{153}. \URLprefix
  \url{http://portal.acm.org/citation.cfm?doid=1569901.1569923}.
  \DOIprefix\doi{10.1145/1569901.1569923}.
\bibitem[{Conti et~al.(2018)Conti, Madhavan, Such, Lehman, Stanley, and
  Clune}]{Conti2018a}
\bibinfo{author}{E.~Conti}, \bibinfo{author}{V.~Madhavan},
  \bibinfo{author}{F.~P. Such}, \bibinfo{author}{J.~Lehman},
  \bibinfo{author}{K.~O. Stanley}, \bibinfo{author}{J.~Clune},
\newblock \bibinfo{title}{Improving exploration in evolution strategies for
  deep reinforcement learning via a population of novelty-seeking agents},
\newblock \bibinfo{journal}{Conference on Neural Information Processing
  Systems, NeurIPS 2018}  (\bibinfo{year}{2018}) \bibinfo{pages}{5027--5038}.
  \href{http://arxiv.org/abs/1712.06560}{{\tt arXiv:1712.06560}}.
\bibitem[{Gravina et~al.(2018)Gravina, Liapis, and Yannakakis}]{Gravina2018}
\bibinfo{author}{D.~Gravina}, \bibinfo{author}{A.~Liapis},
  \bibinfo{author}{G.~N. Yannakakis},
\newblock \bibinfo{title}{{Quality Diversity Through Surprise}},
\newblock \bibinfo{journal}{IEEE Transactions on Evolutionary Computation}
  (\bibinfo{year}{2018}) \bibinfo{pages}{1--14}.
  \DOIprefix\doi{10.1109/TEVC.2018.2877215}.
  \href{http://arxiv.org/abs/1807.02397}{{\tt arXiv:1807.02397}}.
\bibitem[{Mouret and Doncieux(2012)}]{Mouret2012}
\bibinfo{author}{J.~B. Mouret}, \bibinfo{author}{S.~Doncieux},
\newblock \bibinfo{title}{{Encouraging Behavioral Diversity In Evolutionary
  Robotics: An Empirical Study}},
\newblock \bibinfo{journal}{Evolutionary Computation} \bibinfo{volume}{20}
  (\bibinfo{year}{2012}) \bibinfo{pages}{91--133}.
  \DOIprefix\doi{10.1162/EVCO_a_00048}.
\bibitem[{Pugh et~al.(2016)Pugh, Soros, and Stanley}]{Pugh2016}
\bibinfo{author}{J.~K. Pugh}, \bibinfo{author}{L.~B. Soros},
  \bibinfo{author}{K.~O. Stanley},
\newblock \bibinfo{title}{{Quality Diversity: A New Frontier for Evolutionary
  Computation}},
\newblock \bibinfo{journal}{Frontiers in Robotics and AI} \bibinfo{volume}{3}
  (\bibinfo{year}{2016}). \URLprefix
  \url{http://journal.frontiersin.org/Article/10.3389/frobt.2016.00040/abstract}.
  \DOIprefix\doi{10.3389/frobt.2016.00040}.
\bibitem[{Hong et~al.(2018)Hong, Shann, Su, Chang, and Lee}]{Hong2018}
\bibinfo{author}{Z.~W. Hong}, \bibinfo{author}{T.~Y. Shann},
  \bibinfo{author}{S.~Y. Su}, \bibinfo{author}{Y.~H. Chang},
  \bibinfo{author}{C.~Y. Lee},
\newblock \bibinfo{title}{{Diversity-Driven Exploration Strategy for Deep
  Reinforcement Learning}},
\newblock \bibinfo{journal}{6th International Conference on Learning
  Representations, ICLR 2018 - Workshop Track Proceedings}
  (\bibinfo{year}{2018}).
\bibitem[{Cohen et~al.(2019)Cohen, Yu, Qiao, and Tong}]{Cohen2019}
\bibinfo{author}{A.~Cohen}, \bibinfo{author}{L.~Yu}, \bibinfo{author}{X.~Qiao},
  \bibinfo{author}{X.~Tong},
\newblock \bibinfo{title}{{Maximum Entropy Diverse Exploration: Disentangling
  Maximum Entropy Reinforcement Learning}}  (\bibinfo{year}{2019}). \URLprefix
  \url{http://arxiv.org/abs/1911.00828}.
  \href{http://arxiv.org/abs/1911.00828}{{\tt arXiv:1911.00828}}.
\bibitem[{Pong et~al.(2019)Pong, Dalal, Lin, Nair, Bahl, and Levine}]{Pong2019}
\bibinfo{author}{V.~H. Pong}, \bibinfo{author}{M.~Dalal},
  \bibinfo{author}{S.~Lin}, \bibinfo{author}{A.~Nair},
  \bibinfo{author}{S.~Bahl}, \bibinfo{author}{S.~Levine},
\newblock \bibinfo{title}{{Skew-Fit: State-Covering Self-Supervised
  Reinforcement Learning}}  (\bibinfo{year}{2019}). \URLprefix
  \url{http://arxiv.org/abs/1903.03698}.
  \href{http://arxiv.org/abs/1903.03698}{{\tt arXiv:1903.03698}}.
\bibitem[{Gangwani et~al.(2019)Gangwani, Liu, and Peng}]{Hess2019}
\bibinfo{author}{T.~Gangwani}, \bibinfo{author}{Q.~Liu},
  \bibinfo{author}{J.~Peng},
\newblock \bibinfo{title}{{Learning Self-Imitating Diverse Policies}},
\newblock \bibinfo{journal}{7th International Conference on Learning
  Representations, ICLR 2019}  (\bibinfo{year}{2019}).
\bibitem[{Ecoffet et~al.(2020)Ecoffet, Huizinga, Lehman, Stanley, and
  Clune}]{Ecoffet2020}
\bibinfo{author}{A.~Ecoffet}, \bibinfo{author}{J.~Huizinga},
  \bibinfo{author}{J.~Lehman}, \bibinfo{author}{K.~O. Stanley},
  \bibinfo{author}{J.~Clune},
\newblock \bibinfo{title}{{First Return Then Explore}}  (\bibinfo{year}{2020})
  \bibinfo{pages}{1--46}. \URLprefix \url{http://arxiv.org/abs/2004.12919}.
  \href{http://arxiv.org/abs/2004.12919}{{\tt arXiv:2004.12919}}.
\bibitem[{Matheron et~al.(2020)Matheron, Perrin, and Sigaud}]{Matheron2020a}
\bibinfo{author}{G.~Matheron}, \bibinfo{author}{N.~Perrin},
  \bibinfo{author}{O.~Sigaud},
\newblock \bibinfo{title}{{PBCS: Efficient Exploration and Exploitation Using a
  Synergy Between Reinforcement Learning and Motion Planning}},
\newblock \bibinfo{journal}{ICANN 2020} \bibinfo{volume}{12397 LNCS}
  (\bibinfo{year}{2020}) \bibinfo{pages}{295--307}. \URLprefix
  \url{http://dx.doi.org/10.1007/978-3-030-61616-8{\_}24}.
  \DOIprefix\doi{10.1007/978-3-030-61616-8_24}.
  \href{http://arxiv.org/abs/2004.11667}{{\tt arXiv:2004.11667}}.
\bibitem[{Guo and Brunskill(2019)}]{Guo2019}
\bibinfo{author}{Z.~D. Guo}, \bibinfo{author}{E.~Brunskill},
\newblock \bibinfo{title}{{Directed Exploration for Reinforcement Learning}}
  (\bibinfo{year}{2019}). \URLprefix \url{http://arxiv.org/abs/1906.07805}.
  \href{http://arxiv.org/abs/1906.07805}{{\tt arXiv:1906.07805}}.
\bibitem[{Guo et~al.(2019)Guo, Choi, Moczulski, Bengio, Norouzi, and
  Lee}]{Guo2019a}
\bibinfo{author}{Y.~Guo}, \bibinfo{author}{J.~Choi},
  \bibinfo{author}{M.~Moczulski}, \bibinfo{author}{S.~Bengio},
  \bibinfo{author}{M.~Norouzi}, \bibinfo{author}{H.~Lee},
\newblock \bibinfo{title}{{Self-Imitation Learning via Trajectory-Conditioned
  Policy for Hard-Exploration Tasks}}  (\bibinfo{year}{2019})
  \bibinfo{pages}{1--22}. \URLprefix \url{http://arxiv.org/abs/1907.10247}.
  \href{http://arxiv.org/abs/1907.10247}{{\tt arXiv:1907.10247}}.
\bibitem[{Oh et~al.(2018)Oh, Guo, Singh, and Lee}]{Oh2018}
\bibinfo{author}{J.~Oh}, \bibinfo{author}{Y.~Guo}, \bibinfo{author}{S.~Singh},
  \bibinfo{author}{H.~Lee},
\newblock \bibinfo{title}{{Self-Imitation Learning}},
\newblock \bibinfo{journal}{Proceedings of the 35th International Conference on
  Machine Learning}  (\bibinfo{year}{2018}).
\bibitem[{Guo et~al.(2020)Guo, Choi, Moczulski, Feng, Bengio, Norouzi, and
  Lee}]{Guo2020}
\bibinfo{author}{Y.~Guo}, \bibinfo{author}{J.~Choi},
  \bibinfo{author}{M.~Moczulski}, \bibinfo{author}{S.~Feng},
  \bibinfo{author}{S.~Bengio}, \bibinfo{author}{M.~Norouzi},
  \bibinfo{author}{H.~Lee},
\newblock \bibinfo{title}{{Memory Based Trajectory-conditioned Policies for
  Learning from Sparse Rewards}},
\newblock \bibinfo{journal}{Conference on Neural Information Processing
  Systems, NeurIPS 2020}  (\bibinfo{year}{2020}).
\bibitem[{Liu et~al.(2020)Liu, Keramati, Seshadri, Guu, Pasupat, Brunskill, and
  Liang}]{Liu2020b}
\bibinfo{author}{E.~Z. Liu}, \bibinfo{author}{R.~Keramati},
  \bibinfo{author}{S.~Seshadri}, \bibinfo{author}{K.~Guu},
  \bibinfo{author}{P.~Pasupat}, \bibinfo{author}{E.~Brunskill},
  \bibinfo{author}{P.~Liang},
\newblock \bibinfo{title}{{Learning Abstract Models for Strategic Exploration
  and Fast Reward Transfer}}  (\bibinfo{year}{2020}). \URLprefix
  \url{http://arxiv.org/abs/2007.05896}.
  \href{http://arxiv.org/abs/2007.05896}{{\tt arXiv:2007.05896}}.
\bibitem[{Edwards et~al.(2018)Edwards, Downs, and Davidson}]{Edwards2018}
\bibinfo{author}{A.~D. Edwards}, \bibinfo{author}{L.~Downs},
  \bibinfo{author}{J.~C. Davidson},
\newblock \bibinfo{title}{{Forward-Backward Reinforcement Learning}}
  (\bibinfo{year}{2018}). \URLprefix \url{http://arxiv.org/abs/1803.10227}.
  \href{http://arxiv.org/abs/1803.10227}{{\tt arXiv:1803.10227}}.
\bibitem[{Florensa et~al.(2017)Florensa, Held, Wulfmeier, Zhang, and
  Abbeel}]{Florensa2017a}
\bibinfo{author}{C.~Florensa}, \bibinfo{author}{D.~Held},
  \bibinfo{author}{M.~Wulfmeier}, \bibinfo{author}{M.~Zhang},
  \bibinfo{author}{P.~Abbeel},
\newblock \bibinfo{title}{Reverse curriculum generation for reinforcement
  learning}  (\bibinfo{year}{2017}). \URLprefix
  \url{http://arxiv.org/abs/1707.05300}.
  \href{http://arxiv.org/abs/1707.05300}{{\tt arXiv:1707.05300}}.
\bibitem[{Forestier et~al.(2017)Forestier, Mollard, and
  Oudeyer}]{Forestier2017}
\bibinfo{author}{S.~Forestier}, \bibinfo{author}{Y.~Mollard},
  \bibinfo{author}{P.~Y. Oudeyer}, \bibinfo{title}{{Intrinsically Motivated
  Goal Exploration Processes with Automatic Curriculum Learning}},
  \bibinfo{year}{2017}. \URLprefix \url{http://arxiv.org/abs/1708.02190}.
  \href{http://arxiv.org/abs/1708.02190}{{\tt arXiv:1708.02190}}.
\bibitem[{Colas et~al.(2019)Colas, Founder, Sigaud, Chetouani, and
  Oudeyer}]{Colas2019}
\bibinfo{author}{C.~Colas}, \bibinfo{author}{P.~Founder},
  \bibinfo{author}{O.~Sigaud}, \bibinfo{author}{M.~Chetouani},
  \bibinfo{author}{P.~Y. Oudeyer},
\newblock \bibinfo{title}{{CURIOUS: Intrinsically Motivated Modular Multi-goal
  Reinforcement Learning}},
\newblock \bibinfo{journal}{36th International Conference on Machine Learning,
  ICML 2019}  (\bibinfo{year}{2019}) \bibinfo{pages}{2372--2387}.
  \href{http://arxiv.org/abs/1810.06284}{{\tt arXiv:1810.06284}}.
\bibitem[{P{\'{e}}r{\'{e}} et~al.(2018)P{\'{e}}r{\'{e}}, Forestier, Sigaud, and
  Oudeyer}]{Pere2018}
\bibinfo{author}{A.~P{\'{e}}r{\'{e}}}, \bibinfo{author}{S.~Forestier},
  \bibinfo{author}{O.~Sigaud}, \bibinfo{author}{P.~Y. Oudeyer},
\newblock \bibinfo{title}{{Unsupervised Learning Of Goal Spaces For
  Intrinsically Motivated Goal Exploration}},
\newblock \bibinfo{journal}{6th International Conference on Learning
  Representations, ICLR 2018}  (\bibinfo{year}{2018}) \bibinfo{pages}{1--26}.
  \href{http://arxiv.org/abs/1803.00781}{{\tt arXiv:1803.00781}}.
\bibitem[{Vezhnevets et~al.(2017)Vezhnevets, Osindero, Schaul, Heess,
  Jaderberg, Silver, and Kavukcuoglu}]{Vezhnevets2017a}
\bibinfo{author}{A.~S. Vezhnevets}, \bibinfo{author}{S.~Osindero},
  \bibinfo{author}{T.~Schaul}, \bibinfo{author}{N.~Heess},
  \bibinfo{author}{M.~Jaderberg}, \bibinfo{author}{D.~Silver},
  \bibinfo{author}{K.~Kavukcuoglu},
\newblock \bibinfo{title}{{FeUdal Networks for Hierarchical Reinforcement
  Learning}},
\newblock \bibinfo{journal}{34th International Conference on Machine Learning,
  ICML 2017} \bibinfo{volume}{7} (\bibinfo{year}{2017})
  \bibinfo{pages}{5409--5418}. \href{http://arxiv.org/abs/1703.01161}{{\tt
  arXiv:1703.01161}}.
\bibitem[{Hester and Stone(2013)}]{Hester2013a}
\bibinfo{author}{T.~Hester}, \bibinfo{author}{P.~Stone},
\newblock \bibinfo{title}{{Learning Exploration Strategies in Model-Based
  Reinforcement Learning}},
\newblock \bibinfo{journal}{Proceedings of the 12th International Conference on
  Autonomous Agents and Multiagent Systems (AAAI}  (\bibinfo{year}{2013}).
\bibitem[{Kulkarni et~al.(2016)Kulkarni, Narasimhan, Saeedi, and
  Tenenbaum}]{Kulkarni2016a}
\bibinfo{author}{T.~D. Kulkarni}, \bibinfo{author}{K.~R. Narasimhan},
  \bibinfo{author}{A.~Saeedi}, \bibinfo{author}{J.~B. Tenenbaum},
\newblock \bibinfo{title}{{Hierarchical Deep Reinforcement Learning:
  Integrating Temporal Abstraction and Intrinsic Motivation}},
\newblock \bibinfo{journal}{Conference on Neural Information Processing
  Systems, NeurIPS 2016}  (\bibinfo{year}{2016}). \DOIprefix\doi{10.1162/NECO}.
  \href{http://arxiv.org/abs/NIHMS150003}{{\tt arXiv:NIHMS150003}}.
\bibitem[{Riedmiller et~al.(2018)Riedmiller, Hafner, Lampe, Neunert, Degrave,
  van~de Wiele, Mnih, Heess, and Springenberg}]{Riedmiller2018a}
\bibinfo{author}{M.~Riedmiller}, \bibinfo{author}{R.~Hafner},
  \bibinfo{author}{T.~Lampe}, \bibinfo{author}{M.~Neunert},
  \bibinfo{author}{J.~Degrave}, \bibinfo{author}{T.~van~de Wiele},
  \bibinfo{author}{V.~Mnih}, \bibinfo{author}{N.~Heess},
  \bibinfo{author}{T.~Springenberg},
\newblock \bibinfo{title}{{Learning by Playing - Solving Sparse Reward Tasks
  From Scratch}},
\newblock \bibinfo{journal}{Proceedings of the 35th International Conference on
  Machine Learning}  (\bibinfo{year}{2018}).
\bibitem[{Ghafoorian et~al.(2013)Ghafoorian, Taghizadeh, and
  Beigy}]{Ghafoorian2013a}
\bibinfo{author}{M.~Ghafoorian}, \bibinfo{author}{N.~Taghizadeh},
  \bibinfo{author}{H.~Beigy},
\newblock \bibinfo{title}{{Automatic Abstraction in Reinforcement Learning
  Using Ant System Algorithm}},
\newblock \bibinfo{journal}{AAAI Spring Symposium - Technical Report}
  \bibinfo{volume}{SS-13-05} (\bibinfo{year}{2013}) \bibinfo{pages}{9--14}.
\bibitem[{Machado et~al.(2017)Machado, Bellemare, and Bowling}]{Machado2017}
\bibinfo{author}{M.~C. Machado}, \bibinfo{author}{M.~G. Bellemare},
  \bibinfo{author}{M.~Bowling},
\newblock \bibinfo{title}{{A Laplacian Framework for Option Discovery in
  Reinforcement Learning}},
\newblock \bibinfo{journal}{34th International Conference on Machine Learning,
  ICML 2017} \bibinfo{volume}{5} (\bibinfo{year}{2017})
  \bibinfo{pages}{3567--3582}. \href{http://arxiv.org/abs/1703.00956}{{\tt
  arXiv:1703.00956}}.
\bibitem[{Zaki and Meira(2014)}]{zaki_meira}
\bibinfo{author}{M.~J. Zaki}, \bibinfo{author}{W.~Meira, Jr},
  \bibinfo{title}{Data Mining and Analysis: Fundamental Concepts and
  Algorithms}, \bibinfo{publisher}{Cambridge University Press},
  \bibinfo{year}{2014}. \DOIprefix\doi{10.1017/CBO9780511810114}.
\bibitem[{Machado et~al.(2018)Machado, Rosenbaum, Guo, Liu, Tesauro, and
  Campbell}]{Machado2018}
\bibinfo{author}{M.~C. Machado}, \bibinfo{author}{C.~Rosenbaum},
  \bibinfo{author}{X.~Guo}, \bibinfo{author}{M.~Liu},
  \bibinfo{author}{G.~Tesauro}, \bibinfo{author}{M.~Campbell},
\newblock \bibinfo{title}{{Eigenoption Discovery Through The Deep Successor
  Representation}},
\newblock \bibinfo{journal}{6th International Conference on Learning
  Representations, ICLR 2018}  (\bibinfo{year}{2018}).
  \href{http://arxiv.org/abs/arXiv:1710.11089v3}{{\tt
  arXiv:arXiv:1710.11089v3}}.
\bibitem[{Fang et~al.(2020)Fang, Zhu, Savarese, and Fei-Fei}]{Fang2020}
\bibinfo{author}{K.~Fang}, \bibinfo{author}{Y.~Zhu},
  \bibinfo{author}{S.~Savarese}, \bibinfo{author}{L.~Fei-Fei},
\newblock \bibinfo{title}{{Adaptive Procedural Task Generation for
  Hard-Exploration Problems}},
\newblock \bibinfo{journal}{Under Review at ICLR 2021}  (\bibinfo{year}{2020}).
  \URLprefix \url{http://arxiv.org/abs/2007.00350}.
  \href{http://arxiv.org/abs/2007.00350}{{\tt arXiv:2007.00350}}.
\bibitem[{Guestrin et~al.(2002)Guestrin, Patrascu, and
  Schuurmans}]{Guestrin2002a}
\bibinfo{author}{C.~Guestrin}, \bibinfo{author}{R.~Patrascu},
  \bibinfo{author}{D.~Schuurmans},
\newblock \bibinfo{title}{{Algorithm-directed Exploration for Model-Based
  Reinforcement Learning In Factored Mdps}},
\newblock \bibinfo{journal}{Machine Learning International Workshop}
  (\bibinfo{year}{2002}) \bibinfo{pages}{235--242}. \URLprefix
  \url{http://scholar.google.com/scholar?hl=en{\&}btnG=Search{\&}q=intitle:Algorithm-Directed+Exploration+for+Model-Based+Reinforcement+Learning+in+Factored+MDPs{\#}0}.
\bibitem[{Abel et~al.(2016)Abel, Agarwal, Diaz, Krishnamurthy, and
  Schapire}]{Abel2016a}
\bibinfo{author}{D.~Abel}, \bibinfo{author}{A.~Agarwal},
  \bibinfo{author}{F.~Diaz}, \bibinfo{author}{A.~Krishnamurthy},
  \bibinfo{author}{R.~E. Schapire},
\newblock \bibinfo{title}{{Exploratory Gradient Boosting for Reinforcement
  Learning in Complex Domains}}  (\bibinfo{year}{2016}). \URLprefix
  \url{http://arxiv.org/abs/1603.04119}.
  \href{http://arxiv.org/abs/1603.04119}{{\tt arXiv:1603.04119}}.
\bibitem[{Kova{\v{c}} et~al.(2020)Kova{\v{c}}, Laversanne-Finot, and
  Oudeyer}]{Kovac2020}
\bibinfo{author}{G.~Kova{\v{c}}}, \bibinfo{author}{A.~Laversanne-Finot},
  \bibinfo{author}{P.-Y. Oudeyer},
\newblock \bibinfo{title}{{GRIMGEP: Learning Progress for Robust Goal Sampling
  in Visual Deep Reinforcement Learning}}  (\bibinfo{year}{2020})
  \bibinfo{pages}{1--15}. \URLprefix
  \url{https://sites.google.com/view/grimgep}.
  \href{http://arxiv.org/abs/2008.04388v1}{{\tt arXiv:2008.04388v1}}.
\bibitem[{Osband and {Van Roy}(2017)}]{Osband2017a}
\bibinfo{author}{I.~Osband}, \bibinfo{author}{B.~{Van Roy}},
\newblock \bibinfo{title}{{Why is Posterior Sampling Better Than Optimism for
  Reinforcement Learning?}},
\newblock \bibinfo{journal}{34th International Conference on Machine Learning,
  ICML 2017}  (\bibinfo{year}{2017}) \bibinfo{pages}{4133--4148}.
  \href{http://arxiv.org/abs/1607.00215}{{\tt arXiv:1607.00215}}.
\bibitem[{Jung and Stone(2010)}]{Jung2010}
\bibinfo{author}{T.~Jung}, \bibinfo{author}{P.~Stone},
\newblock \bibinfo{title}{{Gaussian Processes for Sample Efficient
  Reinforcement Learning With Rmax-Like Exploration}},
\newblock \bibinfo{journal}{Lecture Notes in Computer Science (including
  subseries Lecture Notes in Artificial Intelligence and Lecture Notes in
  Bioinformatics)} \bibinfo{volume}{6321 LNAI} (\bibinfo{year}{2010})
  \bibinfo{pages}{601--616}. \DOIprefix\doi{10.1007/978-3-642-15880-3_44}.
  \href{http://arxiv.org/abs/1201.6604}{{\tt arXiv:1201.6604}}.
\bibitem[{Xie et~al.(2016)Xie, Patil, Moldovan, Levine, and Abbeel}]{Xie2016}
\bibinfo{author}{C.~Xie}, \bibinfo{author}{S.~Patil},
  \bibinfo{author}{T.~Moldovan}, \bibinfo{author}{S.~Levine},
  \bibinfo{author}{P.~Abbeel},
\newblock \bibinfo{title}{{Model-Based Reinforcement Learning With Parametrized
  Physical Models And Optimism-Driven Exploration}},
\newblock \bibinfo{journal}{IEEE International Conference on Robotics and
  Automation, ICRA 2016}  (\bibinfo{year}{2016}) \bibinfo{pages}{504--511}.
  \DOIprefix\doi{10.1109/ICRA.2016.7487172}.
  \href{http://arxiv.org/abs/1509.06824}{{\tt arXiv:1509.06824}}.
\bibitem[{D'Eramo et~al.(2019)D'Eramo, Cini, and Restelli}]{DEramo2019}
\bibinfo{author}{C.~D'Eramo}, \bibinfo{author}{A.~Cini},
  \bibinfo{author}{M.~Restelli},
\newblock \bibinfo{title}{Exploiting action-value uncertainty to drive
  exploration in reinforcement learning},
\newblock \bibinfo{journal}{Proceedings of the International Joint Conference
  on Neural Networks, IJCNN 2019}  (\bibinfo{year}{2019}).
  \DOIprefix\doi{10.1109/IJCNN.2019.8852326}.
\bibitem[{Osband et~al.(2016)Osband, {Van Roy}, and Wen}]{Osband2016a}
\bibinfo{author}{I.~Osband}, \bibinfo{author}{B.~{Van Roy}},
  \bibinfo{author}{Z.~Wen},
\newblock \bibinfo{title}{Generalization and exploration via randomized value
  functions},
\newblock \bibinfo{journal}{33rd International Conference on Machine Learning,
  ICML 2016}  (\bibinfo{year}{2016}) \bibinfo{pages}{3540--3561}.
  \href{http://arxiv.org/abs/1402.0635}{{\tt arXiv:1402.0635}}.
\bibitem[{Osband et~al.(2019)Osband, {Van Roy}, Russo, and Wen}]{Osband2019b}
\bibinfo{author}{I.~Osband}, \bibinfo{author}{B.~{Van Roy}},
  \bibinfo{author}{D.~J. Russo}, \bibinfo{author}{Z.~Wen},
\newblock \bibinfo{title}{{Deep Exploration via Randomized Value Functions}},
\newblock \bibinfo{journal}{Journal of Machine Learning Research}
  \bibinfo{volume}{20} (\bibinfo{year}{2019}) \bibinfo{pages}{1--62}.
  \href{http://arxiv.org/abs/1703.07608}{{\tt arXiv:1703.07608}}.
\bibitem[{Tang and Agrawal(2018)}]{Tang2018}
\bibinfo{author}{Y.~Tang}, \bibinfo{author}{S.~Agrawal},
\newblock \bibinfo{title}{{Exploration by Distributional Reinforcement
  Learning}},
\newblock \bibinfo{journal}{IJCAI International Joint Conference on Artificial
  Intelligence} \bibinfo{volume}{2018-July} (\bibinfo{year}{2018})
  \bibinfo{pages}{2710--2716}. \DOIprefix\doi{10.24963/ijcai.2018/376}.
  \href{http://arxiv.org/abs/1805.01907}{{\tt arXiv:1805.01907}}.
\bibitem[{Colas et~al.(2018)Colas, Sigaud, and Oudeyer}]{Colas2018}
\bibinfo{author}{C.~Colas}, \bibinfo{author}{O.~Sigaud}, \bibinfo{author}{P.~Y.
  Oudeyer},
\newblock \bibinfo{title}{{GEP-PG: Decoupling Exploration and Exploitation in
  Deep Reinforcement Learning Algorithms}},
\newblock \bibinfo{journal}{Proceedings of the 35th International Conference on
  Machine Learning, ICML 2018}  (\bibinfo{year}{2018}).
\bibitem[{Janz et~al.(2019)Janz, Hron, Mazur, Hofmann, Hern{\'{a}}ndez-Lobato,
  and Tschiatschek}]{Janz2019}
\bibinfo{author}{D.~Janz}, \bibinfo{author}{J.~Hron},
  \bibinfo{author}{P.~Mazur}, \bibinfo{author}{K.~Hofmann},
  \bibinfo{author}{J.~M. Hern{\'{a}}ndez-Lobato},
  \bibinfo{author}{S.~Tschiatschek},
\newblock \bibinfo{title}{{Successor Uncertainties: Exploration and Uncertainty
  in Temporal Difference Learning}},
\newblock \bibinfo{journal}{Conference on Neural Information Processing
  Systems, NeurIPS 2019} \bibinfo{volume}{33} (\bibinfo{year}{2019}).
  \href{http://arxiv.org/abs/1810.06530}{{\tt arXiv:1810.06530}}.
\bibitem[{Stulp(2012)}]{Stulp2012}
\bibinfo{author}{F.~Stulp},
\newblock \bibinfo{title}{{Adaptive Exploration for Continual Reinforcement
  Learning}},
\newblock \bibinfo{journal}{IEEE International Conference on Intelligent Robots
  and Systems, IROS 2012}  (\bibinfo{year}{2012}) \bibinfo{pages}{1631--1636}.
  \DOIprefix\doi{10.1109/IROS.2012.6385818}.
\bibitem[{Akiyama et~al.(2010)Akiyama, Hachiya, and Sugiyama}]{Akiyama2010}
\bibinfo{author}{T.~Akiyama}, \bibinfo{author}{H.~Hachiya},
  \bibinfo{author}{M.~Sugiyama},
\newblock \bibinfo{title}{{Efficient exploration through active learning for
  value function approximation in reinforcement learning}},
\newblock \bibinfo{journal}{Neural Networks} \bibinfo{volume}{23}
  (\bibinfo{year}{2010}) \bibinfo{pages}{639--648}. \URLprefix
  \url{http://dx.doi.org/10.1016/j.neunet.2009.12.010}.
  \DOIprefix\doi{10.1016/j.neunet.2009.12.010}.
\bibitem[{Strens(2000)}]{Strens2000a}
\bibinfo{author}{M.~Strens},
\newblock \bibinfo{title}{{A Bayesian Framework for Reinforcement Learning}},
\newblock \bibinfo{journal}{Proc of the 17th International Conference on
  Machine Learning}  (\bibinfo{year}{2000}) \bibinfo{pages}{943--950}.
  \URLprefix
  \url{http://citeseerx.ist.psu.edu/viewdoc/download?doi=10.1.1.140.1701{\&}rep=rep1{\&}type=pdf}.
\bibitem[{Guez et~al.(2012)Guez, Silver, and Dayan}]{Guez2012}
\bibinfo{author}{A.~Guez}, \bibinfo{author}{D.~Silver},
  \bibinfo{author}{P.~Dayan},
\newblock \bibinfo{title}{{Efficient Bayes-adaptive Reinforcement Learning
  Using Sample-based Search}},
\newblock \bibinfo{journal}{Conference on Neural Information Processing
  Systems, NeurIPS 2012}  (\bibinfo{year}{2012}) \bibinfo{pages}{1025--1033}.
  \href{http://arxiv.org/abs/1205.3109}{{\tt arXiv:1205.3109}}.
\bibitem[{O'Donoghue et~al.(2018)O'Donoghue, Osband, Munos, and
  Mnih}]{ODonoghue2018a}
\bibinfo{author}{B.~O'Donoghue}, \bibinfo{author}{I.~Osband},
  \bibinfo{author}{R.~Munos}, \bibinfo{author}{V.~Mnih},
\newblock \bibinfo{title}{{The Uncertainty Bellman Equation And Exploration}},
\newblock \bibinfo{journal}{35th International Conference on Machine Learning,
  ICML 2018} \bibinfo{volume}{9} (\bibinfo{year}{2018})
  \bibinfo{pages}{6154--6173}. \href{http://arxiv.org/abs/1709.05380}{{\tt
  arXiv:1709.05380}}.
\bibitem[{Nikolov et~al.(2019)Nikolov, Kirschner, Berkenkamp, and
  Krause}]{Nikolov2019}
\bibinfo{author}{N.~Nikolov}, \bibinfo{author}{J.~Kirschner},
  \bibinfo{author}{F.~Berkenkamp}, \bibinfo{author}{A.~Krause},
\newblock \bibinfo{title}{{Information-Directed Exploration for Deep
  Reinforcement Learning}},
\newblock \bibinfo{journal}{7th International Conference on Learning
  Representations, ICLR 2019}  (\bibinfo{year}{2019}).
  \href{http://arxiv.org/abs/1812.07544}{{\tt arXiv:1812.07544}}.
\bibitem[{Osband et~al.(2016)Osband, Blundell, Pritzel, and Roy}]{Osband2016}
\bibinfo{author}{I.~Osband}, \bibinfo{author}{C.~Blundell},
  \bibinfo{author}{A.~Pritzel}, \bibinfo{author}{B.~V. Roy},
\newblock \bibinfo{title}{Deep exploration via bootstrapped dqn},
\newblock \bibinfo{journal}{Conference on Neural Information Processing
  Systems, NeurIPS 2016}  (\bibinfo{year}{2016}).
\bibitem[{Pearce et~al.(2018)Pearce, Anastassacos, Zaki, and
  Neely}]{Pearce2018}
\bibinfo{author}{T.~Pearce}, \bibinfo{author}{N.~Anastassacos},
  \bibinfo{author}{M.~Zaki}, \bibinfo{author}{A.~Neely},
\newblock \bibinfo{title}{{Bayesian Inference with Anchored Ensembles of Neural
  Networks, and Application to Exploration in Reinforcement Learning}},
\newblock \bibinfo{journal}{Exploration in Reinforcement Learning Work- shop at
  the 35th International Conference on Machine Learning}
  (\bibinfo{year}{2018}). \URLprefix \url{http://arxiv.org/abs/1805.11324}.
  \href{http://arxiv.org/abs/1805.11324}{{\tt arXiv:1805.11324}}.
\bibitem[{Shyam et~al.(2019)Shyam, Jaskowski, and Gomez}]{Shyam2019a}
\bibinfo{author}{P.~Shyam}, \bibinfo{author}{W.~Jaskowski},
  \bibinfo{author}{F.~Gomez},
\newblock \bibinfo{title}{{Model-based Active Exploration}},
\newblock \bibinfo{journal}{36th International Conference on Machine Learning,
  ICML 2019}  (\bibinfo{year}{2019}) \bibinfo{pages}{10136--10152}.
  \href{http://arxiv.org/abs/1810.12162}{{\tt arXiv:1810.12162}}.
\bibitem[{Henaff(2019)}]{Henaff2019a}
\bibinfo{author}{M.~Henaff},
\newblock \bibinfo{title}{Explicit explore-exploit algorithms in continuous
  state spaces},
\newblock \bibinfo{journal}{Conference on Neural Information Processing
  Systems, NeurIPS 2019}  (\bibinfo{year}{2019}).
  \href{http://arxiv.org/abs/1911.00617}{{\tt arXiv:1911.00617}}.
\bibitem[{Vecerik et~al.(2017)Vecerik, Hester, Scholz, Wang, Pietquin, Piot,
  Heess, Roth{\"{o}}rl, Lampe, and Riedmiller}]{Vecerik2017}
\bibinfo{author}{M.~Vecerik}, \bibinfo{author}{T.~Hester},
  \bibinfo{author}{J.~Scholz}, \bibinfo{author}{F.~Wang},
  \bibinfo{author}{O.~Pietquin}, \bibinfo{author}{B.~Piot},
  \bibinfo{author}{N.~Heess}, \bibinfo{author}{T.~Roth{\"{o}}rl},
  \bibinfo{author}{T.~Lampe}, \bibinfo{author}{M.~Riedmiller},
\newblock \bibinfo{title}{{Leveraging Demonstrations for Deep Reinforcement
  Learning on Robotics Problems with Sparse Rewards}}  (\bibinfo{year}{2017}).
  \URLprefix \url{http://arxiv.org/abs/1707.08817}.
  \href{http://arxiv.org/abs/1707.08817}{{\tt arXiv:1707.08817}}.
\bibitem[{Hester et~al.(2018)Hester, Schaul, Sendonaris, Vecerik, Piot, Osband,
  Pietquin, Horgan, Dulac-Arnold, Lanctot, Quan, Agapiou, Leibo, and
  Gruslys}]{Hester2018a}
\bibinfo{author}{T.~Hester}, \bibinfo{author}{T.~Schaul},
  \bibinfo{author}{A.~Sendonaris}, \bibinfo{author}{M.~Vecerik},
  \bibinfo{author}{B.~Piot}, \bibinfo{author}{I.~Osband},
  \bibinfo{author}{O.~Pietquin}, \bibinfo{author}{D.~Horgan},
  \bibinfo{author}{G.~Dulac-Arnold}, \bibinfo{author}{M.~Lanctot},
  \bibinfo{author}{J.~Quan}, \bibinfo{author}{J.~Agapiou},
  \bibinfo{author}{J.~Z. Leibo}, \bibinfo{author}{A.~Gruslys},
\newblock \bibinfo{title}{{Deep q-learning From Demonstrations}},
\newblock \bibinfo{journal}{32nd AAAI Conference on Artificial Intelligence,
  AAAI 2018}  (\bibinfo{year}{2018}) \bibinfo{pages}{3223--3230}.
  \href{http://arxiv.org/abs/1704.03732}{{\tt arXiv:1704.03732}}.
\bibitem[{Gulcehr et~al.(2020)Gulcehr, Paine, Shahriari, Denil, Hoffman, Soyer,
  Tanburn, Kapturowski, Rabinowitz, Williams, Barth-Maron, Wang, and
  de~Freitas}]{Gulcehr2020}
\bibinfo{author}{C.~Gulcehr}, \bibinfo{author}{T.~L. Paine},
  \bibinfo{author}{B.~Shahriari}, \bibinfo{author}{M.~Denil},
  \bibinfo{author}{M.~Hoffman}, \bibinfo{author}{H.~Soyer},
  \bibinfo{author}{R.~Tanburn}, \bibinfo{author}{S.~Kapturowski},
  \bibinfo{author}{N.~Rabinowitz}, \bibinfo{author}{D.~Williams},
  \bibinfo{author}{G.~Barth-Maron}, \bibinfo{author}{Z.~Wang},
  \bibinfo{author}{N.~de~Freitas},
\newblock \bibinfo{title}{{Making Efficient Use of Demonstrations To Solve Hard
  Exploration Problems}},
\newblock \bibinfo{journal}{8th International Conference on Learning
  Representations, ICLR 2020}  (\bibinfo{year}{2020}).
\bibitem[{Nair et~al.(2018)Nair, McGrew, Andrychowicz, Zaremba, and
  Abbeel}]{Nair2018c}
\bibinfo{author}{A.~Nair}, \bibinfo{author}{B.~McGrew},
  \bibinfo{author}{M.~Andrychowicz}, \bibinfo{author}{W.~Zaremba},
  \bibinfo{author}{P.~Abbeel},
\newblock \bibinfo{title}{{Overcoming Exploration in Reinforcement Learning
  with Demonstrations}},
\newblock \bibinfo{journal}{Proceedings - IEEE International Conference on
  Robotics and Automation}  (\bibinfo{year}{2018}) \bibinfo{pages}{6292--6299}.
  \DOIprefix\doi{10.1109/ICRA.2018.8463162}.
  \href{http://arxiv.org/abs/1709.10089}{{\tt arXiv:1709.10089}}.
\bibitem[{Salimans and Chen(2018)}]{Salimans2018a}
\bibinfo{author}{T.~Salimans}, \bibinfo{author}{R.~Chen},
\newblock \bibinfo{title}{{Learning Montezuma's Revenge from a Single
  Demonstration}},
\newblock \bibinfo{journal}{Conference on Neural Information Processing
  Systems, NeurIPS 2018}  (\bibinfo{year}{2018}). \URLprefix
  \url{http://arxiv.org/abs/1812.03381}.
  \href{http://arxiv.org/abs/1812.03381}{{\tt arXiv:1812.03381}}.
\bibitem[{Subramanian et~al.(2016)Subramanian, Isbell, and
  Thomaz}]{Subramanian2016a}
\bibinfo{author}{K.~Subramanian}, \bibinfo{author}{C.~L. Isbell},
  \bibinfo{author}{A.~L. Thomaz},
\newblock \bibinfo{title}{{Exploration From Demonstration for Interactive
  Reinforcement Learning}},
\newblock \bibinfo{journal}{Proceedings of the International Joint Conference
  on Autonomous Agents and Multiagent Systems, AAMAS}  (\bibinfo{year}{2016})
  \bibinfo{pages}{447--456}.
\bibitem[{Garcıa and Fernandez(2015)}]{Garcia2015}
\bibinfo{author}{J.~Garcıa}, \bibinfo{author}{F.~Fernandez},
\newblock \bibinfo{title}{{A Comprehensive Survey on Safe Reinforcement
  Learning}},
\newblock \bibinfo{journal}{Journal of Machine Learning Research}
  \bibinfo{volume}{16} (\bibinfo{year}{2015}).
\bibitem[{Garcia and Fernandez(2012)}]{Garcia2012}
\bibinfo{author}{J.~Garcia}, \bibinfo{author}{F.~Fernandez},
\newblock \bibinfo{title}{{Safe Exploration of State and Action Spaces in
  Reinforcement Learning}},
\newblock \bibinfo{journal}{Journal of Artificial Intelligence Research}
  \bibinfo{volume}{45} (\bibinfo{year}{2012}) \bibinfo{pages}{515--564}.
  \DOIprefix\doi{10.1613/jair.3761}.
\bibitem[{Dalal et~al.(2018)Dalal, Dvijotham, Vecerik, Hester, Paduraru, and
  Tassa}]{Dalal2018a}
\bibinfo{author}{G.~Dalal}, \bibinfo{author}{K.~Dvijotham},
  \bibinfo{author}{M.~Vecerik}, \bibinfo{author}{T.~Hester},
  \bibinfo{author}{C.~Paduraru}, \bibinfo{author}{Y.~Tassa},
\newblock \bibinfo{title}{{Safe Exploration in Continuous Action Spaces}}
  (\bibinfo{year}{2018}). \URLprefix \url{http://arxiv.org/abs/1801.08757}.
  \href{http://arxiv.org/abs/1801.08757}{{\tt arXiv:1801.08757}}.
\bibitem[{Garcelon et~al.(2020)Garcelon, Ghavamzadeh, Lazaric, and
  Pirotta}]{Garcelon2020a}
\bibinfo{author}{E.~Garcelon}, \bibinfo{author}{M.~Ghavamzadeh},
  \bibinfo{author}{A.~Lazaric}, \bibinfo{author}{M.~Pirotta},
\newblock \bibinfo{title}{{Conservative Exploration in Reinforcement
  Learning}},
\newblock \bibinfo{journal}{Proceedings of the 23rd International Conference on
  Artificial Intelligence and Statistics}  (\bibinfo{year}{2020}). \URLprefix
  \url{http://arxiv.org/abs/2002.03218}.
  \href{http://arxiv.org/abs/2002.03218}{{\tt arXiv:2002.03218}}.
\bibitem[{Hunt et~al.(2020)Hunt, Fulton, Magliacane, Hoang, Das, and
  Solar-Lezama}]{Hunt2020}
\bibinfo{author}{N.~Hunt}, \bibinfo{author}{N.~Fulton},
  \bibinfo{author}{S.~Magliacane}, \bibinfo{author}{N.~Hoang},
  \bibinfo{author}{S.~Das}, \bibinfo{author}{A.~Solar-Lezama},
\newblock \bibinfo{title}{{Verifiably Safe Exploration for End-to-End
  Reinforcement Learning}}  (\bibinfo{year}{2020}). \URLprefix
  \url{http://arxiv.org/abs/2007.01223}.
  \href{http://arxiv.org/abs/2007.01223}{{\tt arXiv:2007.01223}}.
\bibitem[{Saunders et~al.(2018)Saunders, Stuhlm{\"{u}}ller, Sastry, and
  Evans}]{Saunders2018}
\bibinfo{author}{W.~Saunders}, \bibinfo{author}{A.~Stuhlm{\"{u}}ller},
  \bibinfo{author}{G.~Sastry}, \bibinfo{author}{O.~Evans},
\newblock \bibinfo{title}{{Trial Without Error: Towards Safe Reinforcement
  Learning via Human Intervention}},
\newblock \bibinfo{journal}{Proceedings of the International Joint Conference
  on Autonomous Agents and Multiagent Systems, AAMAS}  (\bibinfo{year}{2018})
  \bibinfo{pages}{2067--2069}. \href{http://arxiv.org/abs/1707.05173}{{\tt
  arXiv:1707.05173}}.
\bibitem[{Turchetta et~al.(2016)Turchetta, Berkenkamp, and
  Krause}]{Turchetta2016a}
\bibinfo{author}{M.~Turchetta}, \bibinfo{author}{F.~Berkenkamp},
  \bibinfo{author}{A.~Krause},
\newblock \bibinfo{title}{{Safe Exploration in Finite Markov Decision Processes
  with Gaussian Processes}},
\newblock \bibinfo{journal}{Conference on Neural Information Processing
  Systems, NeurIPS 2016}  (\bibinfo{year}{2016}) \bibinfo{pages}{4312--4320}.
  \href{http://arxiv.org/abs/1606.04753}{{\tt arXiv:1606.04753}}.
\bibitem[{Gao et~al.(2019)Gao, Kartal, Hernandez-Leal, and Taylor}]{Gao2019a}
\bibinfo{author}{C.~Gao}, \bibinfo{author}{B.~Kartal},
  \bibinfo{author}{P.~Hernandez-Leal}, \bibinfo{author}{M.~E. Taylor},
\newblock \bibinfo{title}{{On Hard Exploration for Reinforcement Learning: a
  Case Study in Pommerman}},
\newblock \bibinfo{journal}{Fifteenth AAAI Conference on Artificial
  Intelligence and Interactive Digital Entertainment}  (\bibinfo{year}{2019}).
  \URLprefix \url{http://arxiv.org/abs/1907.11788}.
  \href{http://arxiv.org/abs/1907.11788}{{\tt arXiv:1907.11788}}.
\bibitem[{Lipton et~al.(2016)Lipton, Azizzadenesheli, Kumar, Li, Gao, and
  Deng}]{Lipton2016}
\bibinfo{author}{Z.~C. Lipton}, \bibinfo{author}{K.~Azizzadenesheli},
  \bibinfo{author}{A.~Kumar}, \bibinfo{author}{L.~Li},
  \bibinfo{author}{J.~Gao}, \bibinfo{author}{L.~Deng},
\newblock \bibinfo{title}{{Combating Reinforcement Learning's Sisyphean Curse
  with Intrinsic Fear}}  (\bibinfo{year}{2016}). \URLprefix
  \url{http://arxiv.org/abs/1611.01211}.
  \href{http://arxiv.org/abs/1611.01211}{{\tt arXiv:1611.01211}}.
\bibitem[{Fatemi et~al.(2019)Fatemi, Sharma, van Seijen, and
  Kahou}]{Fatemi2019}
\bibinfo{author}{M.~Fatemi}, \bibinfo{author}{S.~Sharma},
  \bibinfo{author}{H.~van Seijen}, \bibinfo{author}{S.~E. Kahou},
\newblock \bibinfo{title}{{Dead-ends and Secure Exploration in Reinforcement
  Learning}},
\newblock \bibinfo{journal}{36th International Conference on Machine Learning,
  ICML 2019}  (\bibinfo{year}{2019}) \bibinfo{pages}{3315--3323}.
\bibitem[{Karimpanal et~al.(2020)Karimpanal, Rana, Gupta, Tran, and
  Venkatesh}]{Karimpanal2020a}
\bibinfo{author}{T.~G. Karimpanal}, \bibinfo{author}{S.~Rana},
  \bibinfo{author}{S.~Gupta}, \bibinfo{author}{T.~Tran},
  \bibinfo{author}{S.~Venkatesh},
\newblock \bibinfo{title}{{Learning Transferable Domain Priors for Safe
  Exploration in Reinforcement Learning}}  (\bibinfo{year}{2020})
  \bibinfo{pages}{1--10}. \DOIprefix\doi{10.1109/ijcnn48605.2020.9207344}.
  \href{http://arxiv.org/abs/1909.04307}{{\tt arXiv:1909.04307}}.
\bibitem[{Patrascu and Stacey(1999)}]{Patrascu1999}
\bibinfo{author}{R.~Patrascu}, \bibinfo{author}{D.~Stacey},
\newblock \bibinfo{title}{{Adaptive Exploration in Reinforcement Learning}},
\newblock \bibinfo{journal}{Proceedings of the International Joint Conference
  on Neural Networks} \bibinfo{volume}{4} (\bibinfo{year}{1999})
  \bibinfo{pages}{2276--2281}. \DOIprefix\doi{10.1109/ijcnn.1999.833417}.
\bibitem[{Tokic(2010)}]{Tokic2010}
\bibinfo{author}{M.~Tokic},
\newblock \bibinfo{title}{{Adaptive $\epsilon$-greedy Exploration in
  Reinforcement Learning Based on Value Differences}},
\newblock \bibinfo{journal}{Lecture Notes in Computer Science (including
  subseries Lecture Notes in Artificial Intelligence and Lecture Notes in
  Bioinformatics)} \bibinfo{volume}{6359 LNAI} (\bibinfo{year}{2010})
  \bibinfo{pages}{203--210}. \DOIprefix\doi{10.1007/978-3-642-16111-7_23}.
\bibitem[{Usama and Chang(2019)}]{Usama2019a}
\bibinfo{author}{M.~Usama}, \bibinfo{author}{D.~E. Chang},
\newblock \bibinfo{title}{{Learning-Driven Exploration for Reinforcement
  Learning}}  (\bibinfo{year}{2019}). \URLprefix
  \url{http://arxiv.org/abs/1906.06890}.
  \href{http://arxiv.org/abs/1906.06890}{{\tt arXiv:1906.06890}}.
\bibitem[{Shani et~al.(2019)Shani, Efroni, and Mannor}]{Shani2019a}
\bibinfo{author}{L.~Shani}, \bibinfo{author}{Y.~Efroni},
  \bibinfo{author}{S.~Mannor},
\newblock \bibinfo{title}{{Exploration Conscious Reinforcement Learning
  Revisited}},
\newblock \bibinfo{journal}{36th International Conference on Machine Learning,
  ICML 2019}  (\bibinfo{year}{2019}) \bibinfo{pages}{9986--10012}.
  \href{http://arxiv.org/abs/1812.05551}{{\tt arXiv:1812.05551}}.
\bibitem[{Khamassi et~al.(2017)Khamassi, Velentzas, Tsitsimis, and
  Tzafestas}]{Khamassi2017}
\bibinfo{author}{M.~Khamassi}, \bibinfo{author}{G.~Velentzas},
  \bibinfo{author}{T.~Tsitsimis}, \bibinfo{author}{C.~Tzafestas},
\newblock \bibinfo{title}{{Active Exploration And Parameterized Reinforcement
  Learning Applied to A Simulated Human-Robot Interaction Task}},
\newblock \bibinfo{journal}{2017 1st IEEE International Conference on Robotic
  Computing, IRC 2017}  (\bibinfo{year}{2017}) \bibinfo{pages}{28--35}.
  \DOIprefix\doi{10.1109/IRC.2017.33}.
\bibitem[{Chang(2004)}]{Chang2004a}
\bibinfo{author}{H.~S. Chang},
\newblock \bibinfo{title}{{An Ant System Based Exploration-Exploitation for
  Reinforcement Learning}},
\newblock \bibinfo{journal}{Conference Proceedings - IEEE International
  Conference on Systems, Man and Cybernetics} \bibinfo{volume}{4}
  (\bibinfo{year}{2004}) \bibinfo{pages}{3805--3810}.
  \DOIprefix\doi{10.1109/ICSMC.2004.1400937}.
\bibitem[{Grossberg(1987)}]{Grossberg1987}
\bibinfo{author}{S.~Grossberg},
\newblock \bibinfo{title}{{Competitive Learning: From Interactive Activation To
  Adaptive Resonance}},
\newblock \bibinfo{journal}{Cognitive Science} \bibinfo{volume}{11}
  (\bibinfo{year}{1987}) \bibinfo{pages}{23--63}.
  \DOIprefix\doi{10.1016/S0364-0213(87)80025-3}.
\bibitem[{Teng and Tan(2012)}]{Teng2012a}
\bibinfo{author}{T.~H. Teng}, \bibinfo{author}{A.~H. Tan},
\newblock \bibinfo{title}{{Knowledge-based Exploration for Reinforcement
  Learning in Self-organizing Neural Networks}},
\newblock \bibinfo{journal}{Proceedings - 2012 IEEE/WIC/ACM International
  Conference on Intelligent Agent Technology, IAT 2012}  (\bibinfo{year}{2012})
  \bibinfo{pages}{332--339}. \DOIprefix\doi{10.1109/WI-IAT.2012.154}.
\bibitem[{Wang et~al.(2018)Wang, Zhou, Wang, and Tan}]{Wang2018a}
\bibinfo{author}{P.~Wang}, \bibinfo{author}{W.~J. Zhou},
  \bibinfo{author}{D.~Wang}, \bibinfo{author}{A.~H. Tan},
\newblock \bibinfo{title}{{Probabilistic guided exploration for reinforcement
  learning in self-organizing neural networks}},
\newblock \bibinfo{journal}{Proceedings - 2018 IEEE International Conference on
  Agents, ICA 2018}  (\bibinfo{year}{2018}) \bibinfo{pages}{109--112}.
  \URLprefix \url{http://arxiv.org/abs/1603.04119}.
  \DOIprefix\doi{10.1109/AGENTS.2018.8460067}.
\bibitem[{R{\"{u}}ckstie{\ss} et~al.(2010)R{\"{u}}ckstie{\ss}, Sehnke, Schaul,
  Wierstra, Sun, and Schmidhuber}]{Ruckstiess2010}
\bibinfo{author}{T.~R{\"{u}}ckstie{\ss}}, \bibinfo{author}{F.~Sehnke},
  \bibinfo{author}{T.~Schaul}, \bibinfo{author}{D.~Wierstra},
  \bibinfo{author}{Y.~Sun}, \bibinfo{author}{J.~Schmidhuber},
\newblock \bibinfo{title}{{Exploring Parameter Space in Reinforcement
  Learning}},
\newblock \bibinfo{journal}{Journal of Behavioral Robotics} \bibinfo{volume}{1}
  (\bibinfo{year}{2010}) \bibinfo{pages}{14--24}.
  \DOIprefix\doi{10.2478/s13230-010-0002-4}.
\bibitem[{Shibata and Sakashita(2015)}]{Shibata2015}
\bibinfo{author}{K.~Shibata}, \bibinfo{author}{Y.~Sakashita},
\newblock \bibinfo{title}{{Reinforcement Learning With Internal-Dynamics-Based
  Exploration Using A Chaotic Neural Network}},
\newblock \bibinfo{journal}{Proceedings of the International Joint Conference
  on Neural Networks, IJCNN 2015}  (\bibinfo{year}{2015}).
  \DOIprefix\doi{10.1109/IJCNN.2015.7280430}.
\bibitem[{Fortunato et~al.(2018)Fortunato, Azar, Piot, Menick, Hessel, Osband,
  Graves, Mnih, Munos, Hassabis, Pietquin, Blundell, and Legg}]{Fortunato2018}
\bibinfo{author}{M.~Fortunato}, \bibinfo{author}{M.~G. Azar},
  \bibinfo{author}{B.~Piot}, \bibinfo{author}{J.~Menick},
  \bibinfo{author}{M.~Hessel}, \bibinfo{author}{I.~Osband},
  \bibinfo{author}{A.~Graves}, \bibinfo{author}{V.~Mnih},
  \bibinfo{author}{R.~Munos}, \bibinfo{author}{D.~Hassabis},
  \bibinfo{author}{O.~Pietquin}, \bibinfo{author}{C.~Blundell},
  \bibinfo{author}{S.~Legg},
\newblock \bibinfo{title}{{Noisy networks for exploration}},
\newblock \bibinfo{journal}{6th International Conference on Learning
  Representations, ICLR 2018}  (\bibinfo{year}{2018}) \bibinfo{pages}{1--21}.
  \href{http://arxiv.org/abs/1706.10295}{{\tt arXiv:1706.10295}}.
\bibitem[{Plappert et~al.(2018)Plappert, Houthooft, Dhariwal, Sidor, Chen,
  Chen, Asfour, Abbeel, and Andrychowicz}]{Plappert2018a}
\bibinfo{author}{M.~Plappert}, \bibinfo{author}{R.~Houthooft},
  \bibinfo{author}{P.~Dhariwal}, \bibinfo{author}{S.~Sidor},
  \bibinfo{author}{R.~Y. Chen}, \bibinfo{author}{X.~Chen},
  \bibinfo{author}{T.~Asfour}, \bibinfo{author}{P.~Abbeel},
  \bibinfo{author}{M.~Andrychowicz},
\newblock \bibinfo{title}{{Parameter Space Noise for Exploration}},
\newblock \bibinfo{journal}{6th International Conference on Learning
  Representations, ICLR 2018}  (\bibinfo{year}{2018}) \bibinfo{pages}{1--18}.
  \href{http://arxiv.org/abs/1706.01905}{{\tt arXiv:1706.01905}}.
\bibitem[{R{\"{u}}ckstie{\ss} et~al.(2008)R{\"{u}}ckstie{\ss}, Felder, and
  Schmidhuber}]{Ruckstiess2008}
\bibinfo{author}{T.~R{\"{u}}ckstie{\ss}}, \bibinfo{author}{M.~Felder},
  \bibinfo{author}{J.~Schmidhuber},
\newblock \bibinfo{title}{{State-dependent Exploration for Policy Gradient
  Methods}},
\newblock \bibinfo{journal}{Lecture Notes in Computer Science (including
  subseries Lecture Notes in Artificial Intelligence and Lecture Notes in
  Bioinformatics)} \bibinfo{volume}{5212 LNAI} (\bibinfo{year}{2008})
  \bibinfo{pages}{234--249}. \DOIprefix\doi{10.1007/978-3-540-87481-2_16}.
\bibitem[{Osband et~al.(2020)Osband, Doron, Hessel, Aslanides, Sezener,
  Saraiva, McKinney, Lattimore, Szepezvari, Singh, van Roy, Sutton, Silver, and
  van Hasselt}]{Osband2020}
\bibinfo{author}{I.~Osband}, \bibinfo{author}{Y.~Doron},
  \bibinfo{author}{M.~Hessel}, \bibinfo{author}{J.~Aslanides},
  \bibinfo{author}{E.~Sezener}, \bibinfo{author}{A.~Saraiva},
  \bibinfo{author}{K.~McKinney}, \bibinfo{author}{T.~Lattimore},
  \bibinfo{author}{C.~Szepezvari}, \bibinfo{author}{S.~Singh},
  \bibinfo{author}{B.~van Roy}, \bibinfo{author}{R.~Sutton},
  \bibinfo{author}{D.~Silver}, \bibinfo{author}{H.~van Hasselt},
\newblock \bibinfo{title}{{Behaviour Suite for Reinforcement Learning}},
\newblock \bibinfo{journal}{8th International Conference on Learning
  Representations, ICLR 2020}  (\bibinfo{year}{2020}).
  \href{http://arxiv.org/abs/1908.03568}{{\tt arXiv:1908.03568}}.
\bibitem[{Jaegle et~al.(2019)Jaegle, Mehrpour, and Rust}]{Jaegle2019a}
\bibinfo{author}{A.~Jaegle}, \bibinfo{author}{V.~Mehrpour},
  \bibinfo{author}{N.~Rust},
\newblock \bibinfo{title}{{Visual Novelty, Curiosity, And Intrinsic Reward in
  Machine Learning And The Brain}},
\newblock \bibinfo{journal}{Current Opinion in Neurobiology}
  \bibinfo{volume}{58} (\bibinfo{year}{2019}) \bibinfo{pages}{167--174}.
  \URLprefix \url{https://doi.org/10.1016/j.conb.2019.08.004}.
  \DOIprefix\doi{10.1016/j.conb.2019.08.004}.
  \href{http://arxiv.org/abs/1901.02478}{{\tt arXiv:1901.02478}}.
\bibitem[{OpenAI(2021)}]{OpenAI2021}
\bibinfo{author}{OpenAI},
\newblock \bibinfo{title}{{Asymmetric Self-Play for Automatic Goal Discovery in
  Robotic Manipulation}}  (\bibinfo{year}{2021}). \URLprefix
  \url{http://arxiv.org/abs/2101.04882}.
  \href{http://arxiv.org/abs/2101.04882}{{\tt arXiv:2101.04882}}.
\bibitem[{Zhang et~al.(2019)Zhang, Tang, and Yao}]{Zhang2019a}
\bibinfo{author}{L.~Zhang}, \bibinfo{author}{K.~Tang},
  \bibinfo{author}{X.~Yao},
\newblock \bibinfo{title}{{Explicit Planning for Efficient Exploration in
  Reinforcement Learning}},
\newblock \bibinfo{journal}{Conference on Neural Information Processing
  Systems, NeurIPS 2019} \bibinfo{volume}{32} (\bibinfo{year}{2019}).

\end{thebibliography}

\end{document}